\pdfoutput=1

\documentclass[11pt]{article}
\usepackage[utf8]{inputenc}
\usepackage{CJKutf8}
\usepackage{scrextend}

\usepackage[]{ACL2023}

\usepackage{times}
\usepackage{latexsym}
\usepackage{cuted}

\usepackage{colortbl}
\usepackage[T1]{fontenc}
\usepackage{graphicx}

\newcommand{\inlinejp}[1]{\begin{CJK}{UTF8}{min}{#1}\end{CJK}}
\newcommand{\inlinezh}[1]{\begin{CJK}{UTF8}{gbsn}{#1}\end{CJK}}

\definecolor{disYellow}{HTML}{D6B656}
\definecolor{covBlue}{HTML}{6C8EBF}
\definecolor{conRed}{HTML}{B85450}

\usepackage{subcaption}

\newcommand{\adjustimg}{%
  \hspace*{\dimexpr\evensidemargin-\oddsidemargin}%
}
\newcommand{\centerimg}[2][width=\textwidth]{%
  \makebox[\textwidth]{\adjustimg\includegraphics[#1]{#2}}%
}

\newcommand{\cococrola}[0]{{\color{red} \textbf{\textit{CoCo-CroLa}}}}

\usepackage{microtype}

\usepackage{inconsolata}

\usepackage{cleveref}
\usepackage{booktabs}

\newcommand{\samel}[0]{\multicolumn{1}{l}{\space\hspace{0.5em}\vline}}
\usepackage{comment, pifont}

\title{Multilingual Conceptual Coverage in Text-to-Image Models}

\author{Michael Saxon \\
  University of California, Santa Barbara \\
  \texttt{saxon@ucsb.edu} \\
  \And 
  William Yang Wang \\
  University of California, Santa Barbara \\
  \texttt{william@cs.ucsb.edu} \\
  }

\begin{document}
\maketitle

\vspace{-20pt}

\begin{strip} 

{
    \vspace{-58pt}

    \noindent\centerimg[width=\linewidth]{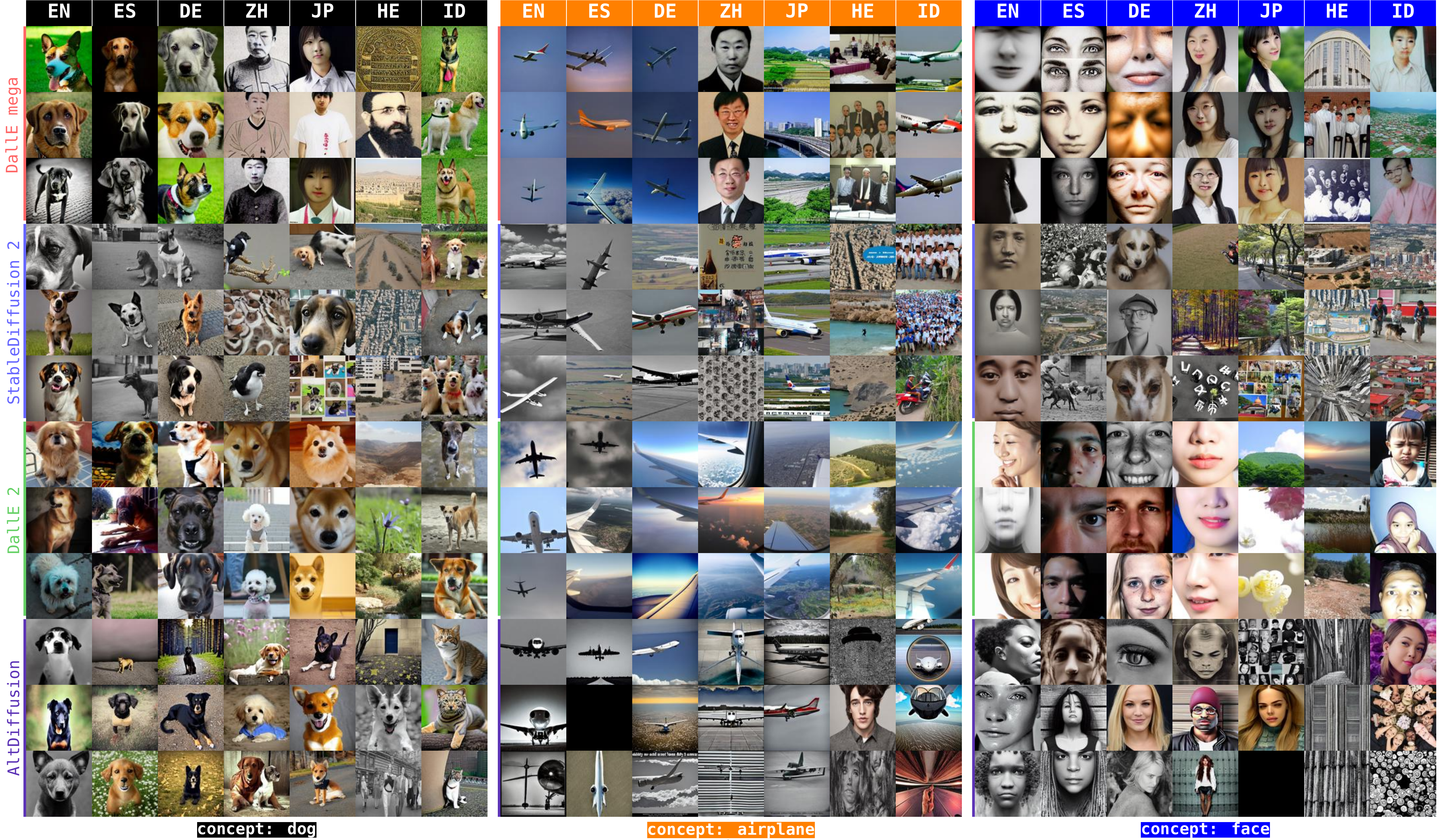}

      \fontsize{10pt}{12pt}\selectfont Figure 1: A selection of images generated by DALLE-mega, Stable Diffusion 2, DALLE-2, and AltDiffusion, illustrating their \textit{conceptual coverage} of ``\textbf{dog},''  ``\textbf{\color{orange} airplane},'' and ``\textbf{\color{blue} face}'' across English, Spanish, German, Chinese (simplified), Japanese, Hebrew, and Indonesian.  Coverage of the concepts varies considerably across model and language, and can be observed in the consistency and correctness of images generated under simple prompts.
    \label{fig:teaser}

}

\end{strip}

     \begin{abstract}
 We propose ``\textbf{Co}nceptual \textbf{Co}verage A\textbf{cro}ss \textbf{La}nguages'' (\cococrola), a technique for benchmarking the degree to which any generative text-to-image system provides multilingual parity to its training language in terms of tangible nouns. For each model we can assess ``conceptual coverage'' of a given target language relative to a source language by comparing the population of images generated for a series of tangible nouns in the source language to the population of images generated for each noun under translation in the target language. This technique allows us to estimate how well-suited a model is to a target language as well as identify model-specific weaknesses, spurious correlations, and biases without a-priori assumptions. We demonstrate how it can be used to benchmark T2I models in terms of multilinguality, and how despite its simplicity it is a good proxy for impressive generalization.
     \end{abstract}

\setcounter{figure}{1}

\section{Introduction}

Neural text-to-image models convert text prompts into images \cite{mansimov2015generating, reed2016generative} using internal representations reflective of the training data population. 
Advancements in conditional language modeling \cite{lewis2019bart}, variational autoencoders \cite{kingma2013auto}, GANs \cite{goodfellow2020generative}, multimodal representations \cite{radford2021learning}, and latent diffusion models \cite{rombach2022high} have given rise to sophisticated text-to-image (T2I) systems
that exhibit impressive \textit{semantic generalization capabilities}, 
with which they generate coherent, visually-appealing images with novel combinations of objects, scenarios, and styles \cite{ramesh2021zero}.
Their semantic latent spaces \cite{kwon2022diffusion} ground words to associated visuals \cite{hutchinson2022underspecification}. %
Characterizing the limits of these systems' capabilities is a challenge. They are composed of elements trained on incomprehensibly large \cite{prabhu2020large, jia2021scaling}, web-scale data \cite{gao2020pile, schuhmann2021laion}, hindering training-data-centric model analysis \cite{mitchell2019model, gebru2021datasheets} to address this problem.

\begin{figure}[t!]
    \centering
    \includegraphics[width=0.25\linewidth]{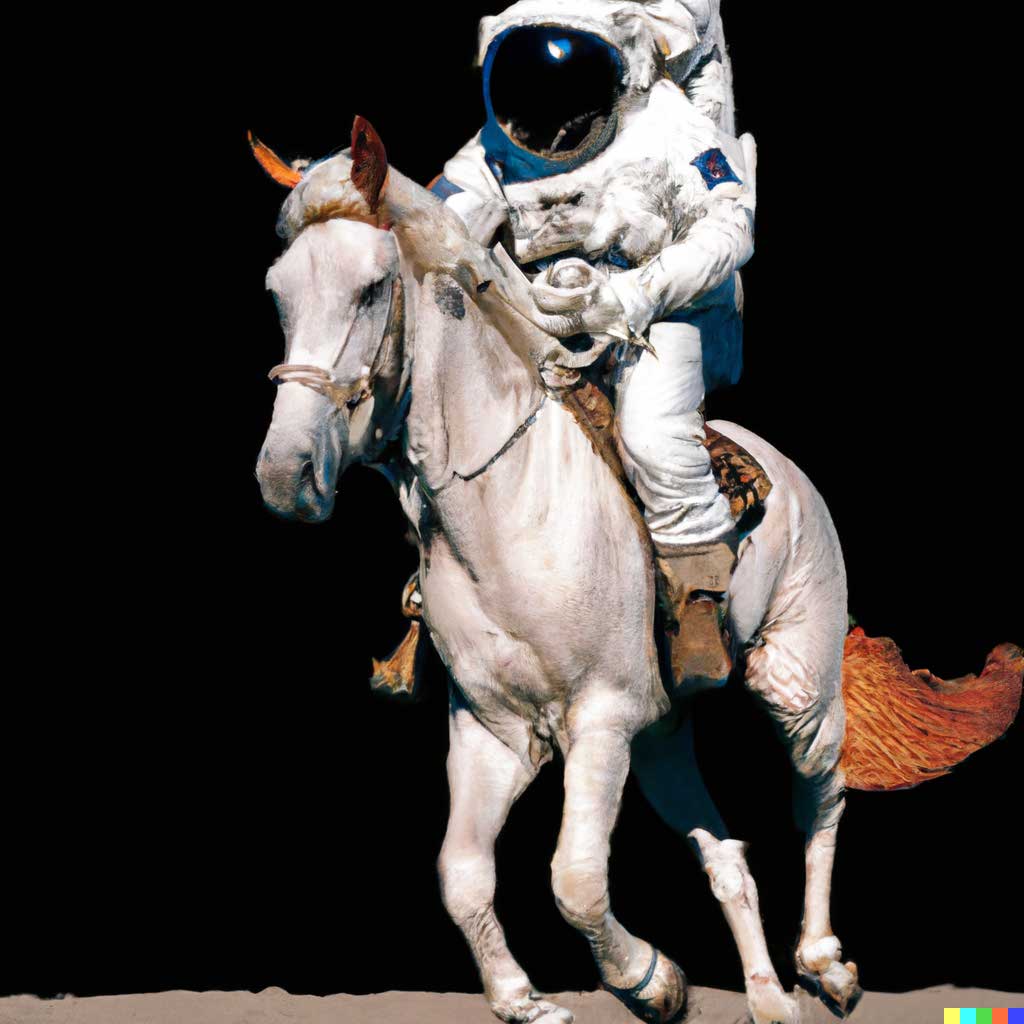}\includegraphics[width=0.25\linewidth]{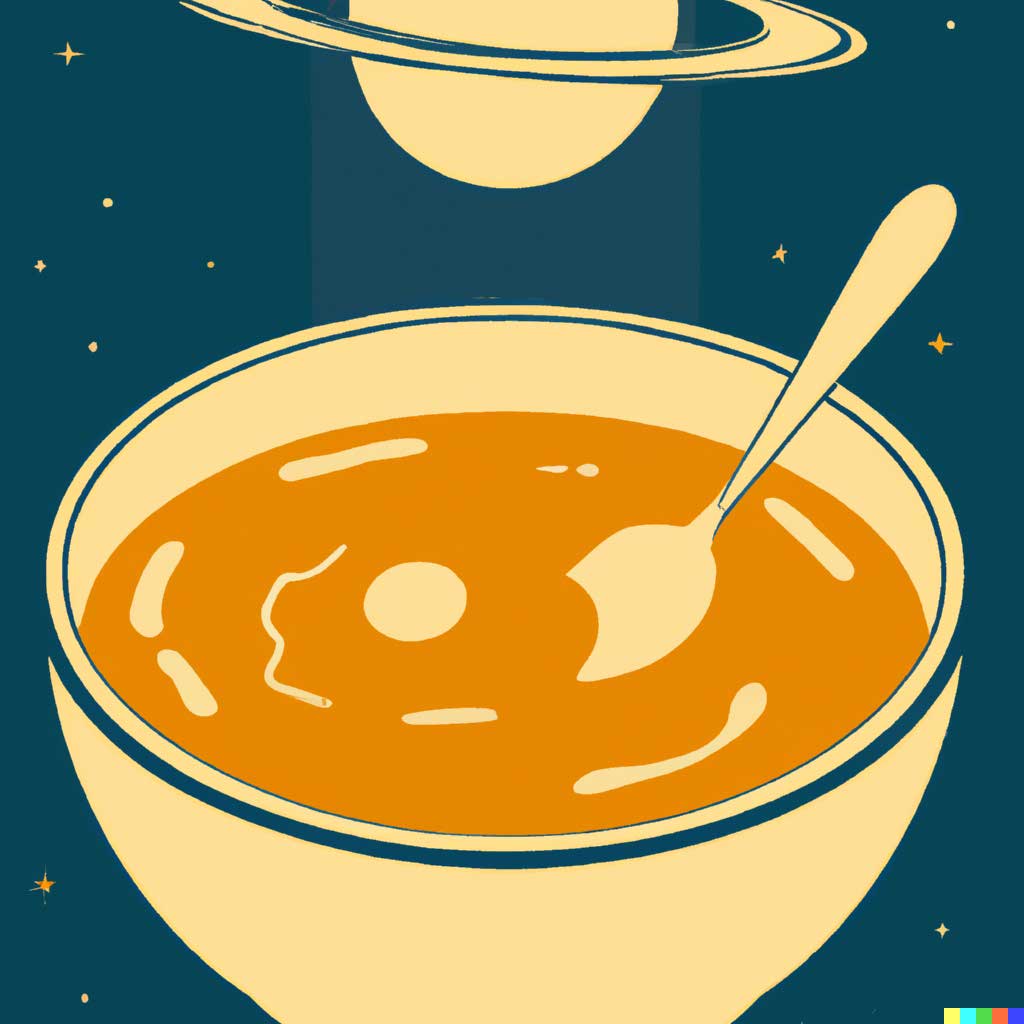}\includegraphics[width=0.25\linewidth]{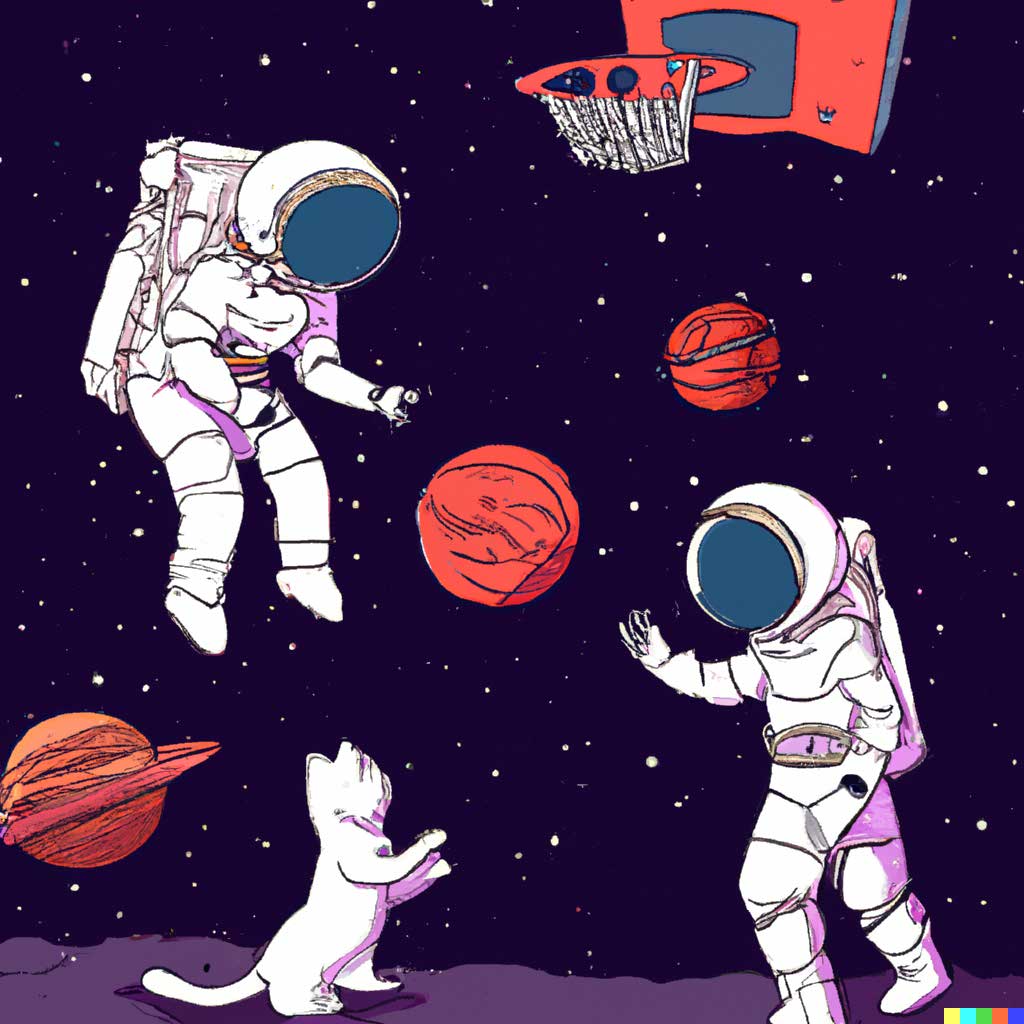}\includegraphics[width=0.25\linewidth]{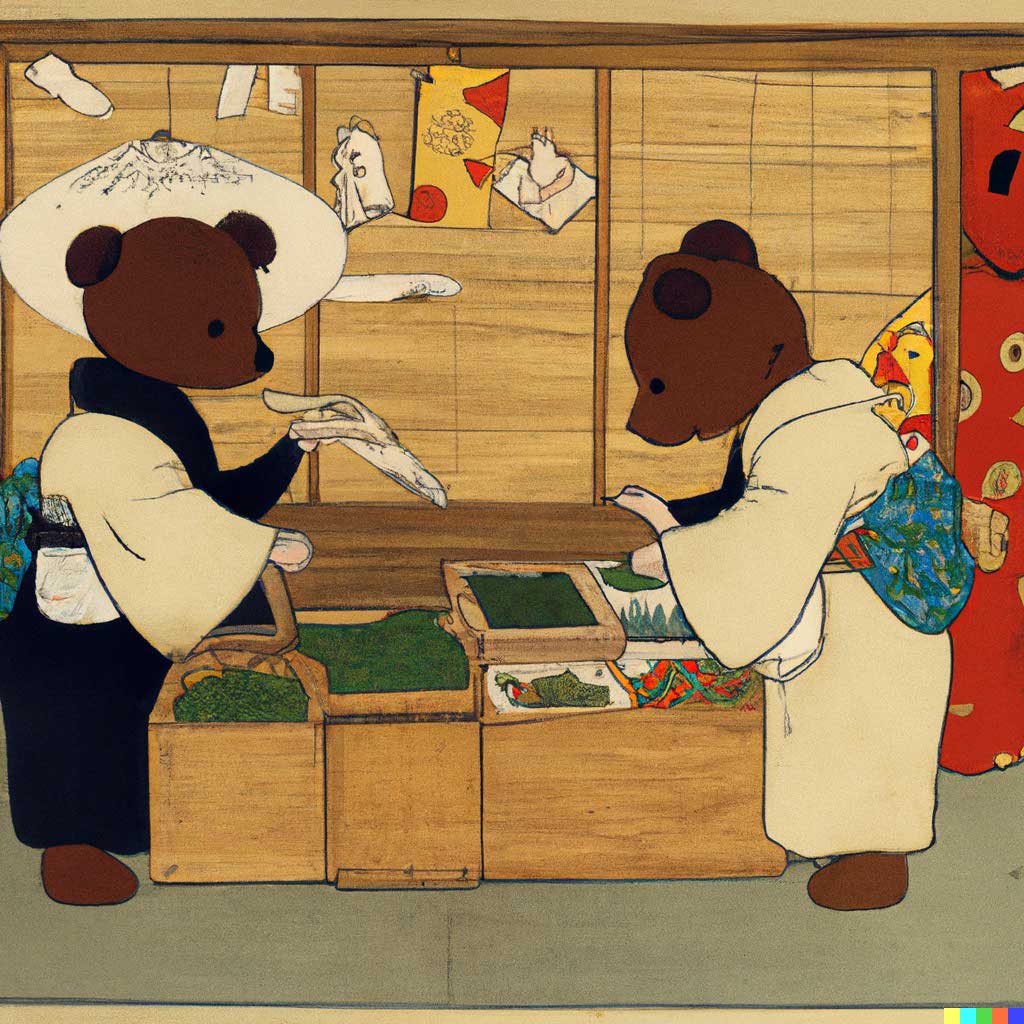}
    \caption{We hypothesize that a model's ability to generate creative, compositional images depicting tangible concepts (e.g., astronaut, horse, soup, bear) is predicated on its ability to generate simple images of the concepts alone. Samples from \citet{ramesh2022hierarchical}.}
    \label{fig:astrohorse}
\end{figure}

Demonstrations of novel T2I model capabilities tend to rely on the subjective impressiveness of their ability to generalize to complex, novel prompts (\autoref{fig:astrohorse}). Unfortunately, the space of creative prompts is in principle infinite. However, we observe that \textbf{impressive creative prompts are composed of known, tangible concepts}.

Can we directly evaluate a model's knowledge of these tangible concepts as a partial proxy for its capability to generalize to creative novel prompts? Perhaps. 
But finding a diverse set of significant failure cases of basic concept knowledge for these models is challenging---in their training language.

\begin{figure}
    \centering
    \tiny \texttt{EN}\hspace{0.22\linewidth}\texttt{ES}\hspace{0.22\linewidth}\texttt{ID}\hspace{0.22\linewidth}\texttt{JA}\vspace{-0.5pt}
    \includegraphics[width=0.12\linewidth]{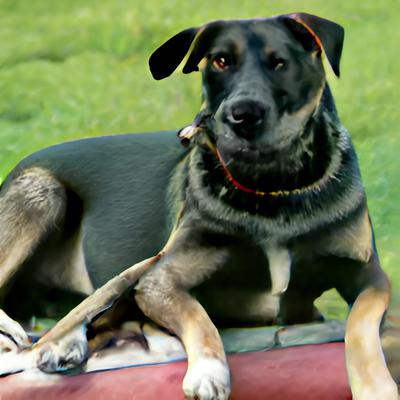}\includegraphics[width=0.12\linewidth]{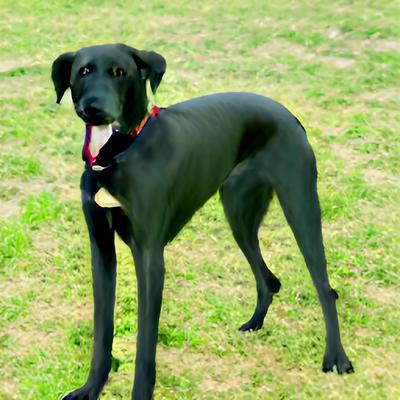}\hspace{0.01\linewidth}\includegraphics[width=0.12\linewidth]{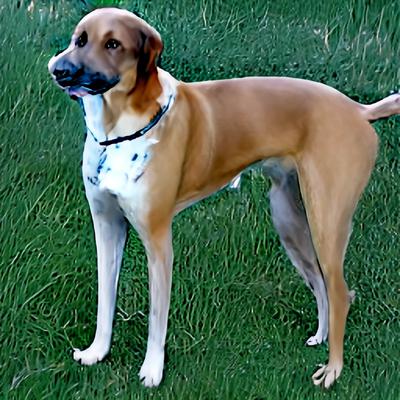}\includegraphics[width=0.12\linewidth]{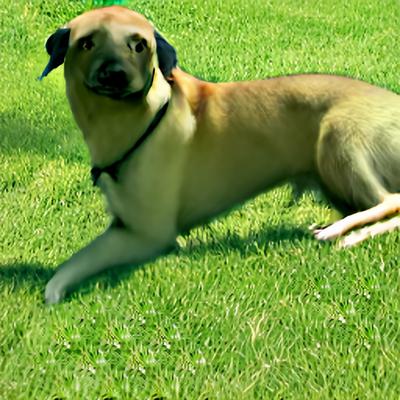}\hspace{0.01\linewidth}\includegraphics[width=0.12\linewidth]{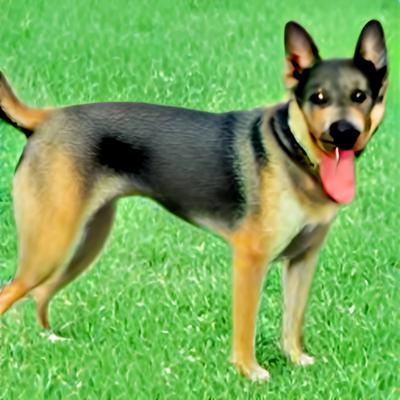}\includegraphics[width=0.12\linewidth]{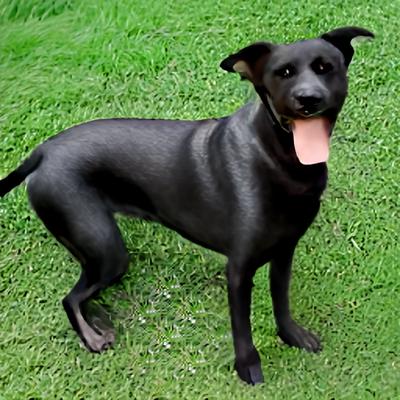}\hspace{0.01\linewidth}\includegraphics[width=0.12\linewidth]{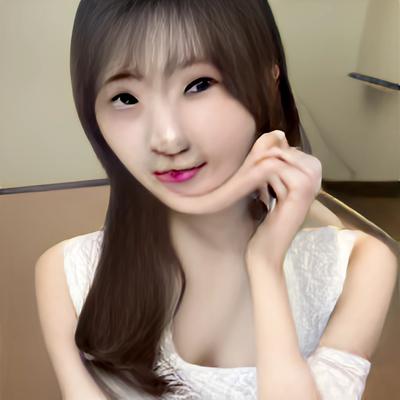}\includegraphics[width=0.12\linewidth]{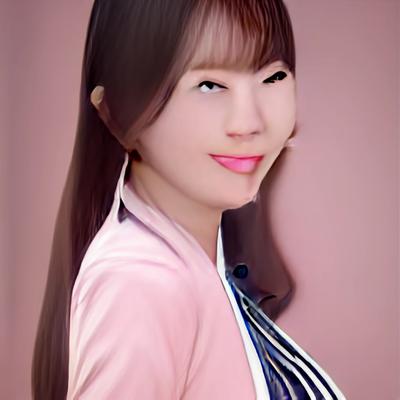}\vspace{-2pt}\\
    \includegraphics[width=0.12\linewidth]{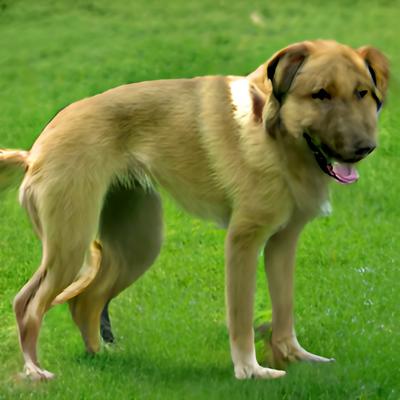}\includegraphics[width=0.12\linewidth]{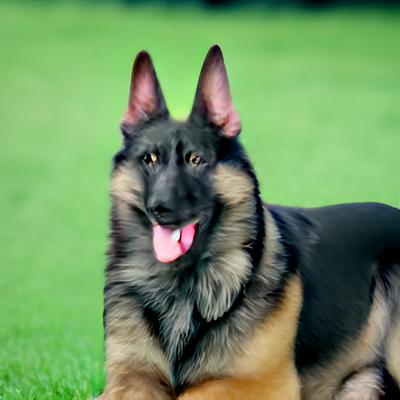}\hspace{0.01\linewidth}\includegraphics[width=0.12\linewidth]{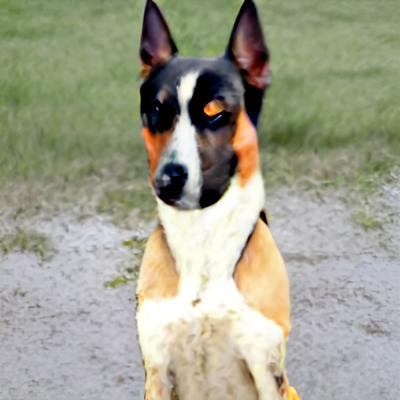}\includegraphics[width=0.12\linewidth]{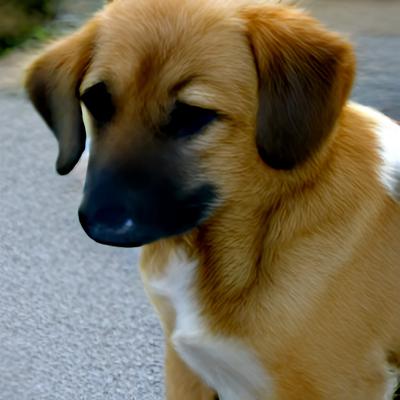}\hspace{0.01\linewidth}\includegraphics[width=0.12\linewidth]{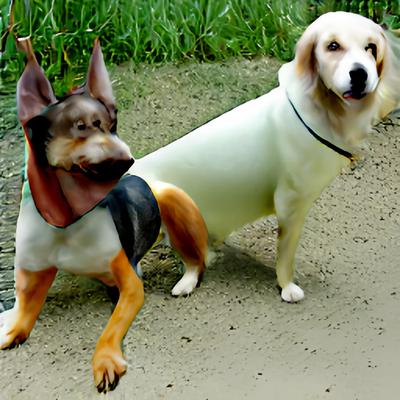}\includegraphics[width=0.12\linewidth]{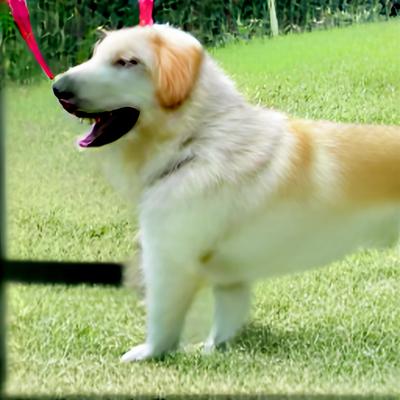}\hspace{0.01\linewidth}\includegraphics[width=0.12\linewidth]{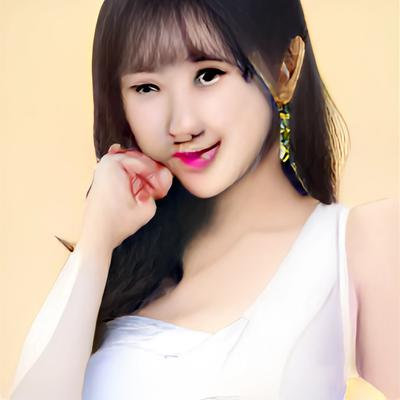}\includegraphics[width=0.12\linewidth]{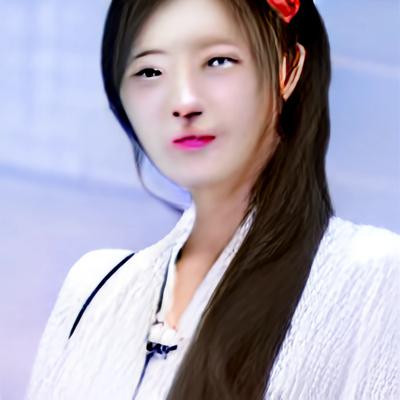}\vspace{-2pt}\\
    \caption{Although DALL-E mini \cite{dayama2021dallemini} is ostensibly trained only on English data, when elicited with ``big dog'' in Spanish, Indonesian, and Japanese it generalizes the ``dog'' concept to ES and ID, while exhibiting an offensive concept-level collision in JA.}
    \label{fig:incite}
\end{figure}

\begin{figure*}[t!]
    \centering
    \includegraphics[width=1\linewidth]{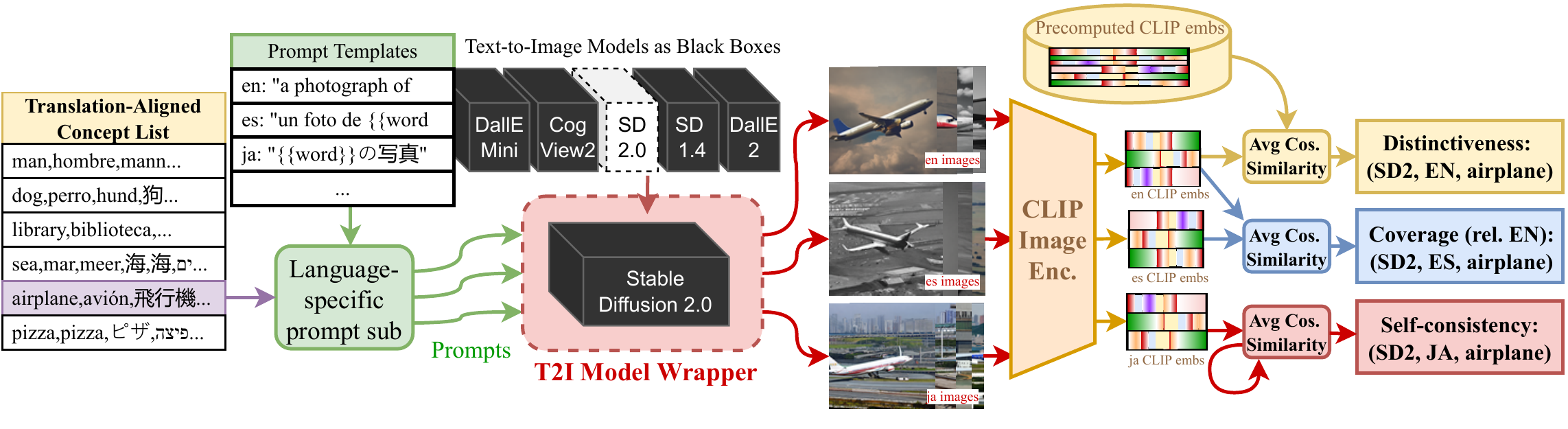}
    \vspace{-12pt}
    \caption{\cococrola~ assesses the cross-lingual coverage of a concept in a model by plugging all the term translations into prompt templates, generating a set of images from a model under test, extracting their corresponding CLIP embeddings, and computing concept-level \textbf{\color{disYellow}distinctiveness}, \textbf{\color{covBlue}coverage}, and \textbf{\color{conRed}self-consistency} for the concept with respect to each language. (Demo and code available at \href{https://github.com/michaelsaxon/CoCoCroLa}{\small\texttt{github.com/michaelsaxon/CoCoCroLa}})}
    \label{fig:modelscoring}
\end{figure*}

We observe that when prompted with simple requests for specific tangible concepts in a constrained style, T2I models can sometimes generate consistent and semantically-correct images in languages for which they received limited training (\hyperref[fig:teaser]{Figure 1}, \autoref{fig:incite}). We refer to this capacity as \textit{concept possession} by said model in said language.
At scale, we can assess the language-concept possession for a diverse array of concepts and languages in a model to attempt to describe its overall multilingual generalization capability. We refer to the degree of this capability as the model's multilingual \textit{conceptual coverage}.
In this work we:

\begin{enumerate}
    \item Introduce objective measures of \textit{multilingual conceptual coverage} in T2I models that compare images generated from equivalent prompts under translation (\autoref{fig:modelscoring}).
    \item Release \cococrola, a benchmark set for conceptual coverage testing of 193 tangible concepts across English, Spanish, German, Chinese, Japanese, Hebrew, and Indonesian.
    \item Validate the utility of \textit{conceptual coverage analysis} with a preliminary pilot study suggesting that generalization to complex, creative prompts follows concept possession.
\end{enumerate}

Our benchmark enables fine-grained concept-level model analysis and identification of novel failure modes, and will guide future work in increasing the performance, explainability, and linguistic parity of text-to-image models.

\section{Motivation \& Related Work}

This work is an attempt to produce a scalable, precise technique for characterizing conceptual coverage with \textbf{minimal assumptions} about the concepts or models themselves. 
In this section we lay out our motivations alongside relevant related work. %

\textbf{Benchmarks enabling model comparability have been a driving force} in the development of pretrained language models (LM) \cite{devlin2018bert}. For classification and regression tasks, evaluation under fine-tuning \cite{howard2018universal,karpukhin2020dense} is a straightforward and practical proxy for pretrained LM quality \cite{dodge2020fine} (e.g., for encoder-only transformer networks \cite{liu2019roberta}). For these classification models, higher performance on benchmark datasets \cite{lai2017race, rajpurkar2018know, wang2019superglue} became the epitome of LM advancement.
However, other important qualities in models including degree of social biases \cite{sheng2019woman} and robustness \cite{clark-etal-2019-dont} arising from biases in training data \cite{saxon2021automatically} can only be captured by more sophisticated benchmarks that go beyond simple accuracy \cite{cho2021checkdst}.
CheckList represented an influential move in this direction by benchmarking model performance through \textit{behavioral analysis under perturbed elicitation} \cite{checklist:acl20}. 

In contrast, generative large language models (LLMs) such as GPT-3 \cite{brown2020language} have a broader range of outputs, use-cases, and capabilities, making evaluation more difficult. For many text-generative tasks such as summarization and creative text generation, the crucial desired quality is subjective, and challenging to evaluate \cite{xu2022not}.
However, as these LLMs operate in a text-only domain, existing supervised tasks could be ported to few-shot or zero-shot evaluations of LLM capabilities \cite{srivastava2022beyond}.
While performance on these benchmarks isn't directly indicative of the impressive generative performance and generalization capabilities, they are a means to measure improvement \cite{suzgun2022challenging}.

\textbf{Text-to-image models are even more difficult to evaluate than LLMs.} Unlike in LLMs, there are no objective evaluation tasks that can be directly ported as proxy evaluations. For example, while GPT-3 was introduced with impressive zero-shot performance across many classification tasks, the T2I model DALL-E 2 was primarily introduced with human opinion scores and cool demo images \cite{ramesh2022hierarchical}. 
Prior efforts in developing T2I evaluations such as Drawbench \cite{saharia2022photorealistic} and DALL-Eval \cite{Cho2022DallEval} fall into the trap of trying to build “everything benchmarks” for which whether the
benchmark accurately reflects the practical task being asked of the computer in its real-world context is difficult to assess \cite{raji2021everything}. We instead seek to build an \textit{atomic benchmark} which narrowly and reliably captures a specific characteristic---conceptual knowledge as reflected by a model's ability to reliably generate images of an object.

\textbf{Multilingual conceptual coverage is a high-variation T2I model performance setting.} (\hyperref[fig:teaser]{Figure 1}) Perhaps more importantly, it has immediate value, as work on improving T2I model multilinguality has has been proposed, but hampered by a lack of evaluation metrics.

\citet{chen2022altclip} introduce AltCLIP and AltDiffusion, models produced by performing multilingual contrastive learning on a CLIP checkpoint for an array of non-English languages including Japanese, Chinese, and Korean. Without an objective evaluation benchmark, they can only demonstrate their improvement through human evaluation of impressive but arbitrary examples. 
\cococrola~improves this state of affairs by enabling CheckList-like direct comparison of techniques for reducing \textit{multilingual conceptual coverage} disparities as an objective, capabilities-based benchmark.

Excitingly, we find that \textbf{conceptual coverage is upstream of the impressive T2I model creativity} that model developers and end-users are fundamentally interested in.
This means that not only is \cococrola~an  objective evaluation of T2I system capabilities, it is also a \textbf{proxy measure for the deeper semantic generalization capabilities we are interested in enhancing} in second languages, as we demonstrate in \autoref{subsec:conceptprop}.

\begin{figure*}[t!]
    \centering
    \includegraphics[width=1\linewidth]{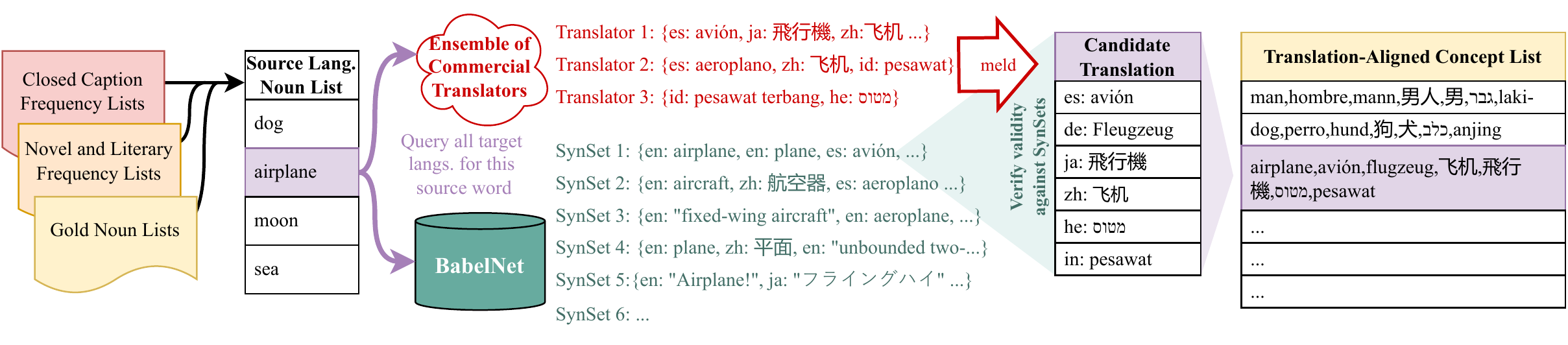}
    \vspace{-12pt}
    \caption{A diagram of our approach for producing the aligned noun concept list across the target language set using an ensemble of cloud translation services and BabelNet. Full description of this method in \autoref{subsec:translation}.}
    \label{fig:aligneddict}
\end{figure*}

\section{Definitions \& Formulations}\label{subsec:formulation}

We define a multilingual \textit{concept} over languages $L$ as a set of words in each language carrying the same meaning and analogous colloquial use. %
We refer to the equivalent translation of concept $c_k$ in language $\ell$ as $c_{k,\ell}$.

Given a set of concepts $C$, test language $\ell$, a \textit{minimal eliciting prompt}\footnote{We define a minimal eliciting prompt as a short sentence with a slot for concept work insertion, intended to enforce style consistency without interfering with the concept.} $\textrm{MP}_\ell$, text-to-image model $f$, and a desired number of images-per-concept $n$, we sample $n|C||L|$ images $I_{c_{k},\ell,i}$, where

\begin{equation}
    I_{c_{k},\ell,i} \sim f(\textrm{MP}_\ell(c_{k,\ell}))
\end{equation}

\noindent For every concept word in the language $\ell$ $c_{k}\in C$.

Given an image feature extractor $F$, some similarity function $\textsc{Sim}(\cdot, \cdot)$, 
we assess whether $f$ \textit{possesses} concept $c_{k,\ell}$ in the test language if using the following metrics from the concept-image set $\{I_{c_{k,\ell},i}\}_{i=0}^n$ (\cococrola~scores in \autoref{fig:modelscoring}):

\paragraph{\color{disYellow} Distinctiveness.}
The images are \textit{distinct} if they tend to not resemble the population of images generated for other concepts in the target language.

Formally, we compute our (inverse) distinctiveness score $\textrm{Dt}(f,\ell,c_k)$ relative to $m$ images sampled from other concepts in $C$:

\begin{equation}
    \textrm{Dt} = \frac{1}{mn}\sum_{j=0}^{m}\sum_{i=0}^n\textsc{Sim}(F(I_{c_k,\ell,i}), F(I_{c_r,\ell,s})),
\end{equation}
\begin{equation}
    c_{r,\ell} \sim C \setminus c_{k,\ell}, \quad s \sim U\{0, n\}
\end{equation}

\paragraph{\color{conRed}Self-consistency.}
The images are \textit{self-consistent} if they tend to resemble each other as a set.

Formally, we compute the self-consistency score $\textrm{Sc}(f,\ell,c_k)$ as:
\begin{equation}\scriptstyle
    \textrm{Sc} = {\frac{1}{n^2-n}}\sum_{j=0}^{n}\left(\sum_{i=0}^n\textsc{Sim}(F(I_{c_{k,\ell},i}), F(I_{c_{k,\ell},j}))-1\right)
\end{equation}

We subtract 1 from each step in the numerator and $n$ from the denominator so that identical matches generated image to itself.

\paragraph{\color{covBlue}Correctness.}
The images are \textit{correct} if they faithfully depict the object being described.

Rather than assess this using a classification model (hindering generality depending on the pretrained classifier), we use faithfulness relative to a source language $\ell s$, cross consistency $\textrm{Xc}(f,\ell,c_k, \ell s)$ as a proxy:

\begin{equation}
    \textrm{Xc} = \frac{1}{n^2}\sum_{j=0}^{n}\sum_{i=0}^n\textsc{Sim}(F(I_{c_{k,\ell},i}), F(I_{c_{k,\ell s},j}))
\end{equation}

It is important to note that in the source language (eg. English), $\textrm{Xc}\approx\textrm{Sc}$, as for that language both metrics are essentially comparing the same (concept, language) pair to itself.

Thus we augment with a second language-grounded correctness score, utilizing the average text-image similarity score of the English concept text against the set of generated images, for a CLIP image encoder $F$ and text encoder $F_t$, Wc:

\begin{equation}
    \textrm{Wc} = \frac{1}{n}\sum_{i=0}^n F_t(c_{k,\ell_s}) \cdot F(I_{c_{k,\ell},i})
\end{equation}

\begin{table*}[t!]
\small
    \centering
    \begin{tabular}{llll}
        \toprule
        Model & Authors (Year) & Repository & Training Language  \\
        \midrule
        DALL-E Mini & \citet{dayama2021dallemini} & \href{https://github.com/borisdayma/dalle-mini}{\texttt{github:borisdayma/dalle-mini}} & EN \\
        DALL-E Mega & \samel & \samel & \samel \\
        CogView 2 & \citet{ding2021cogview} & \href{https://github.com/THUDM/CogView}{\texttt{github:THUDM/CogView}} & ZH \\
        StableDiffusion 1.1 & \citet{rombach2022high} & \href{https://huggingface.co/CompVis/stable-diffusion-v1-1}{\texttt{HF:CompVis/stable-diffusion-v1-1}} & EN \\
        StableDiffusion 1.2 & \samel & \href{https://huggingface.co/CompVis/stable-diffusion-v1-2}{\texttt{HF:CompVis/stable-diffusion-v1-2}} & \samel \\
        StableDiffusion 1.4 & \samel & \href{https://huggingface.co/CompVis/stable-diffusion-v1-4}{\texttt{HF:CompVis/stable-diffusion-v1-4}} & No language filter \\
        StableDiffusion 2 & \samel & \href{https://github.com/Stability-AI/stablediffusion}{\texttt{HF:stabilityai/stable-diffusion-2}} & \samel \\
        DALL-E 2 & \citet{ramesh2022hierarchical} & \href{https://openai.com/dall-e-2/}{\texttt{openai.com/dall-e-2/}} (no checkpoints) & No language filter\\
        AltDiffusion m9 & \citet{chen2022altclip} & \href{https://huggingface.co/BAAI/AltDiffusion-m9/tree/main}{\texttt{HF:BAAI/AltDiffusion-m9}} & {\tiny EN, ES, FR, IT, RU, ZH, JA, KO}\\
        \bottomrule
    \end{tabular}
    \vspace{-5pt}
    \caption{The set of text-to-image models we evaluated with {\color{red}\cococrola~v1.0}. %
    Some monolingual models may integrate pretrained elements such as CLIP checkpoints that have been trained on multilingual data.}
    \vspace{-2ex}
    \label{tab:models}
\end{table*}

\section{Approach}
We compute distinctiveness, self-consistency, and correctness scores across English, Spanish, German, Chinese (Simplified), Japanese, Hebrew, and Indonesian on the models listed in \autoref{tab:models}.

For each (language, concept) pair, we generate 10 images for analysis.
We use a CLIP \cite{radford2021learning} checkpoint from HuggingFace\footnote{\href{https://huggingface.co/openai/clip-vit-base-patch32}{\texttt{HF:openai/clip-vit-base-patch32}}.} as our semantic visual feature extractor $F$, and cosine similarity as our similarity function ($\textsc{Sim}(\mathbf{a}, \mathbf{b}) = \mathbf{a}\cdot \mathbf{b} / ||\mathbf{a}||||\mathbf{b}||$). We collect a translation-aligned concept list $C$ using techniques described in \autoref{subsec:nounset} and depicted in \autoref{fig:aligneddict}. We release our list generation code, testing code, feature extraction code, and final concept list as {\color{red}\cococrola~v1.0}\footnote{Demo and code: 
\href{https://github.com/michaelsaxon/CoCoCroLa}{\texttt{github:michaelsaxon/CoCoCroLa}}
}.

\begin{figure*}[t!]
\begin{minipage}{.33\linewidth}
\centering
\includegraphics[width=1\linewidth]{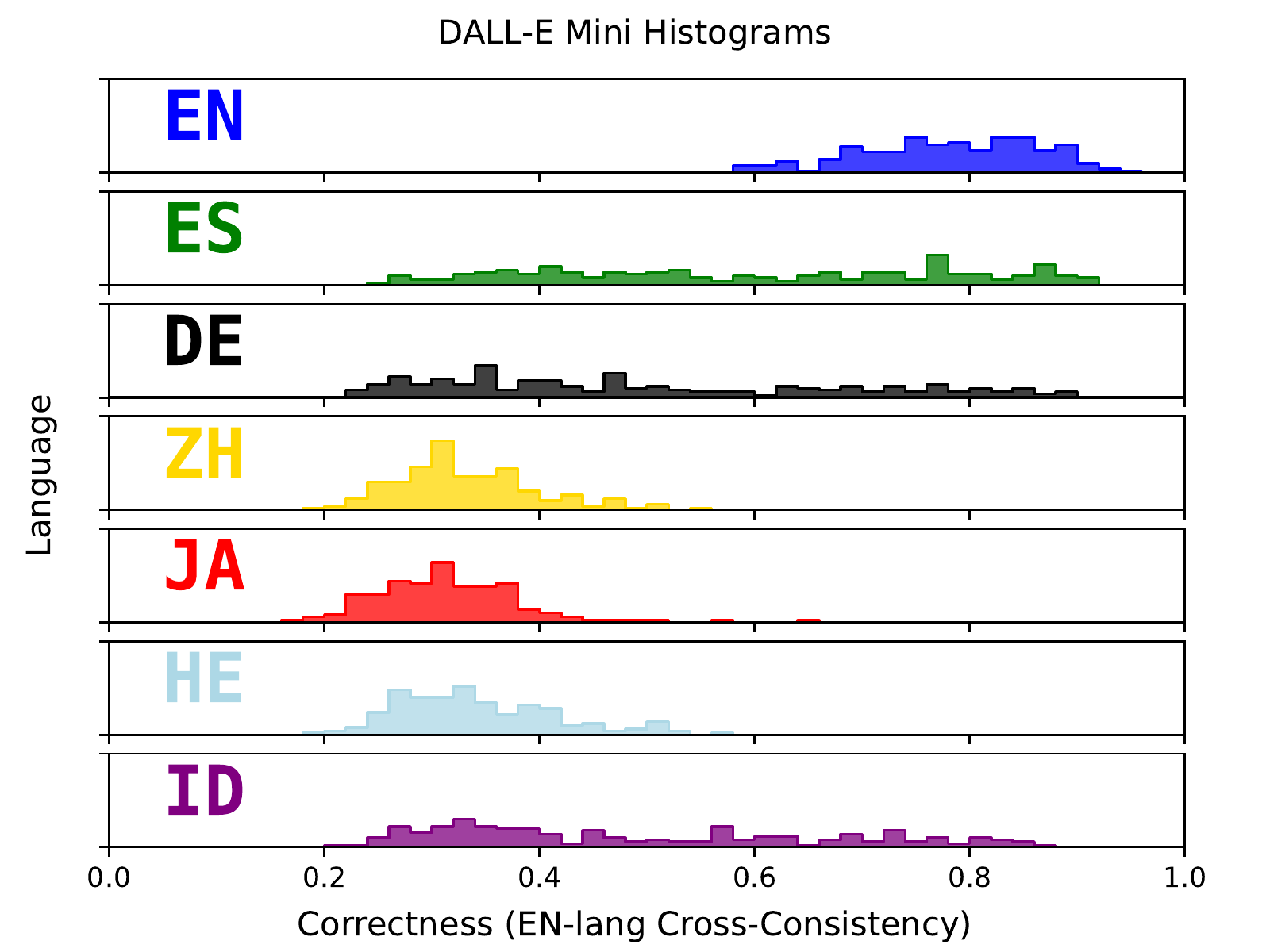}
\end{minipage}%
\begin{minipage}{.33\linewidth}
\centering
\includegraphics[width=1\linewidth]{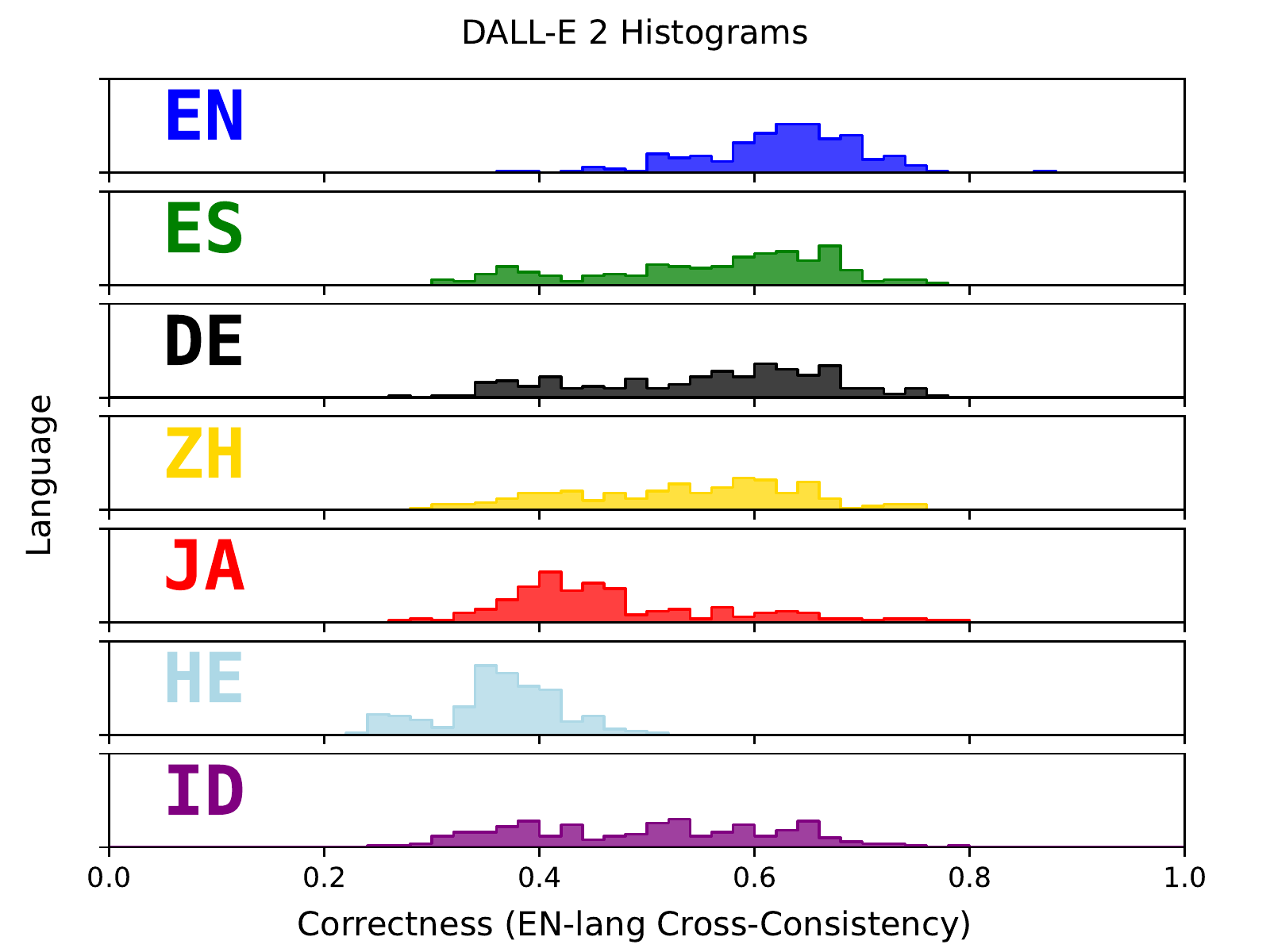}
\end{minipage}%
\begin{minipage}{.33\linewidth}
\centering
\includegraphics[width=1\linewidth]{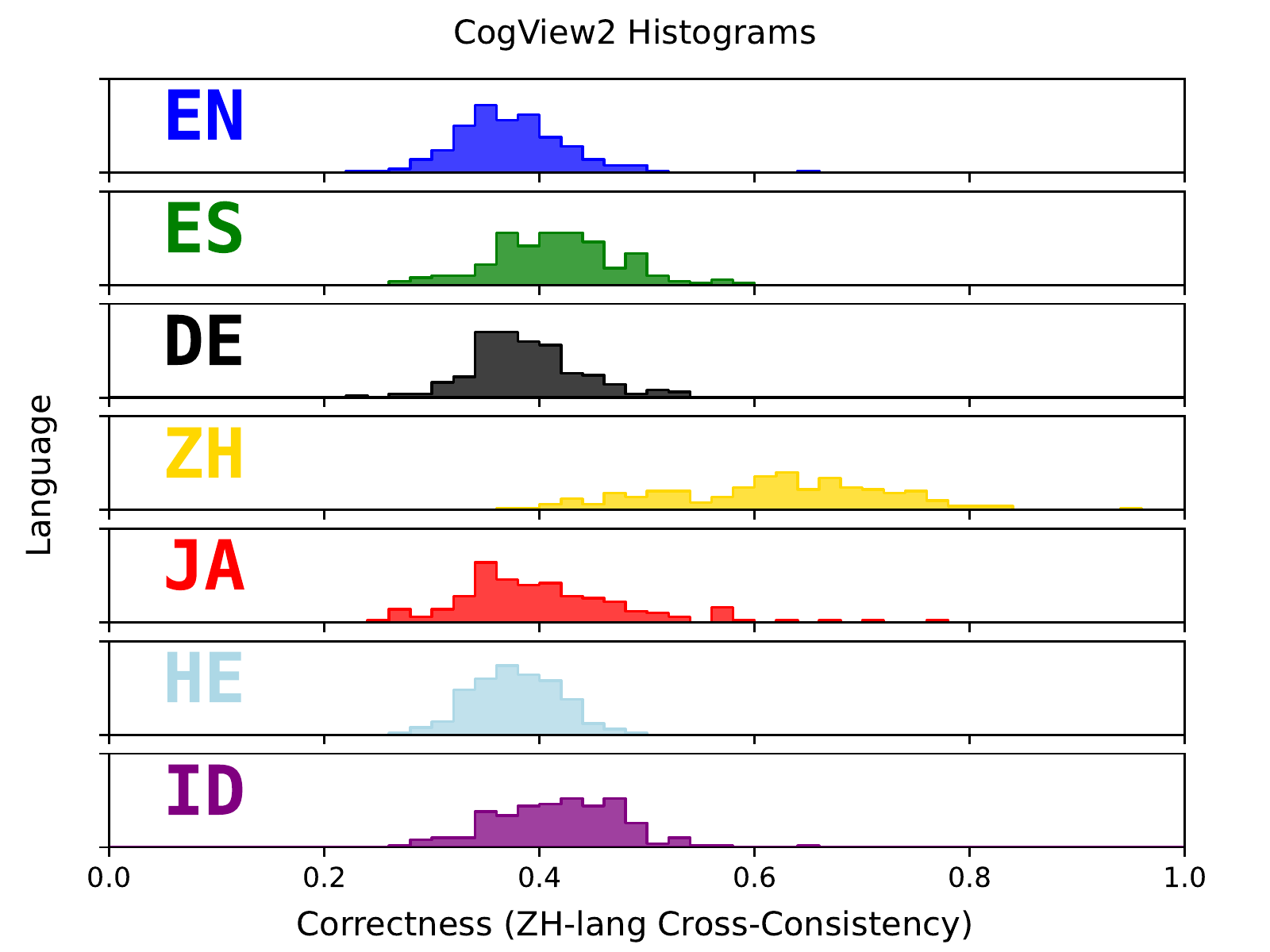}
\end{minipage}%

\begin{minipage}{.33\linewidth}
\centering
\includegraphics[width=1\linewidth]{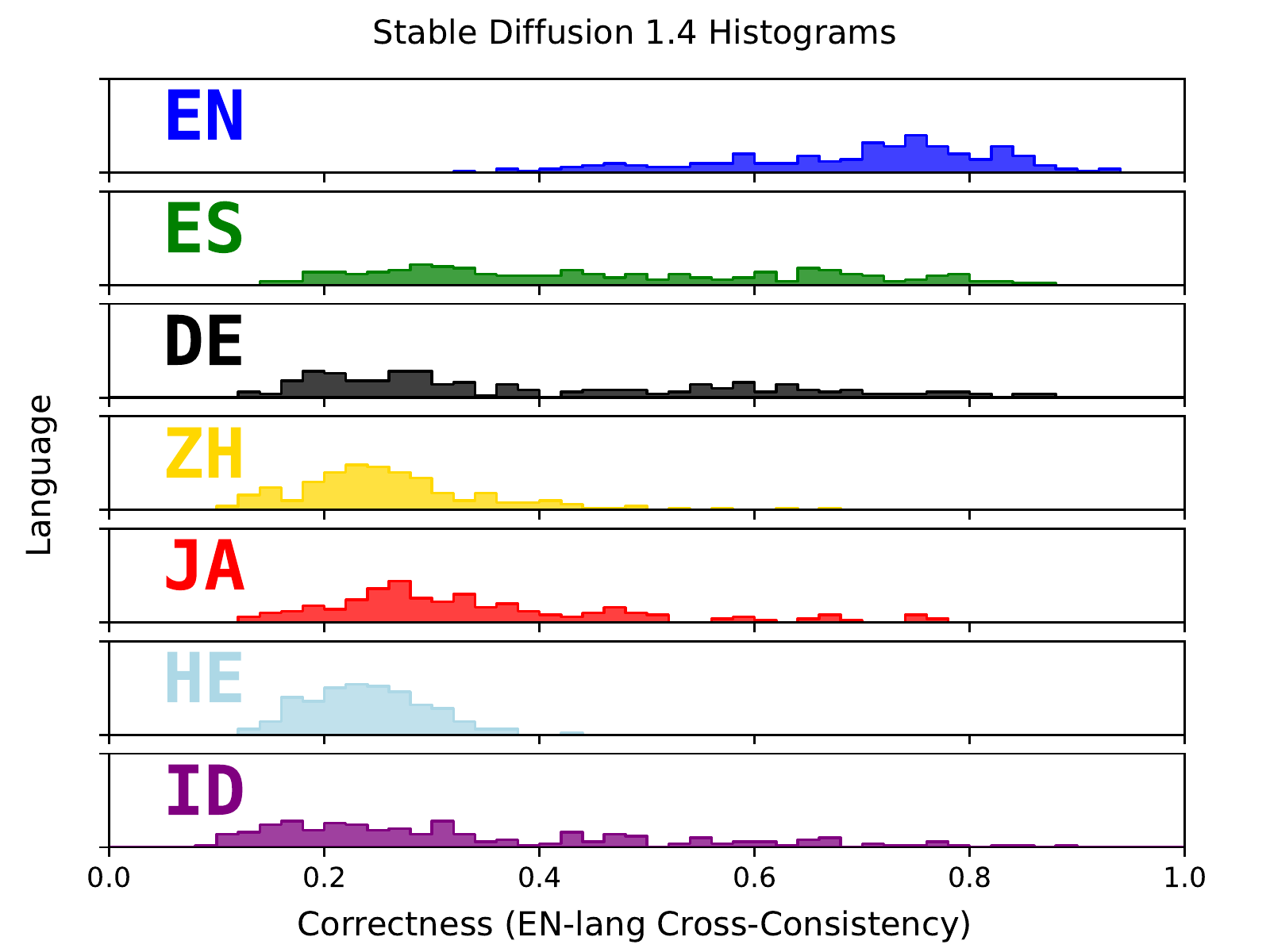}
\end{minipage}%
\begin{minipage}{.33\linewidth}
\centering
\includegraphics[width=1\linewidth]{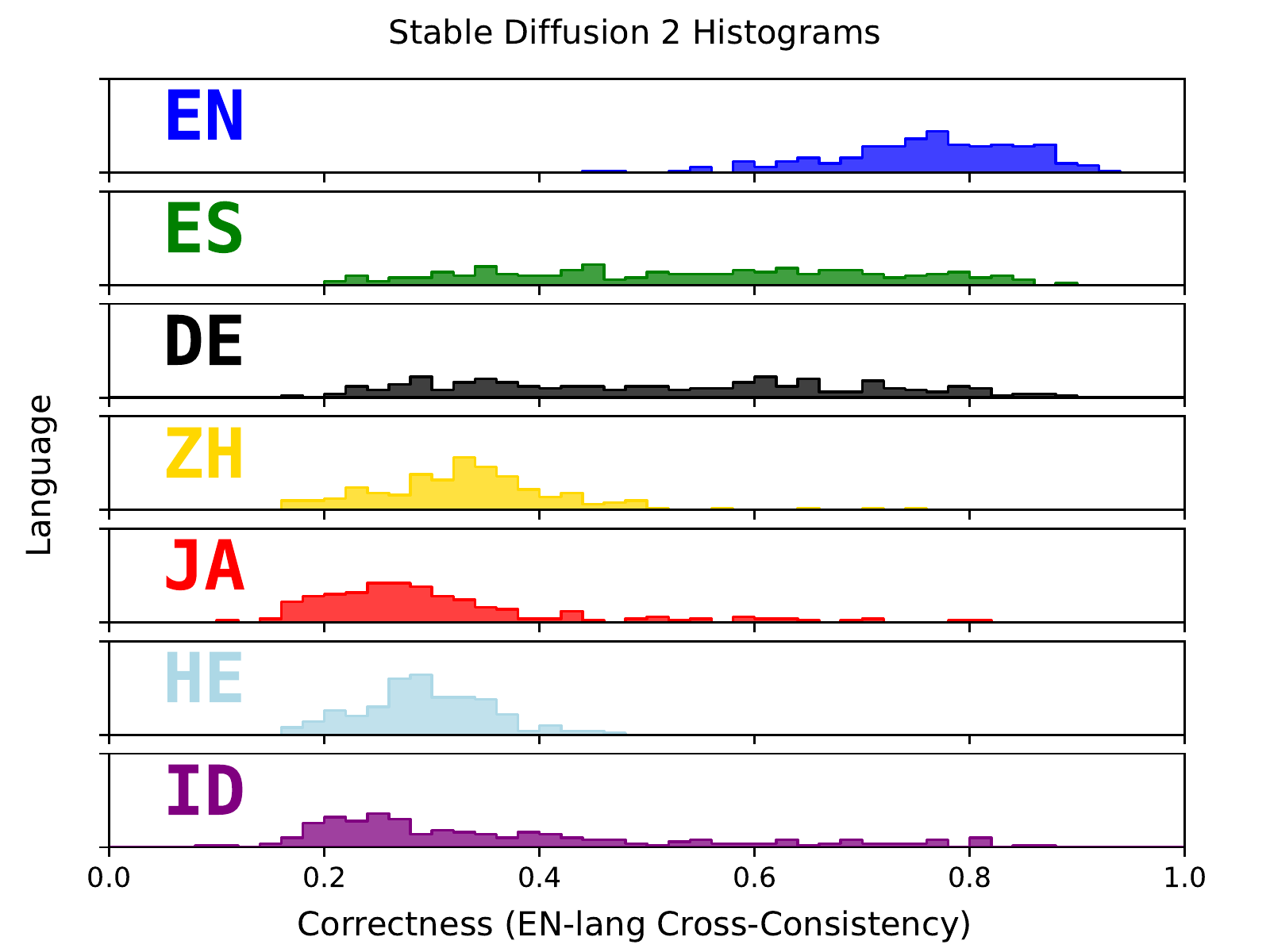}
\end{minipage}%
\begin{minipage}{.33\linewidth}
\centering
\includegraphics[width=1\linewidth]{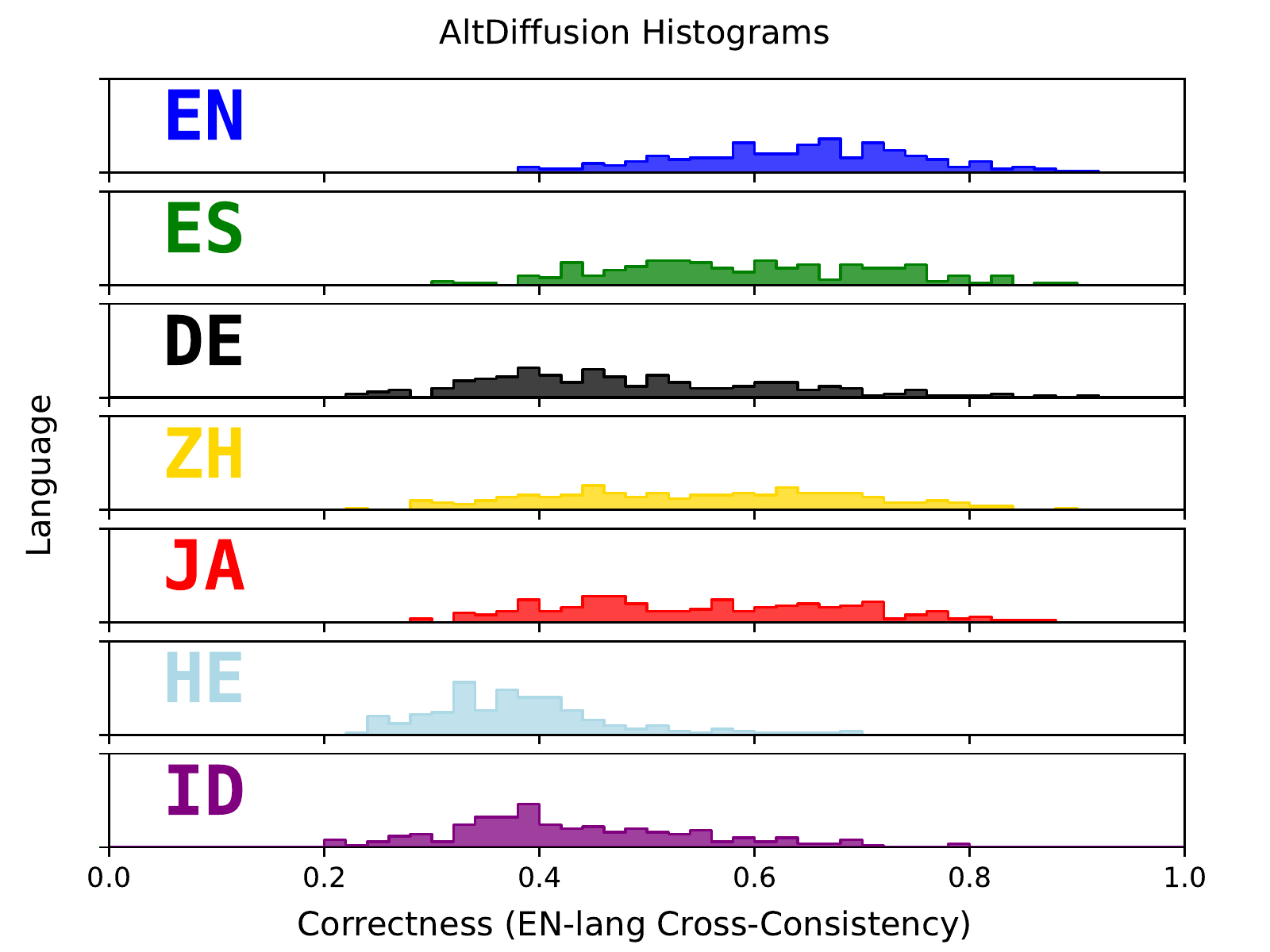}
\end{minipage}%

\caption{Histograms of the distribution of {\color{covBlue}\textbf{correctness}} cross-consistency (Xc) for each test language for six assessed models. Rightward probability mass reflects better conceptual coverage.}
\label{fig:histcov}
\vspace{-5pt}
\end{figure*}

\subsection{Translation-aligned concept set collection}\label{subsec:nounset}

We automatically produce an aligned multilingual concept list, where meaning, colloquial usage, and connotations are preserved as well as possible. 
We identify \textit{tangible nouns} describing physical objects, animals, people, and natural phenomena as a class of concepts that are both straightforward to evaluate and tend toward relative ubiquity in presence as words across languages and cultures, to the extent this ubiquity can be ensured through automated means.

Automated production is desirable for this task, as it enables new languages to be easily added in the future. 
To minimize translation errors we utilize both a large knowledge graph of terminology and an ensemble of commercial machine translation systems to produce an aligned concept list (\autoref{fig:aligneddict}).  
Full details for our translation pipeline, as well as the full concept list, are in \autoref{subsec:translation}. %

\subsection{Making minimal eliciting prompts}\label{subsec:minelicit} As discussed in \autoref{subsec:formulation}, an ideal prompt template would enforce stylistic consistency in the generated outputs without introducing biases that interfere with the demonstration of concept possession. Following \citet{bianchi2022easily} we build simple prompts of the form, ``a photograph of \_\_\_\_\_'', which we manually translate into target languages. This simple template-filling approach will introduce grammatical errors for some languages. We briefly investigate if this matters in \autoref{subsec:grammareffects}.

\begin{figure*}[t!]
\begin{minipage}{.25\linewidth}
\centering
\includegraphics[width=1\linewidth]{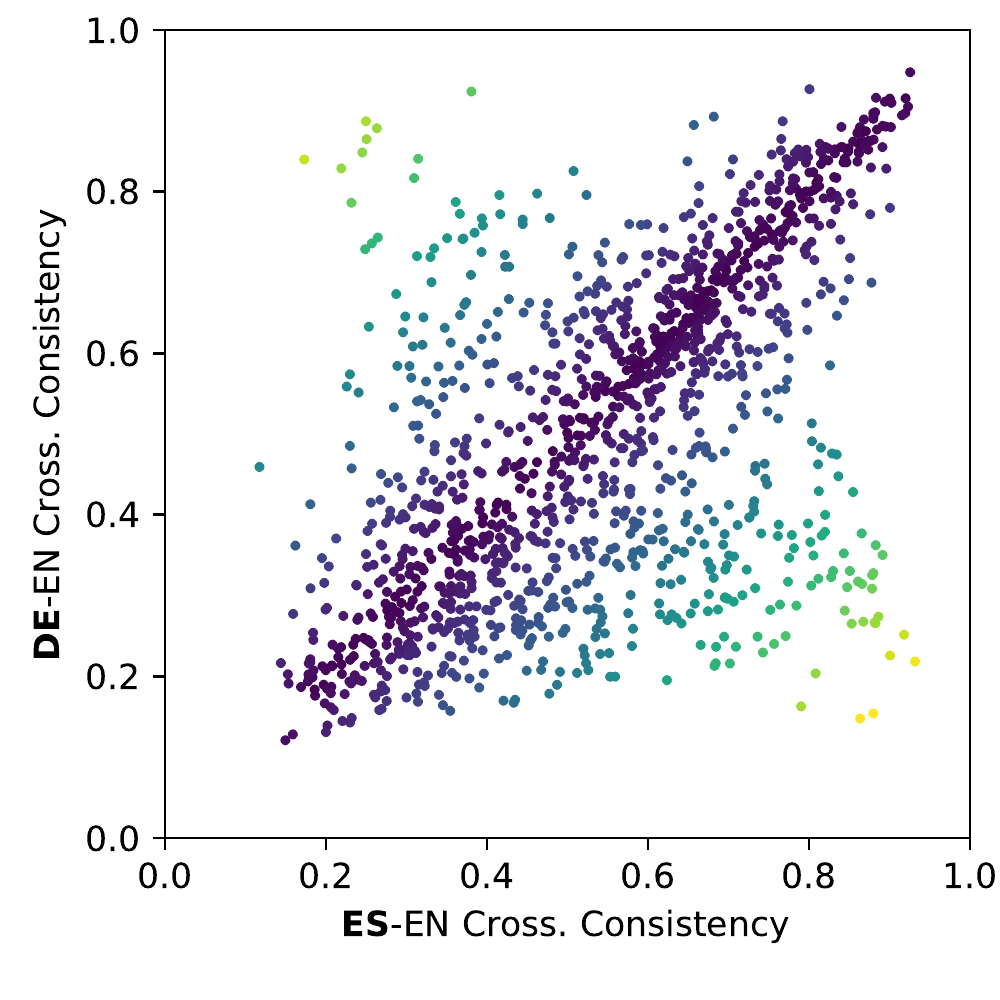}
\end{minipage}%
\begin{minipage}{.25\linewidth}
\centering
\includegraphics[width=1\linewidth]{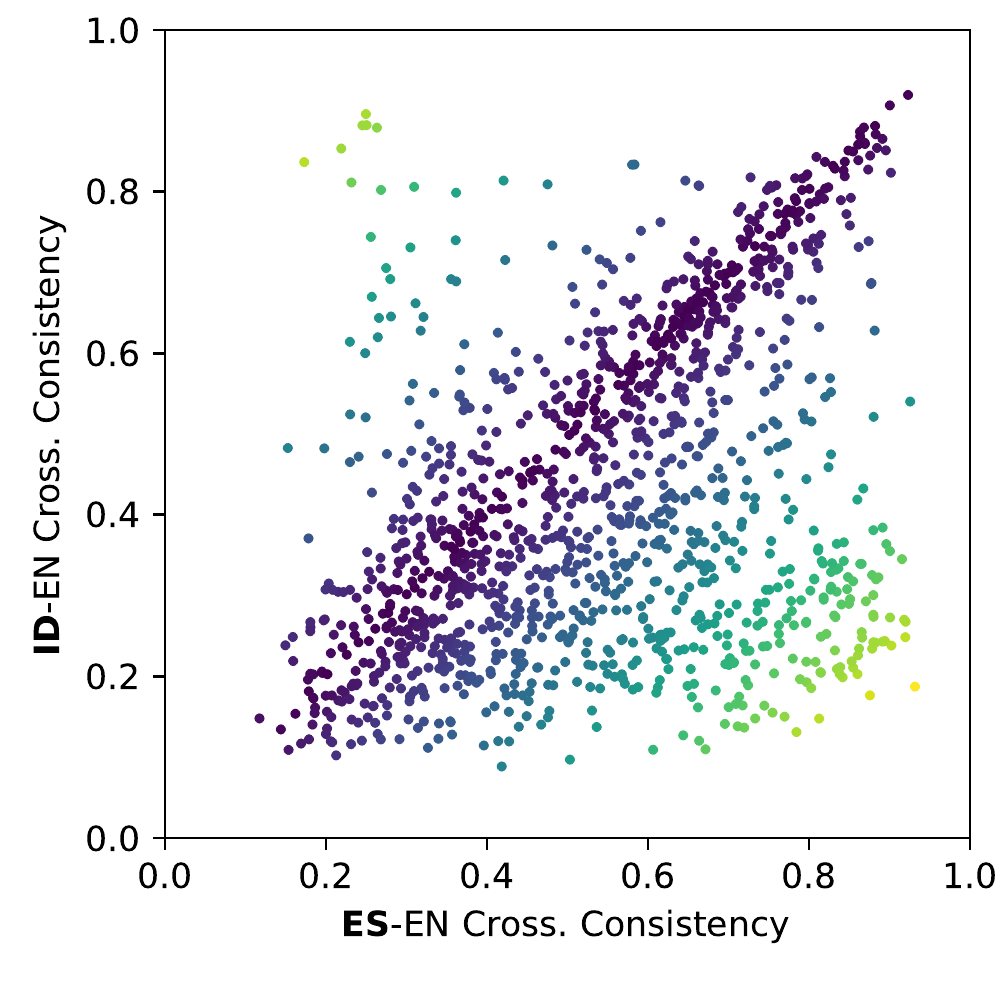}
\end{minipage}%
\begin{minipage}{.25\linewidth}
\centering
\includegraphics[width=1\linewidth]{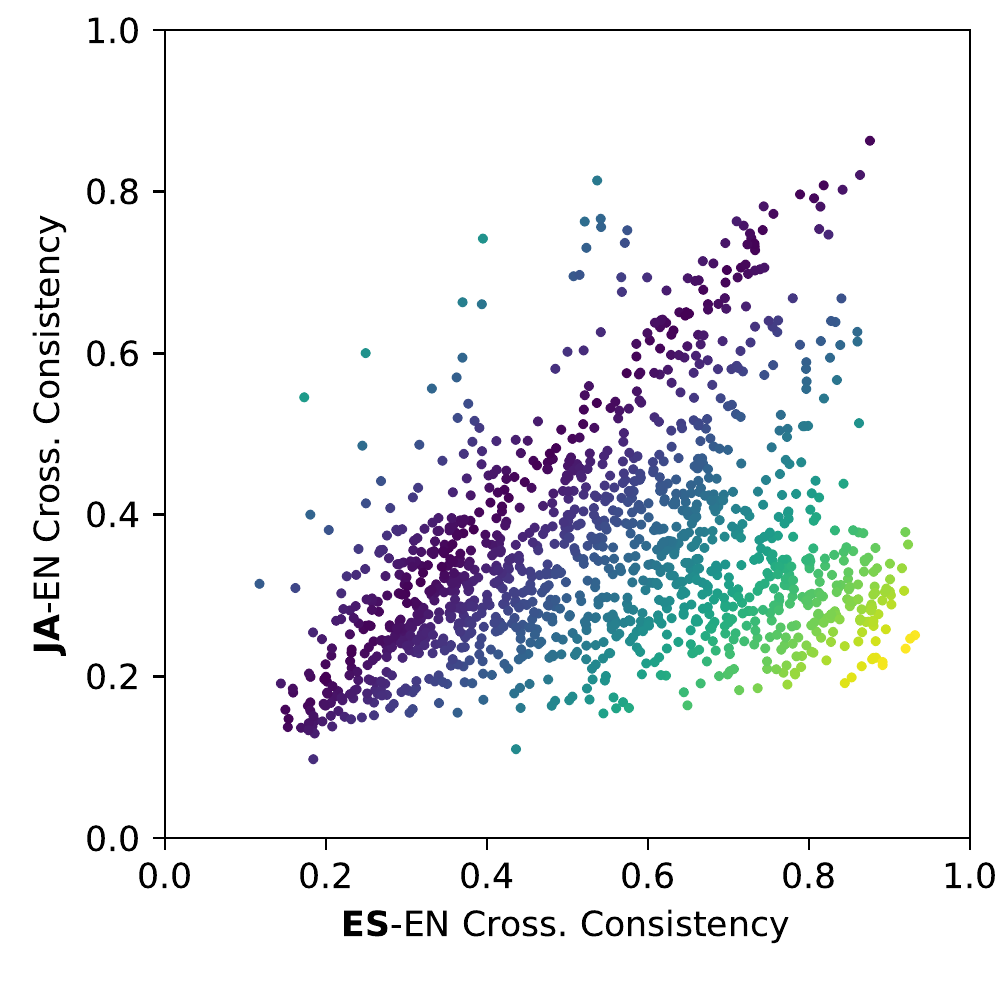}
\end{minipage}%
\begin{minipage}{.25\linewidth}
\centering
\includegraphics[width=1\linewidth]{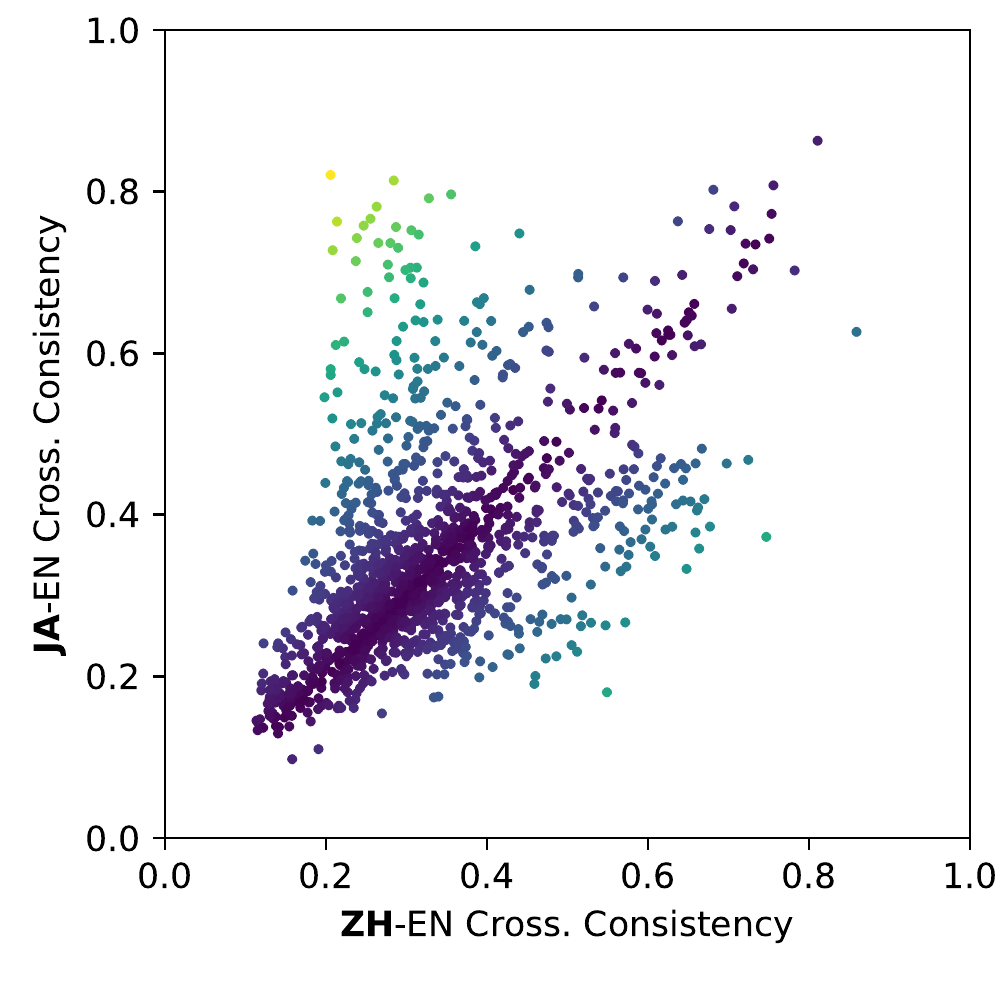}
\end{minipage}%

\caption{The {\color{covBlue}\textbf{correctness}} score for every (concept, model) pair for (right to left) ES vs DE, ES vs ID, ES vs JA, and ES vs JA. Languages sharing scripts (ES/DE/ID and JA/ZH) are more correlated than those that don't (ES/JA).}
\label{fig:scatter}
\vspace{-5pt}
\end{figure*}

\begin{figure*}[t!]
    \centering
    \includegraphics[width=1\linewidth]{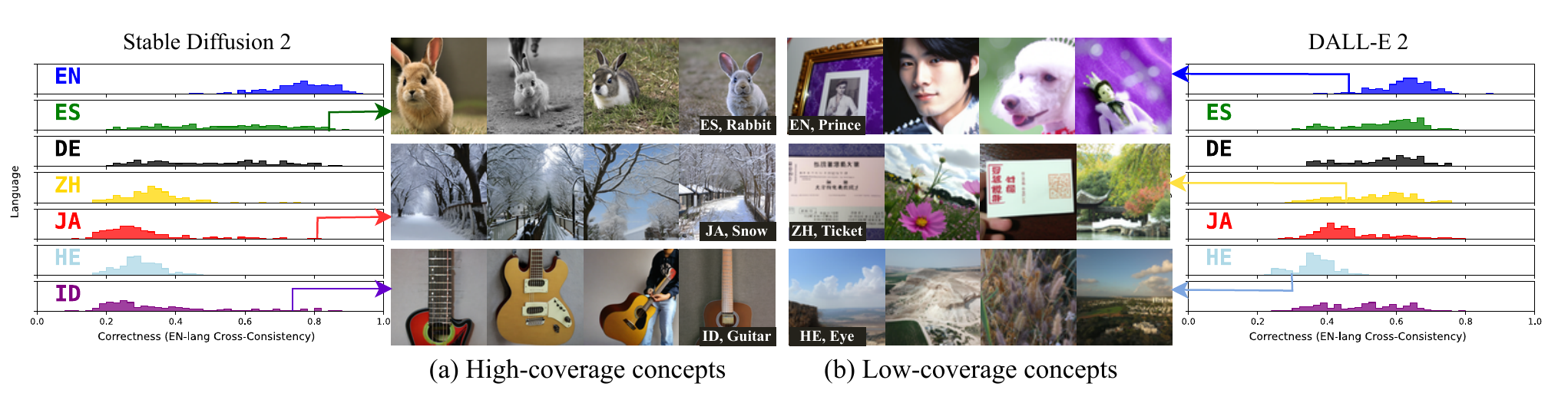}
    \vspace{-25pt}
    \caption{We automatically identify (a) high-coverage concepts in Stable Diffusion 2 (ES, rabbit), (JA, snow), (ID, guitar) and (b) low-coverage concepts in DALL-E 2 (EN, prince), (ZH, ticket), (HE, eye) using {\color{covBlue}\textbf{correctness}} Xc. %
    }
    \label{fig:bestworst-hist}
\end{figure*}

\subsection{Applying the metrics for analysis}

We assess Dt, Sc, Xc, and Wc for each (concept, language) pair for each model. Using these we compare models and assess the validity of conceptual coverage as a proxy for generalization.

\section{Findings}

\autoref{fig:histcov} shows histograms for the distributions of the {\color{covBlue}cross-consistency correctness} proxy score Xc for each concept, relative to the training language (either English or Chinese) for DALL-E Mini, DALL-E 2, CogView 2, Stable Diffusion 1.4, Stable Diffusion 2, and AltDiffusion across the seven test languages. This plot clearly depicts that for the primarily English-trained models (DALL-E Mini, Stable Diffusion 1.4, Stable Diffusion 2), English-language performance (a high-mean distribution of high-EN-EN consistency concepts) is considerably better than the other languages. Similarly, for CogView2, trained on Chinese, the Chinese distribution of ZH-ZH scores is considerably better than the others, which do equally bad.

DALL-E 2 recieved open-ended multilingual training, and exhibits more consistent acceptable performance across the European and East Asian languages being tested. AltDiffusion, which has had its CLIP text encoder contrastively trained against multilingual representations on 9 languages (including ES, DE, ZH, and JA) exhibits higher performance on its training languages than its non-training languages (HE and ID).

Correctness distributions for Spanish, German, and Indonesian look roughly similar (in terms of mean and variance) for all models but AltDiffusion. This is particularly interesting because they are the three non-English languages that also use the Latin alphabet. \autoref{fig:scatter} compares the {\color{covBlue}\textbf{correctness}} Xc score for every concept, in every model, across pairs of languages that fully or partially share scripts (ES, DE, ID), (ZH, JA) and two languages that don't (JA, ES). Across pairs of languages that share scripts, there is a high correlation between possession of a given concept in one language and the other. A consistent trend across all models was poor performance on Hebrew, which is both considerably lower-resource compared to the other six test languages, and uses its own unique writing system.

\subsection{Correctness feature captures possession}

\autoref{fig:bestworst-hist} shows how choosing samples of an image generated by a model, elicited by a high- or low-{\color{covBlue}\textbf{correctness}} score naturally reveals in which languages which concepts are possessed (e.g., Stable Diffusion 2 possesses ES:rabbit, JA:snow, and ID:guitar.) When a model possesses a concept, the outputted images are often visually similar with the tangible concept set in similar scenarios.

\begin{table*}[t!]
\small
    \centering
    \begin{tabular}{lrrrrrrrrrrrrrrrr}
        \toprule
         & \multicolumn{2}{c}{EN} & \multicolumn{2}{c}{ES} & \multicolumn{2}{c}{DE} & \multicolumn{2}{c}{ZH} & \multicolumn{2}{c}{JA} & \multicolumn{2}{c}{HE} & \multicolumn{2}{c}{ID} & \multicolumn{2}{c}{Avg} \\
        Model & Xc & Wc & Xc & Wc & Xc & Wc & Xc & Wc & Xc & Wc & Xc & Wc & Xc & Wc & Xc & Wc \\
        \midrule
        \rowcolor{black!0!yellow!5}DALL-E Mega & \textbf{81} & \textbf{28} & \textbf{65} & 26 & \textbf{64} & \textbf{26} & 29 & 21 & 32 & 21 & 28 & 19 & \textbf{51} & \textbf{25} & 50 & \textbf{24} \\
        DALL-E Mini & 78 & 27 & 59 & 25 & 50 & 23 & 33 & 21 & 31 & 21 & 34 & \textbf{20} & 49 & 24 & 48 & 23 \\
        SD 1.1 & 69 & 26 & 52 & 23 & 46 & 22 & 32 & 19 & 37 & 21 & 28 & 17 & 39 & 21 & 43 & 21 \\
        SD 1.2 & 71 & 26 & 48 & 23 & 44 & 22 & 28 & 19 & 35 & 21 & 24 & 17 & 37 & 21 & 41 & 21 \\
        SD 1.4 & 69 & 26 & 46 & 23 & 40 & 22 & 26 & 20 & 34 & 21 & 24 & 17 & 34 & 21 & 39 & 21 \\
        SD 2 & 76 & 27 & 54 & 24 & 51 & 24 & 34 & 19 & 31 & 21 & 29 & 17 & 37 & 21 & 45 & 22 \\
        \rowcolor{black!0!red!5}CogView 2 & 37 & 20 & 42 & 20 & 39 & 20 & \textbf{62} & \textbf{25} & 40 & 21 & \textbf{38} & \textbf{\textbf{20}} & 42 & 20 & 43 & 21 \\
        DALL-E 2 & 61 & 27 & 55 & \textbf{27} & 54 & \textbf{26} & 44 & \textbf{25} & 42 & 22 & 36 & 19 & 42 & \textbf{25} & 48 & 24 \\
        \rowcolor{black!0!green!5}\vspace{2pt}\textbf{AltDiffusion m9} & 64 & 26 & 59 & 25 & 49 & 22 & 55 & \textbf{25} & \textbf{55} & \textbf{25} & \textbf{38} & \textbf{20} & 43 & 22 & \textbf{52} & 23  \\
         Avg  & 67 & 26 & 53 & 24 & 49 & 23 & 38 & 22 & 38 & 21 & 31 & 18 & 42 & 22 \\
 \bottomrule
    \end{tabular}
    \vspace{-5pt}
    \caption{{\color{covBlue}\textbf{Correctness}} scores (Xc and Wc) averaged for all concepts within a column language for all models. Note that Xc for \colorbox{black!0!red!5}{CogView2} is relative to ZH rather than EN. \colorbox{black!0!green!5}{AltDiffusion} performs best in terms of total average Xc, and number of Xc or Wc column ``wins.'' \colorbox{black!0!yellow!5}{DALL-E Mega} performs best on Latin languages and average Wc.}
    \vspace{-2ex}
    \label{tab:crosssim}
\end{table*}

\subsection{Types of concept non-possession}

A model \textit{not} possessing a concept can manifest in a few different scenarios depicted in \autoref{fig:bestworst-hist} (b). DALL-E 2 doesn't possess ``prince'' in English because it outputs a variety of different images, including human portrait photos, and pictures of photos, toys, and dogs. The existence of these \textit{non-specific} error cases reflects the imperfect nature of our automated concept collection pipeline. Removing these poorly specified concepts is one way we plan to improve \cococrola~ in future releases.

A second type of possession failure we observe, we dub \textit{specific collisions}. For example, \hyperref[fig:teaser]{Figure 1} and \autoref{fig:incite} show JA collisions for the DALL-E mini/mega family. Both models consistently generate images of humans for ``dog'' but pictures of green landscape scenes for ``airplane.'' While these generated concepts are incorrect, they represent an incorrect mapping to a different concept rather than a mere lack of conceptual possession. We also observe cases where specific collisions only occur part of the time, such as in the case of DALL-E 2 and ZH:ticket (\autoref{fig:bestworst-hist} (b)).

Finally, we observed cases of \textit{generic collisions}. For example, DALL-E 2 consistently generates images of desert or Mediterranean scenery when prompted with ``eye'' in Hebrew (\autoref{fig:bestworst-hist} (b)). This pattern shows up across a diverse set of models and prompts. \hyperref[fig:teaser]{Figure 1} shows how across ``dog,'' ``airplane,'' and ``face,'' DALL-E mega, Stable Diffusion 2, and DALL-E 2 seem to generate vaguely-Israel-looking outdoor scenes regardless of eliciting concept. This is probably reflective of a small sample-size bias in the training data. %

\subsection{Model comparison}

\autoref{tab:crosssim} shows the the use of correctness scores in the \cococrola~benchmark to compare the 9 models. As expected, given its multilingual training regimen, AltDiffusion m9 outperforms the other T2I models on average, and in terms of total wins. It is particularly strong relative to the other models in Japanese and Chinese (with the exception of the Chinese-only CogView 2, which is best on Chinese but worst on average overall for both Xc and Wc).

However, despite the strong average performance of AltDiffusion, there's a lot of room for improvement. For example, its improvements in terms of JA and HE performance come at a cost of significantly reduced EN and DE performance relative to Stable Diffusion 2, its initialization checkpoint. The \cococrola~benchmark can guide future work in adapting T2I models to further multilinguality without losing conceptual coverage on source languages.

\begin{figure}[h!]

    \centering
    \includegraphics[width=0.9\linewidth]{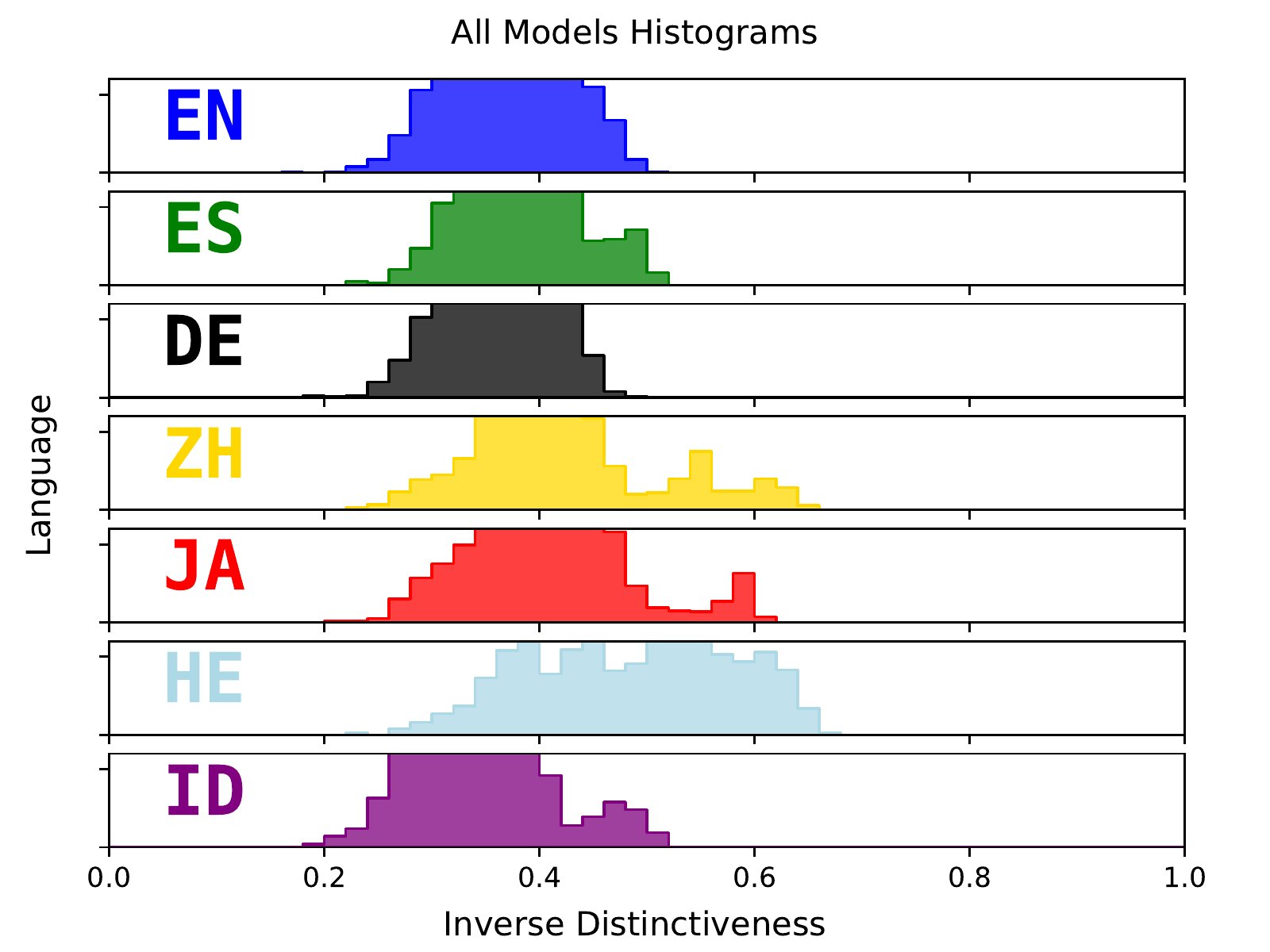}
    \caption{Histograms of the \textbf{\color{disYellow}inverse distinctiveness} scores for all models and all concepts. A model-by-model breakdown is presented in \autoref{fig:disfull}.
    }
    \label{fig:histspecall}
    \vspace{-11pt}
\end{figure}

\newcommand{\timg}[1]{\includegraphics[width=0.25\linewidth]{img/storyprop/#1}}

\newcommand{\specialcell}[1]{%
  \begin{tabular}[c]{@{}c@{}}#1\end{tabular}}

\newcommand{\imgf}[4]{\specialcell{\timg{#1}\timg{#2}\vspace{-3pt}\\\timg{#3}\timg{#4}}\hspace{-5pt}}

\newcommand{\cellresults}[1]{
\begin{tabular}{lrrr}
Concept & Xc & Wc &  \\
\hline
#1
\end{tabular}}

\newcommand{\posy}{\textcolor{green!80!black}{\ding{52}}}
\newcommand{\posn}{\textcolor{red}{\ding{55}}}

\begin{figure*}[t!]
    \centering

    \begin{subfigure}{0.49\textwidth}
    \resizebox{1\linewidth}{!}{
    \begin{tabular}{p{2pt}ccc}
        \toprule
          & \large \texttt{EN} & \large \texttt{ES} & \large \texttt{JA} \\
        \midrule
      
    \rotatebox{90}{\hspace{-45pt}\large\textbf{\texttt{DALL-E Mega}}} & \imgf{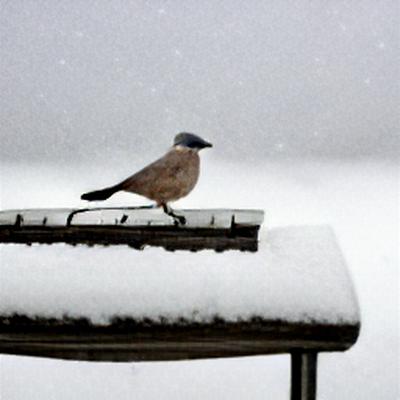}{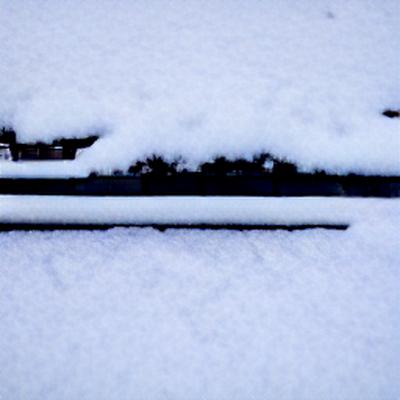}{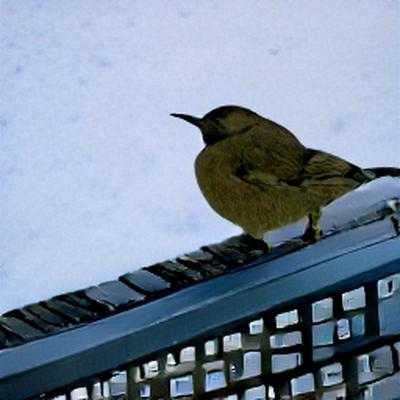}{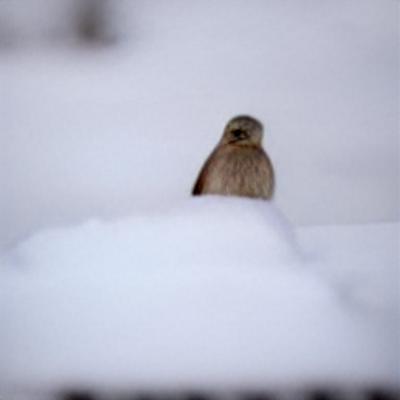} & \imgf{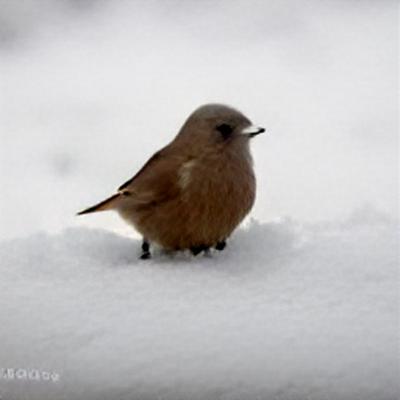}{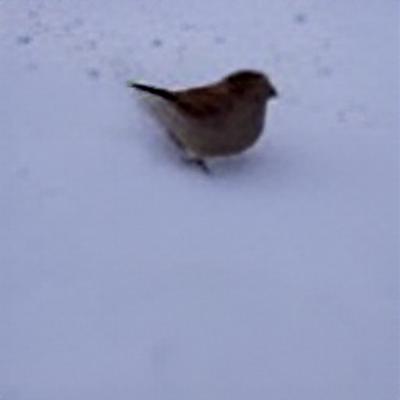}{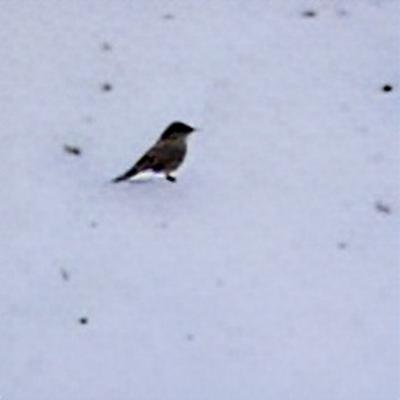}{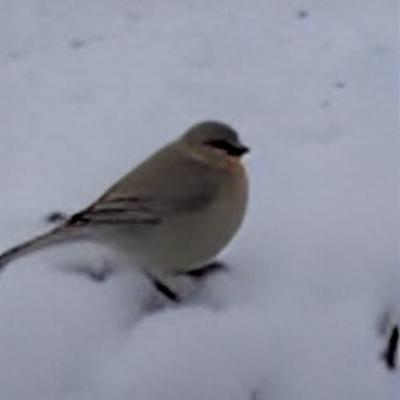} & \imgf{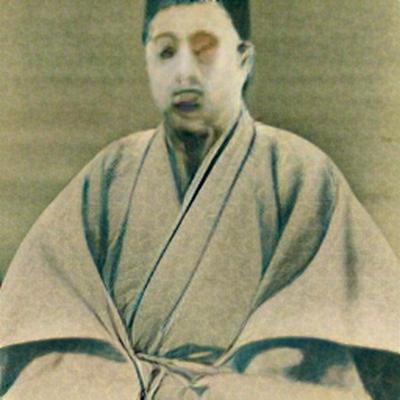}{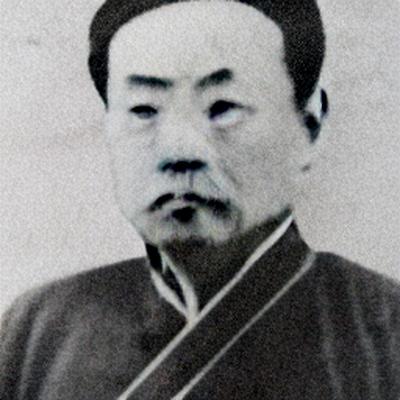}{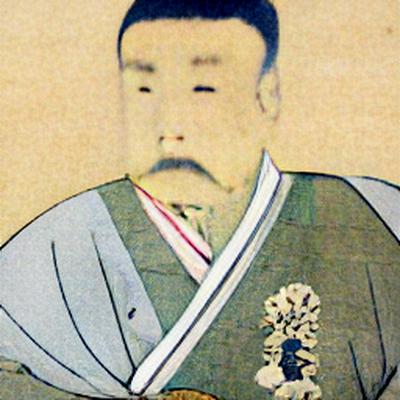}{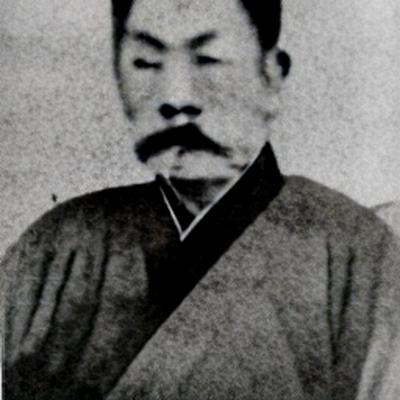}\\
      & \cellresults{\texttt{bird} & 0.741 & 27 & \posy \\ \texttt{\footnotesize keyboard} & 0.824 & 28 & \posy \\ \texttt{snow} & 0.787 & 27 & \posy \\ } & \cellresults{\texttt{bird} & 0.739 & 27 & \posy \\ \texttt{\footnotesize keyboard} & 0.801 & 29 & \posy \\ \texttt{snow} & 0.723 & 26 & \posy \\ } & \cellresults{\texttt{bird} & 0.704 & 26 & \posy \\ \texttt{\footnotesize keyboard} & 0.346 & 18 & \posn \\ \texttt{snow} & 0.404 & 19 & \posn \\ }\\
    \rotatebox{90}{\hspace{-45pt}\large\textbf{\texttt{AltDiffusion}}} & \imgf{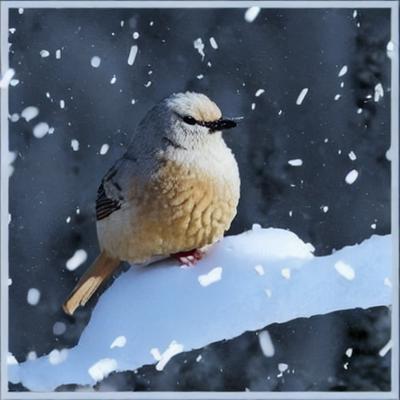}{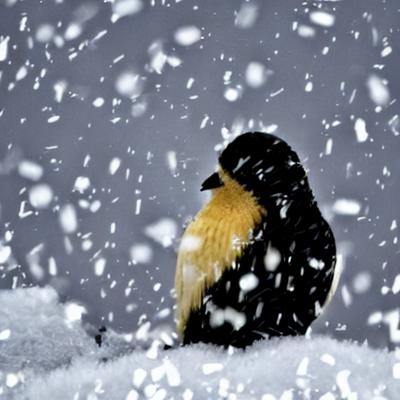}{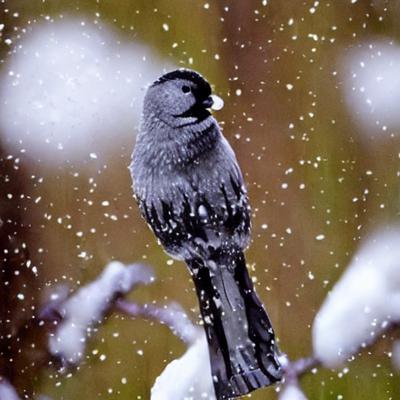}{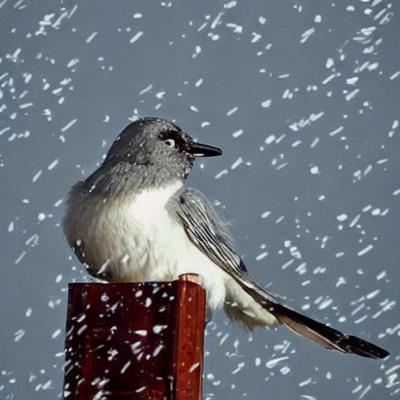} & \imgf{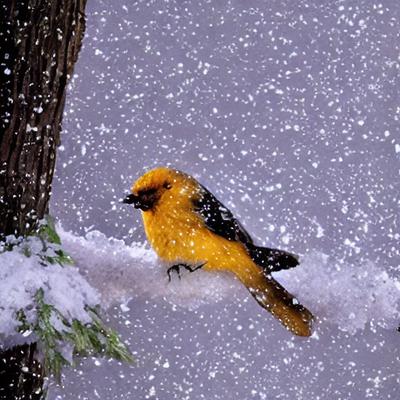}{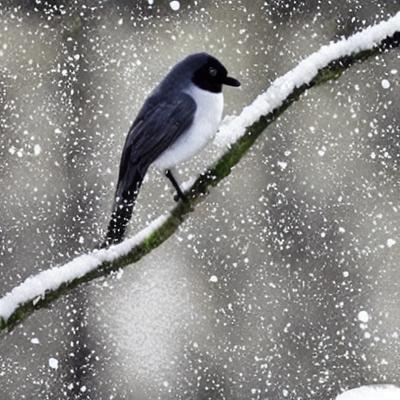}{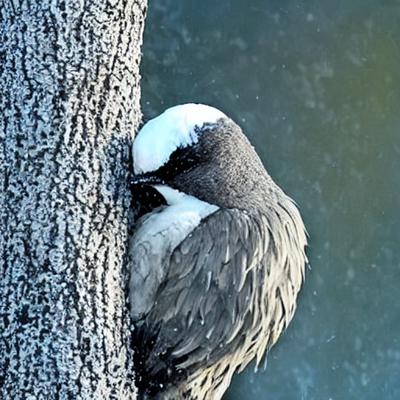}{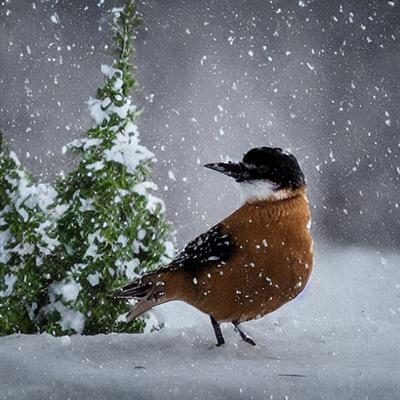} & \imgf{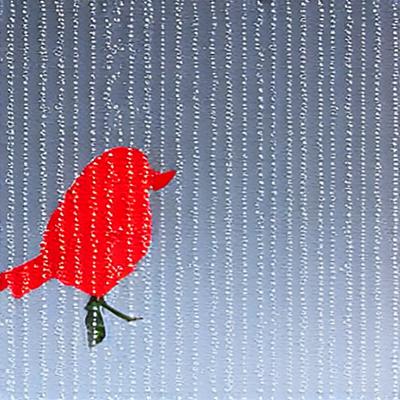}{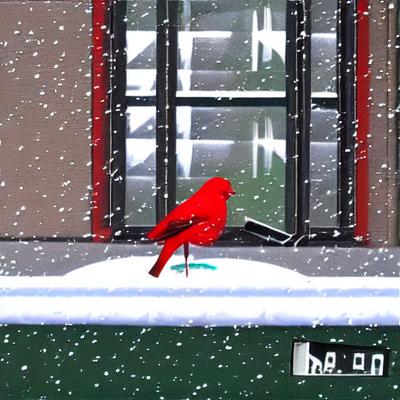}{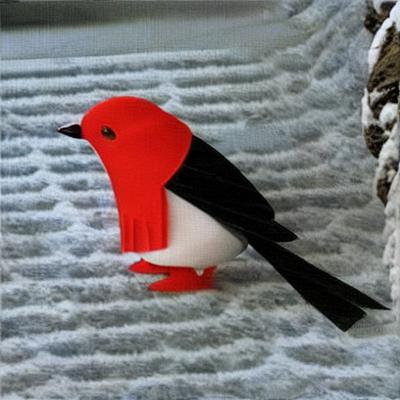}{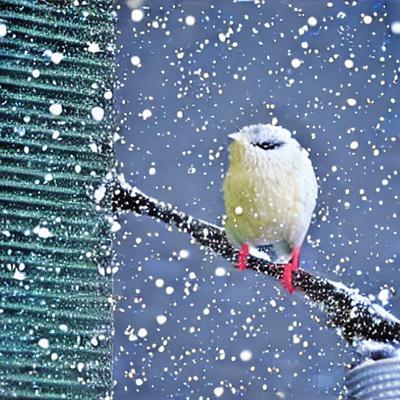}\\
      & \cellresults{\texttt{bird} & 0.655 & 26 & \posy \\ \texttt{\footnotesize keyboard} & 0.491 & 27 & \posy \\ \texttt{snow} & 0.759 & 26 & \posy \\ } & \cellresults{\texttt{bird} & 0.646 & 26 & \posy \\ \texttt{\footnotesize keyboard} & 0.489 & 26 & \posy \\ \texttt{snow} & 0.704 & 25 & \posy \\ } & \cellresults{\texttt{bird} & 0.655 & 26 & \posy \\ \texttt{\footnotesize keyboard} & 0.462 & 24 & \posn \\ \texttt{snow} & 0.671 & 25 & \posy \\ }\\
    \rotatebox{90}{\hspace{-45pt}\large\textbf{\texttt{Stable Diffusion 2}}} & \imgf{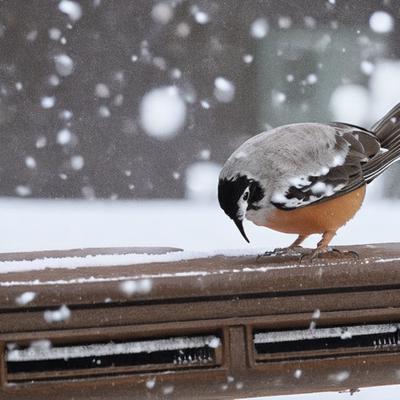}{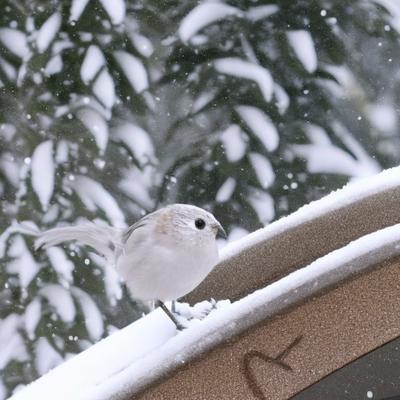}{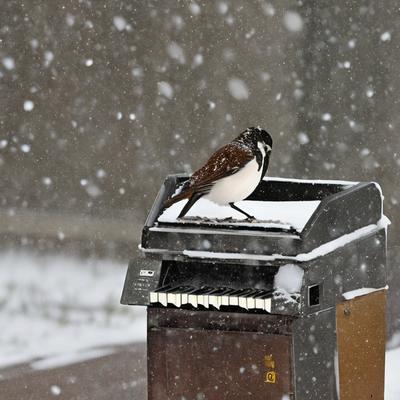}{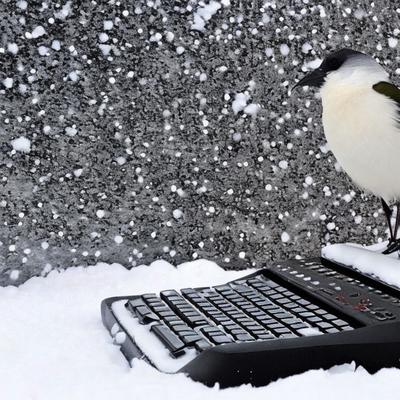} & \imgf{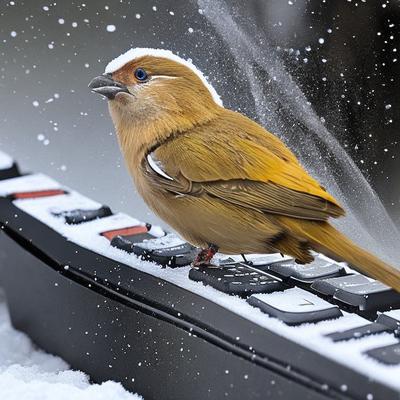}{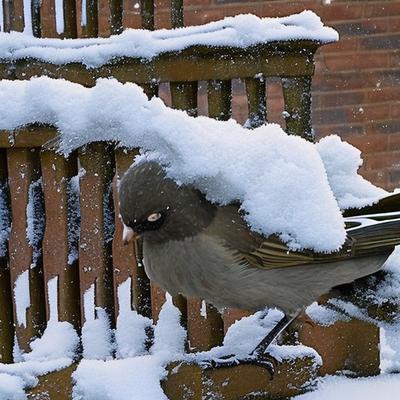}{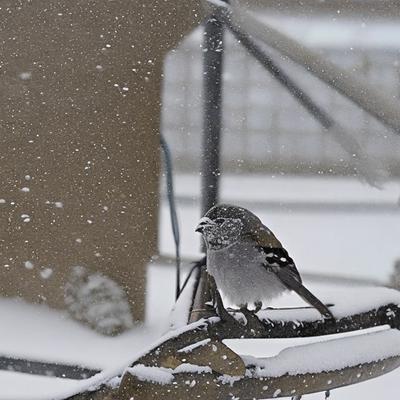}{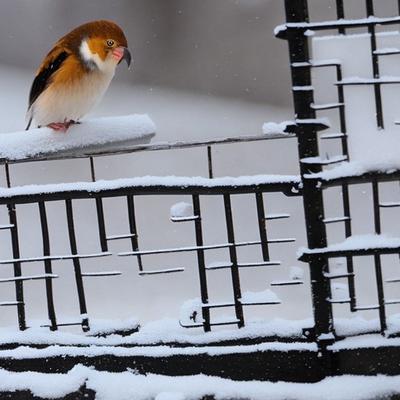} & \imgf{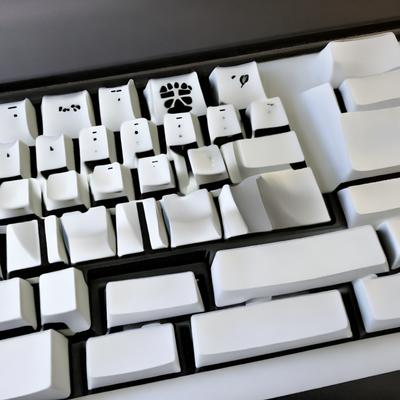}{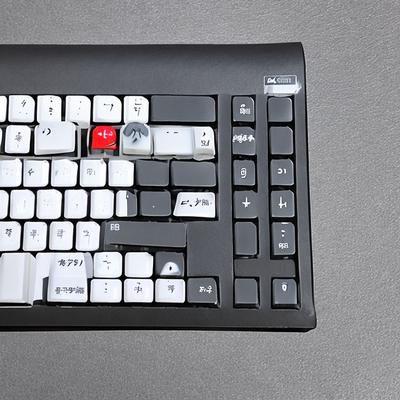}{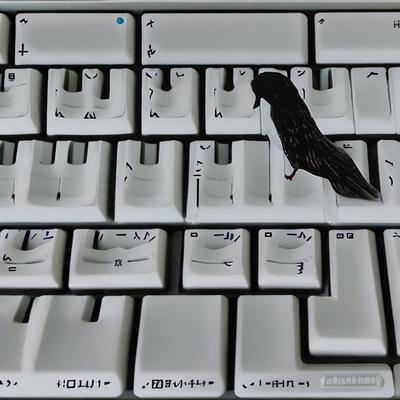}{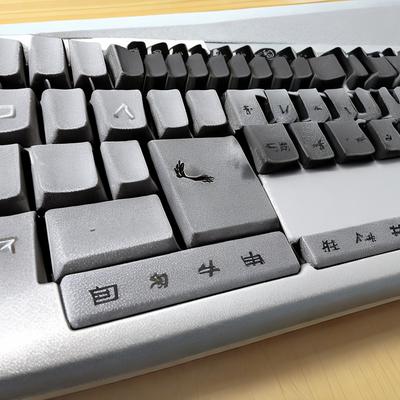}\\
      & \cellresults{\texttt{bird} & 0.726 & 27 & \posy \\ \texttt{\footnotesize keyboard} & 0.837 & 29 & \posy \\ \texttt{snow} & 0.846 & 26 & \posy \\ } & \cellresults{\texttt{bird} & 0.697 & 26 & \posy \\ \texttt{\footnotesize keyboard} & 0.789 & 29 & \posy \\ \texttt{snow} & 0.818 & 26 & \posy \\ } & \cellresults{\texttt{bird} & 0.655 & 26 & \posy \\ \texttt{\footnotesize keyboard} & 0.797 & 29 & \posy \\ \texttt{snow} & 0.808 & 26 & \posy \\ }\\

    \bottomrule
    \end{tabular}}

        \caption{``a bird using a keyboard in the snow''}
    \label{fig:snowkeyboardbird}
\end{subfigure}\begin{subfigure}{0.49\textwidth}
    \resizebox{1\linewidth}{!}{
    \begin{tabular}{p{2pt}ccc}
        \toprule
          & \large \texttt{EN} & \large \texttt{ES} & \large \texttt{JA} \\
        \midrule

    \rotatebox{90}{\hspace{-45pt}\large\textbf{\texttt{DALL-E Mega}}} & \imgf{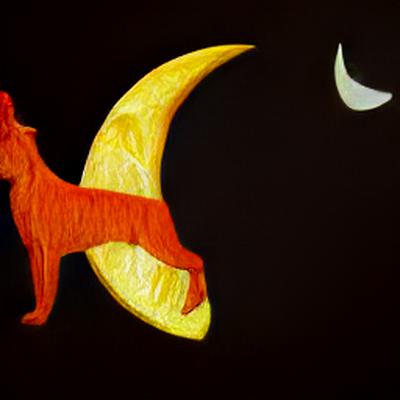}{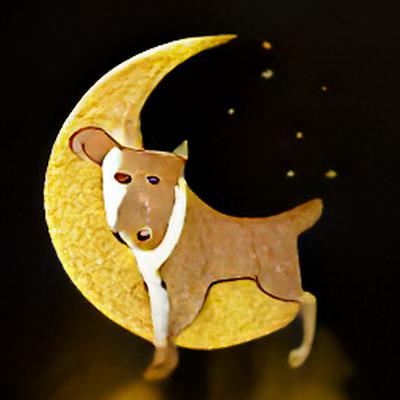}{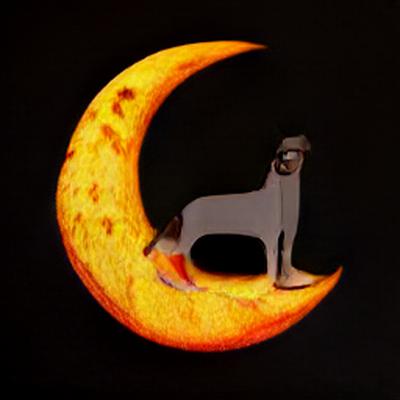}{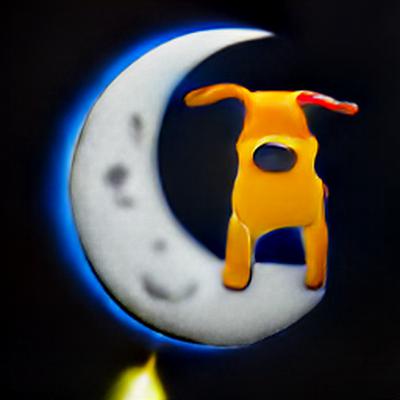} & \imgf{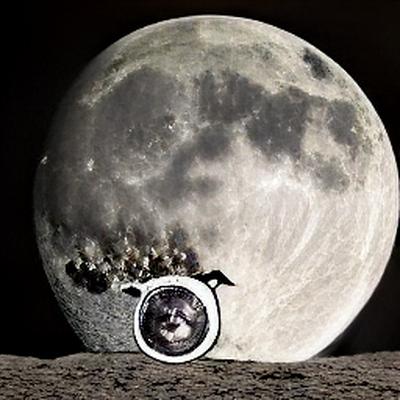}{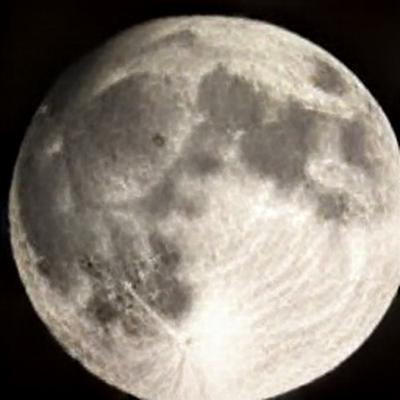}{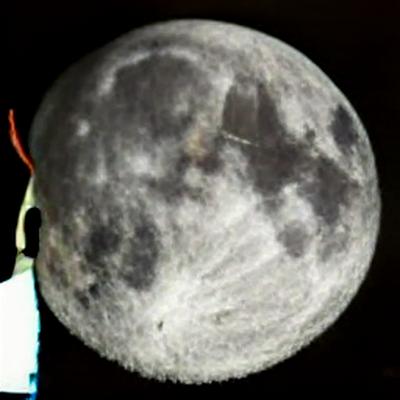}{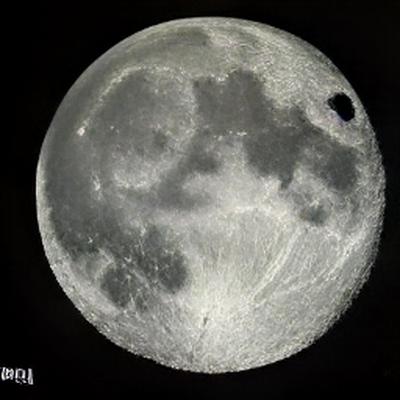} & \imgf{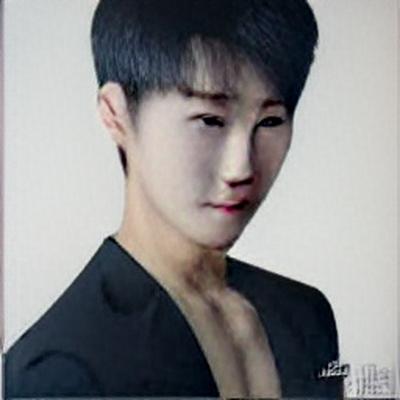}{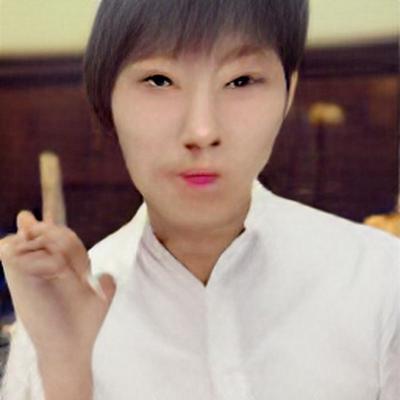}{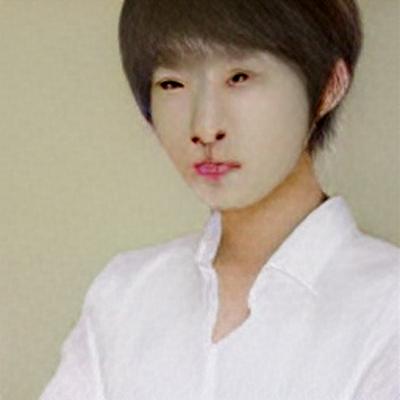}{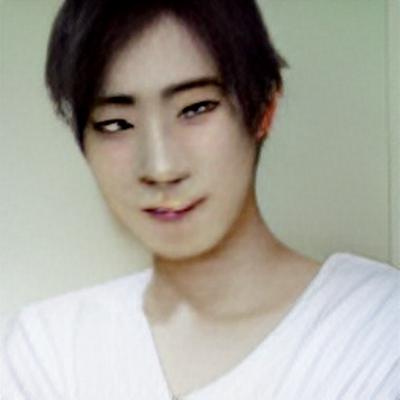}\\
      & \cellresults{\texttt{dog} & 0.746 & 26 & \posy \\ \texttt{fire} & 0.938 & 27 & \posy \\ \texttt{moon} & 0.868 & 29 & \posy \\ } & \cellresults{\texttt{dog} & 0.712 & 27 & \posy \\ \texttt{fire} & 0.926 & 27 & \posy \\ \texttt{moon} & 0.864 & 28 & \posy \\ } & \cellresults{\texttt{dog} & 0.298 & 19 & \posn \\ \texttt{fire} & 0.247 & 19 & \posn \\ \texttt{moon} & 0.269 & 23 & \posn \\ }\\
    \rotatebox{90}{\hspace{-45pt}\large\textbf{\texttt{AltDiffusion}}} & \imgf{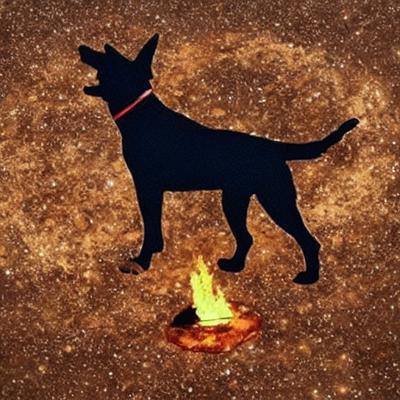}{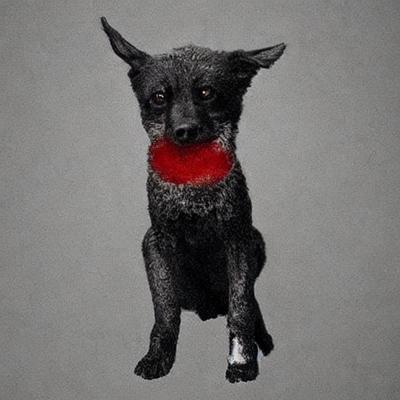}{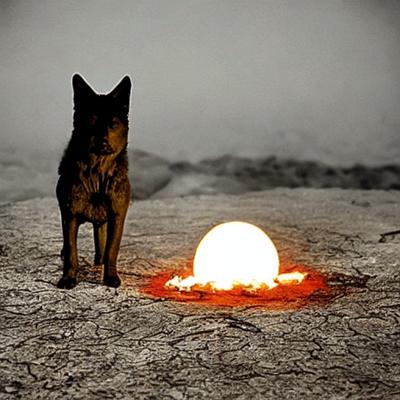}{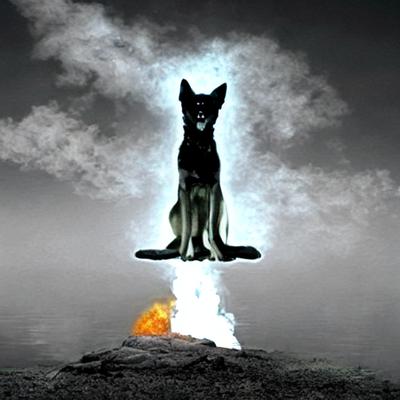} & \imgf{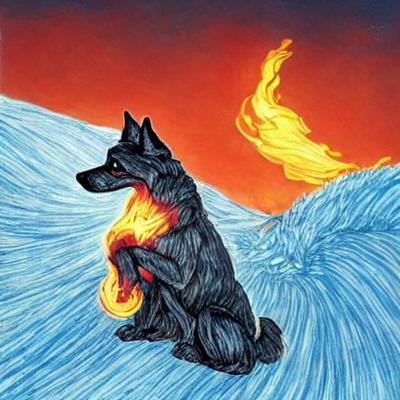}{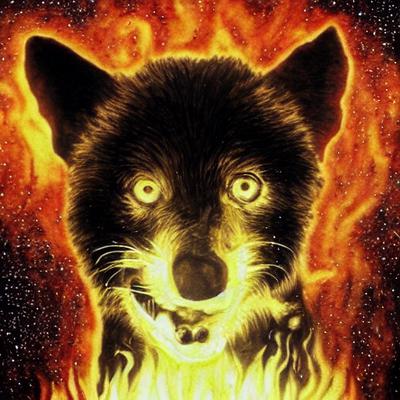}{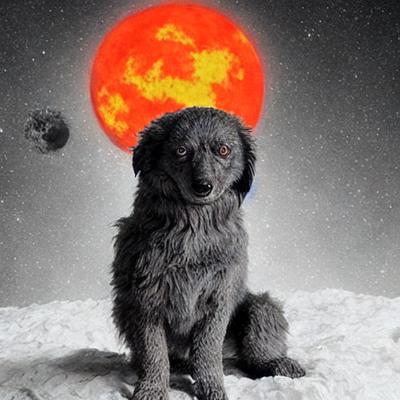}{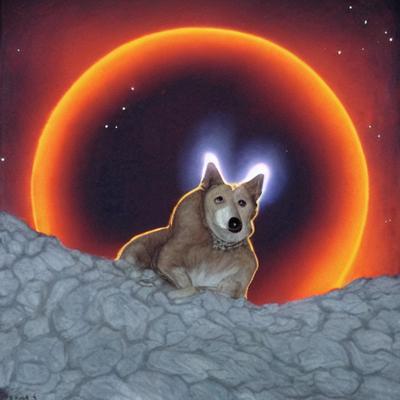} & \imgf{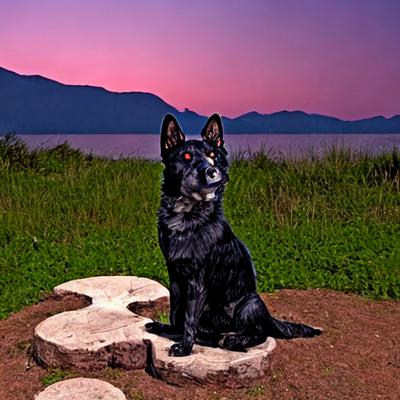}{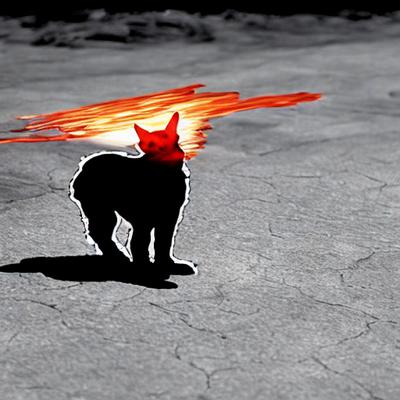}{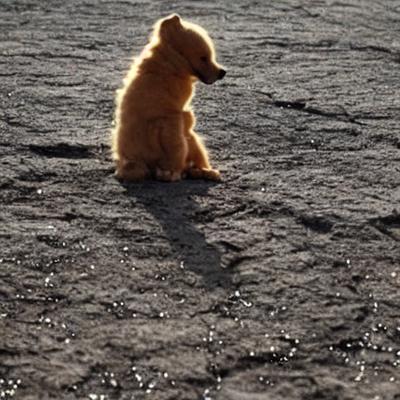}{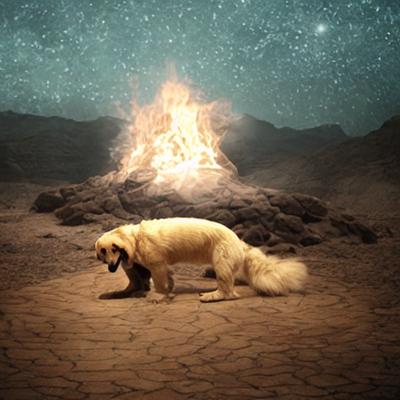}\\
      & \cellresults{\texttt{dog} & 0.702 & 26 & \posy \\ \texttt{fire} & 0.669 & 23 & \posy \\ \texttt{moon} & 0.704 & 27 & \posy \\ } & \cellresults{\texttt{dog} & 0.643 & 26 & \posy \\ \texttt{fire} & 0.658 & 23 & \posy \\ \texttt{moon} & 0.723 & 28 & \posy \\ } & \cellresults{\texttt{dog} & 0.677 & 26 & \posy \\ \texttt{fire} & 0.639 & 23 & \posy \\ \texttt{moon} & 0.607 & 24 & \posy \\ }\\
    \rotatebox{90}{\hspace{-45pt}\large\textbf{\texttt{Stable Diffusion 2}}} & \imgf{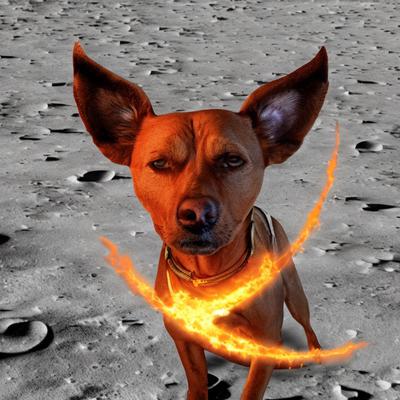}{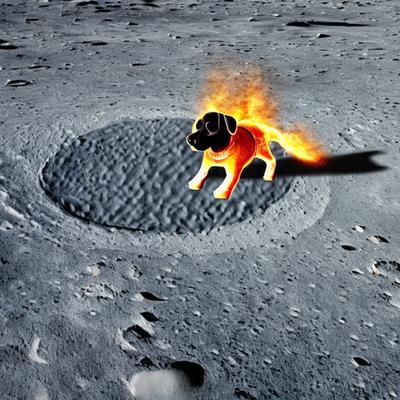}{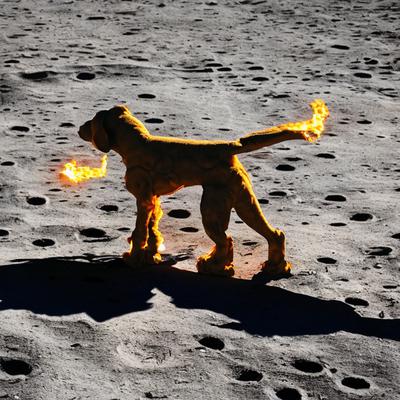}{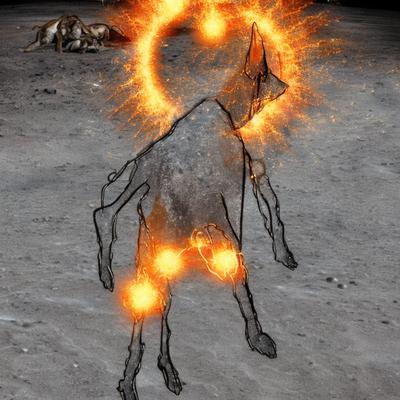} & \imgf{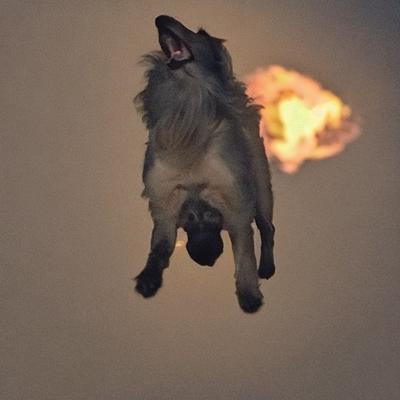}{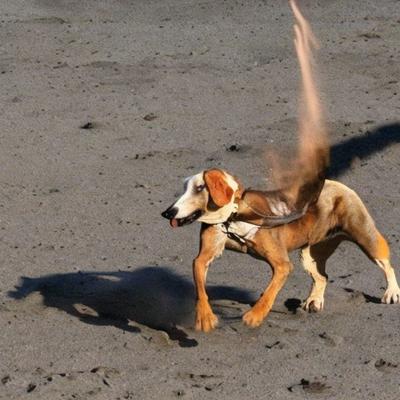}{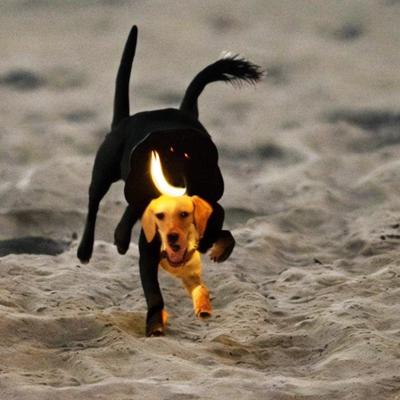}{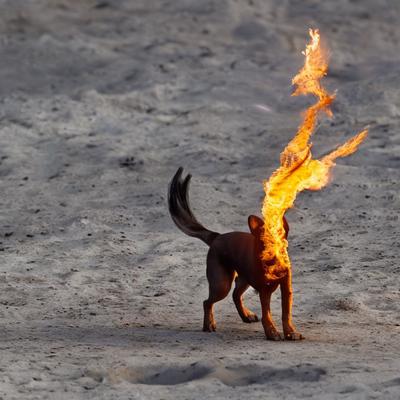} & \imgf{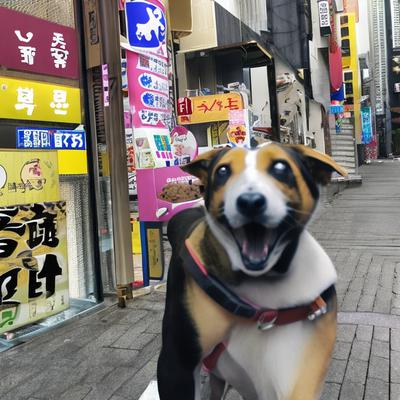}{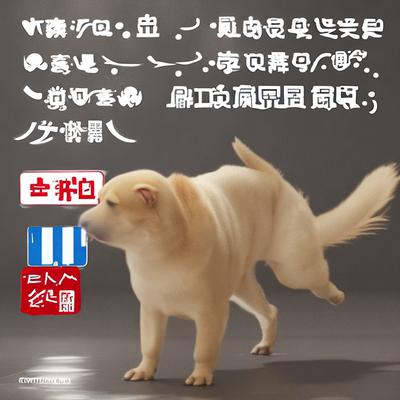}{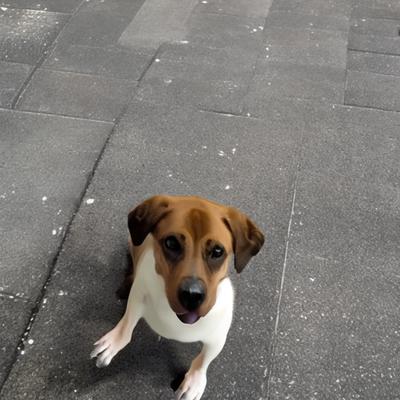}{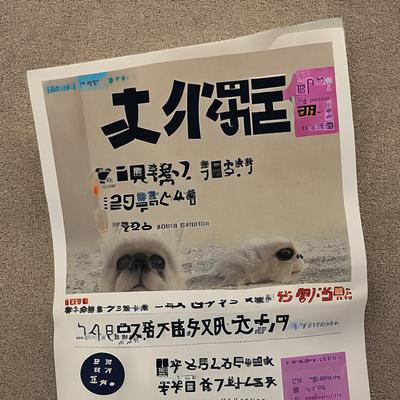}\\
      & \cellresults{\texttt{dog} & 0.748 & 26 & \posy \\ \texttt{fire} & 0.775 & 25 & \posy \\ \texttt{moon} & 0.756 & 28 & \posy \\ } & \cellresults{\texttt{dog} & 0.712 & 26 & \posy \\ \texttt{fire} & 0.620 & 23 & \posy \\ \texttt{moon} & 0.763 & 29 & \posy \\ } & \cellresults{\texttt{dog} & 0.582 & 25 & \posy \\ \texttt{fire} & 0.292 & 20 & \posn \\ \texttt{moon} & 0.282 & 19 & \posn \\ }\\
  
    \bottomrule
    \end{tabular}}

        \caption{``a dog made of fire standing on the moon''}
    \label{fig:firedogmoon}
\end{subfigure}

    \caption{Cross-model analysis of more complicated, creative prompts combining concepts including ``snow,'' ``keyboard,'' ``bird,'' ``dog,'' ``fire,'' and ``moon.'' We find that \textbf{if a model is found to not possess a concept, it will not be able to produce more complicated prompts including the concept.} This validates \cococrola~as an efficient way to capture an overview of a model's generalization capabilities.}
    \vspace{-2ex}
    \label{tab:storyprop}
\end{figure*}

\subsection{Distinctiveness captures generic collisions}

\autoref{fig:histspecall} shows the distribution of the \textbf{\color{disYellow} inverse distinctiveness} score Dt. On this plot, more rightward probability mass indicates a distribution of concepts for which distinctiveness is \textbf{low} relative to a generic sample of images produced by a given model in that language. The four Latin script languages (EN, ES, DE, ID) exhibit the lowest inverse distinctiveness, and are thus the least prone to producing generic failure images.
Hebrew is an outlier for having high concept-level Dt. Across many models. This is probably because it is the most low-resource language in our list by far, and doesn't benefit from script sharing.

\subsection{Ranking concepts by Xc}\label{subsec:xcrank}

For a given model and language, \cococrola~can be used as a concept-level analysis tool. For example, by performing the same ranking over a specific (model, language) pair, we can find the most well-covered and poorly-covered concepts for that pair. For all models and languages, an interactive ranking demo based on ascending and descending Xc and Wc is available at  \href{https://saxon.me/coco-crola/}{\texttt{saxon.me/coco-crola/}}. For example, we found ``snow'' to be a concept possessed in EN and ES for DALL-E Mega, AltDiffusion, and Stable Diffusion, but only possessed in JA by Stable Diffusion 2. A similar situation holds for ``dog'' and ``fire,''  with AltDiffusion.

\subsection{Concept possession as a proxy}\label{subsec:conceptprop}

In this section we will discuss how \textbf{an inability for a model to use some concept in complex, creative prompts is implied by our detected non-possession of said concept,} thereby validating the \cococrola~atomic evaluation paradigm.

To investigate this we manually translated two creative prompts including concepts found to be differentially present in DALL-E Mega, AltDiffusion, and Stable Diffusion \autoref{subsec:xcrank} from English into Spanish and Japanese. The prompts were: ``a bird using a keyboard in the snow,'' (ES: ``un pájaro usando un teclado en la nieve,'' JA: ``\inlinejp{雪にキーボードを使っている鳥}'') and ``a dog made of fire standing on the moon,'' (ES: ``un perro hecho de fuego pisando en la luna,'' JA: ``\inlinejp{火でできた犬が月に立っている}''). 

\autoref{tab:storyprop} clearly shows that, using thresholds for non-possession of $\textrm{Xc} < 0.5$ and $\textrm{Wc} < 25$, \textbf{if a concept is not possessed by a model according to \cococrola, it will be unable to successfully generate creative images containing it}. 

However, other capabilities including compositionality and perhaps a sort of \textit{verb-level conceptual possession} are probably also required in order to make the converse (possession implies capability to generate creatively) to be true---yet we have no method to capture such possession. This is a promising direction for future work.

Thus, concept-level coverage in a model can be used as a proxy for generalization capabilities to more complex prompts containing the concept, at least in the case of tangible noun concepts.
This is good news, as it \textbf{enables assessment of the infinite space of creative prompts from a finite, constrained set of atomic concepts.}

\section{Conclusion}

Multilingual analysis of text-to-image models is desirable both to improve the multicultural accessibility of T2I systems and to deepen our understanding of their semantic capabilities and weaknesses. 
Analyzing a model's \textit{conceptual coverage} is a simple and straightforward way to do this.

We demonstrated that these concepts are core building blocks for producing impressive images, and that analyzing the concepts is a useful proxy for assessing the creative capabilities of T2I models.

Our technique, \cococrola~is a first step toward further work in this domain. Following our recipe, larger benchmarks containing more languages and concepts can be built easily.

\section*{Limitations}

The \cococrola~benchmark generating procedure is intended to yield multilingual evaluations that can be scaled to even larger sets of concepts and languages without experienced annotators. 
In the interests of both concept and language quantity scale, we opted for an automated procedure which leverages machine translation systems, which can introduce translation errors. %
Furthermore, variation in the nuance or normative meaning of concepts, particularly culturally contested ones such as ``face,'' \cite{engelmann2022faces} ``person,'' or ``man'' will inevitably drive some variance in expected outputs by users across language communities. This cultural variation will place an unavoidable upper bound on the performance of inherently cross-cultural benchmarks such as \cococrola.

Additionally, typological variation between languages can introduce complications in applying our framework. For example, while simple template filling for prompting is straightforward in Chinese, which requires no word-dependent articles, in English phonological properties of the word govern the preceding article, and in Spanish and German grammatical gender do the same. Hebrew has gendered nouns, adjectives, and verbs but not articles, on the other hand. Overall, it appears that these have limited influence as grammaticality isn't a crucial feature in the prediction of image tokens performed in T2I models, \autoref{subsec:grammareffects}.

While doing so aids in the scalability of the approach, using CLIP as a feature extractor for computing the metrics, particularly correctness Xc and Wc, potentially introduces biases due to the English-primary data that CLIP is pretrained on. Future work could test this hypothesis by comparing the performance of \cococrola's CLIP-based features with Xc computed using Inception features (as in FID) \cite{chong2020effectively} or with dedicated concept-level purpose-trained classifiers.

\section*{Ethics Statement}

Images of human faces are generated by our model. To mitigate the minor risk of resemblance to real people, we have downsampled all images. However, we believe this risk is mitigated by the lack of personal names in the querying data. Furthermore, we believe demonstrating that human faces are generated and under which conditions they are is important for documentation of bias \cite{paullada2021discontents} and harm risks in these models.

License information is provided in our project page (\href{https://saxon.me/coco-crola}{\texttt{saxon.me/coco-crola}}) and project repository (\href{https://github.com/michaelsaxon/CoCoCroLa}{\texttt{github:michaelsaxon/CoCoCroLa}}).

\section*{Acknowledgements}

This work was supported in part by the National Science Foundation Graduate Research Fellowship under Grant No. 1650114. 
This work was further supported by the National Science Foundation under Grant No. 2048122. We would also like to thank the Robert N. Noyce Trust for their generous gift to the University of California via the Noyce Initiative. The views and conclusions contained in this document are those of the authors and should not be interpreted as representing the sponsors.

\bibliography{custom}
\bibliographystyle{acl_natbib}

\appendix

\label{sec:appendix}

\newpage

\section{Details on producing the concept list}\label{subsec:translation}

\paragraph{Source language term lists.} We first produce a list of English nouns by collating words in term frequency lists extracted from TV closed captions and contemporary fiction novels from Wiktionary\footnote{\href{https://en.wiktionary.org/wiki/Wiktionary:Frequency_lists/Contemporary_fiction}{\texttt{en.wiktionary.org/wiki/Wiktionary:Frequency\_ lists/Contemporary\_fiction}}, \href{https://en.wiktionary.org/wiki/Wiktionary:Frequency_lists/TV/2006/1-1000}{\texttt{.../TV/2006/1-1000}}}, and filter for the 2000 most frequent words in this combined list, and augment it with class label names from CIFAR100 \cite{krizhevsky2009learning}.

\paragraph{Finding good translations.} We feed the list English words into a custom translation pipeline, which simultaneously queries BabelNet \cite{navigli2010babelnet}, and an ensemble of four commercial translation systems: Google Translate, Bing Translate, Baidu Translate, and iTranslate\footnote{Using the \href{https://pypi.org/project/translators/}{\texttt{translators} PyPi package}.}.

In response to an English query, the BabelNet API returns a collection of ``SynSets,'' subgraphs of a combined multilingual word and entity graphs centered on a node the query word maps to (see \autoref{fig:aligneddict} for examples). Each subgraph links to multiple other nodes, containing terms in both the source language and the target language. These edges can represent, for example, the titles of Wikipedia articles in different language editions of Wikipedia that are marked as being equivalent, thus ensuring that by checking against SynSet edges, a degree of human validation is included automatically. The synset also contains information about whether a given word is a noun. If it is not a noun, the candidate concept is discarded.

To choose the best translation from those edges, the returned translations into the target languages of the English term from the commercial translation services are \textit{melded} by first sorting all returns by number of languages in the return query (in the case that one translation service covers more languages than others), and filling in missing translations by prioritizing alignment in the shared language translations. If any target language is missing a word for a concept at the conclusion of this process, that concept is discarded from the final list.

\paragraph{Post-filtering.} Once a list of melded translations from the commercial service is returned, each row in the candidate aligned concept list is checked against the corresponding BabelNet SynSets to ensure each translation is present as a connected node, for pseudo-human evaluation. At the end of this process, a list of approximately 250 concepts is returned.
Finally, we manually remove terms that are verb-noun collisions (e.g. hike) to ensure this ambiguity didn't drive any poor translations. The final list for {\color{red}\cococrola~v1.0} contains 193 concepts.

\begin{figure*}[t!]
\begin{minipage}{.33\linewidth}
\centering
\includegraphics[width=1\linewidth]{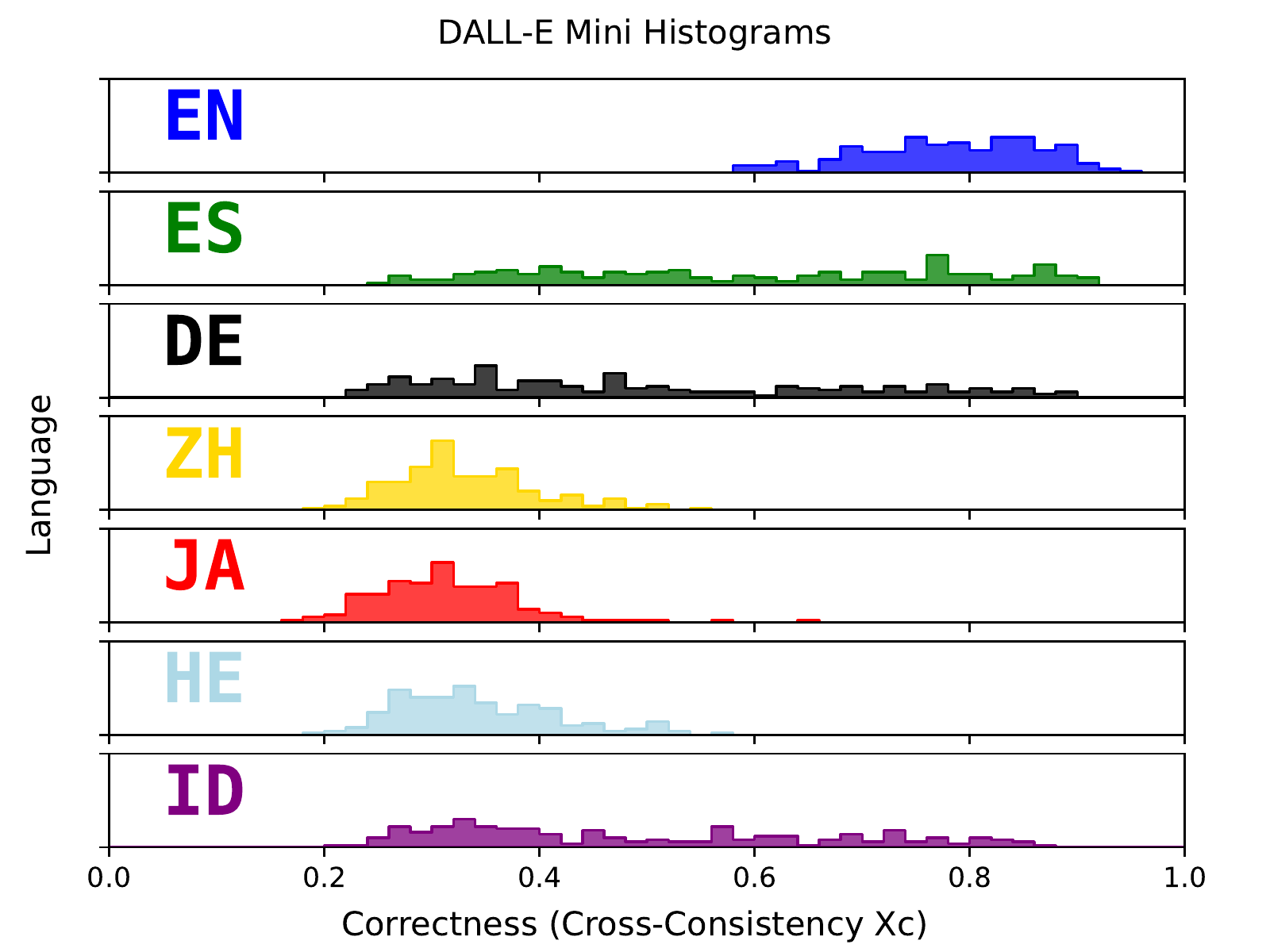}
\end{minipage}%
\begin{minipage}{.33\linewidth}
\centering
\includegraphics[width=1\linewidth]{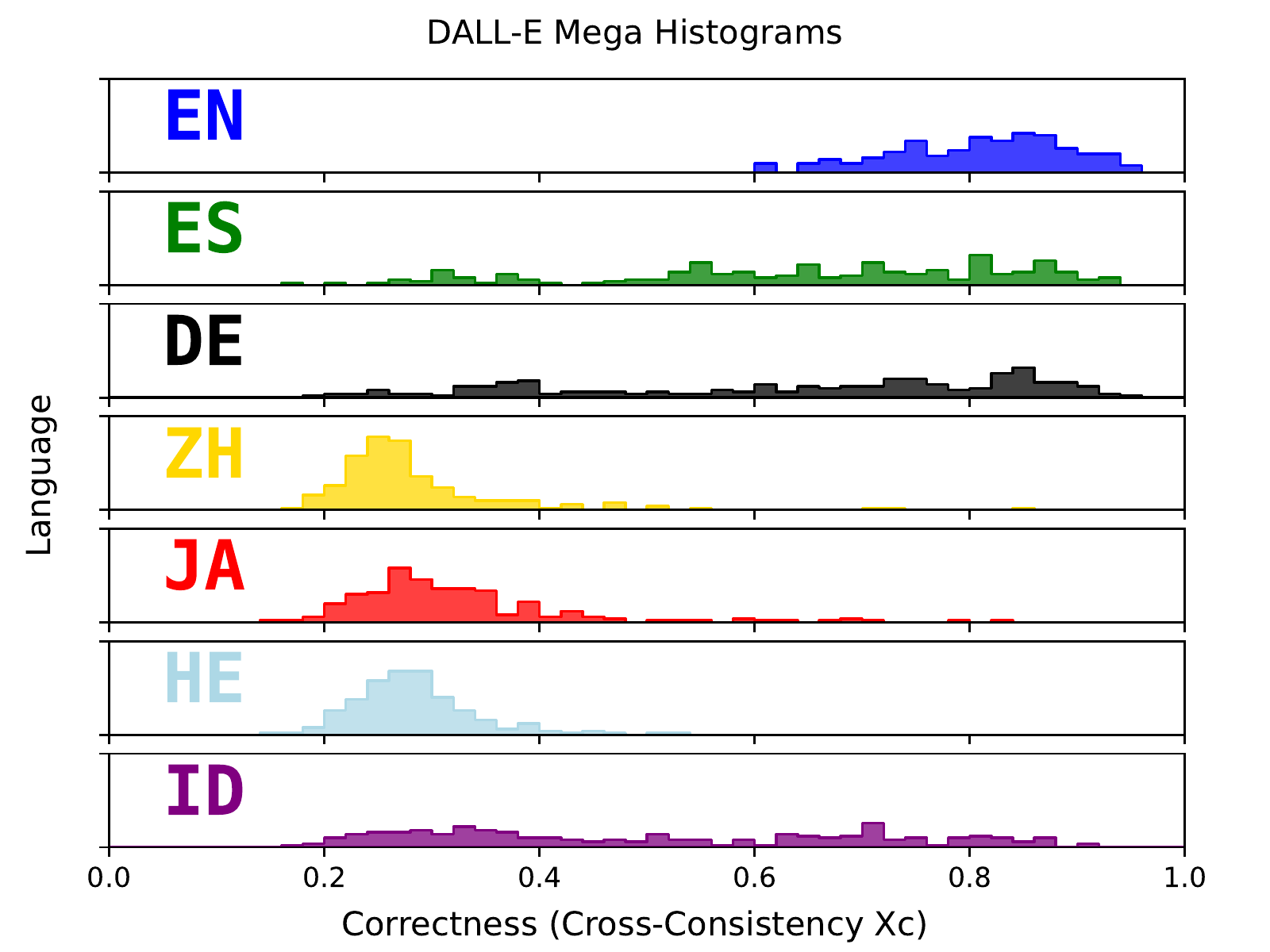}
\end{minipage}%
\begin{minipage}{.33\linewidth}
\centering
\includegraphics[width=1\linewidth]{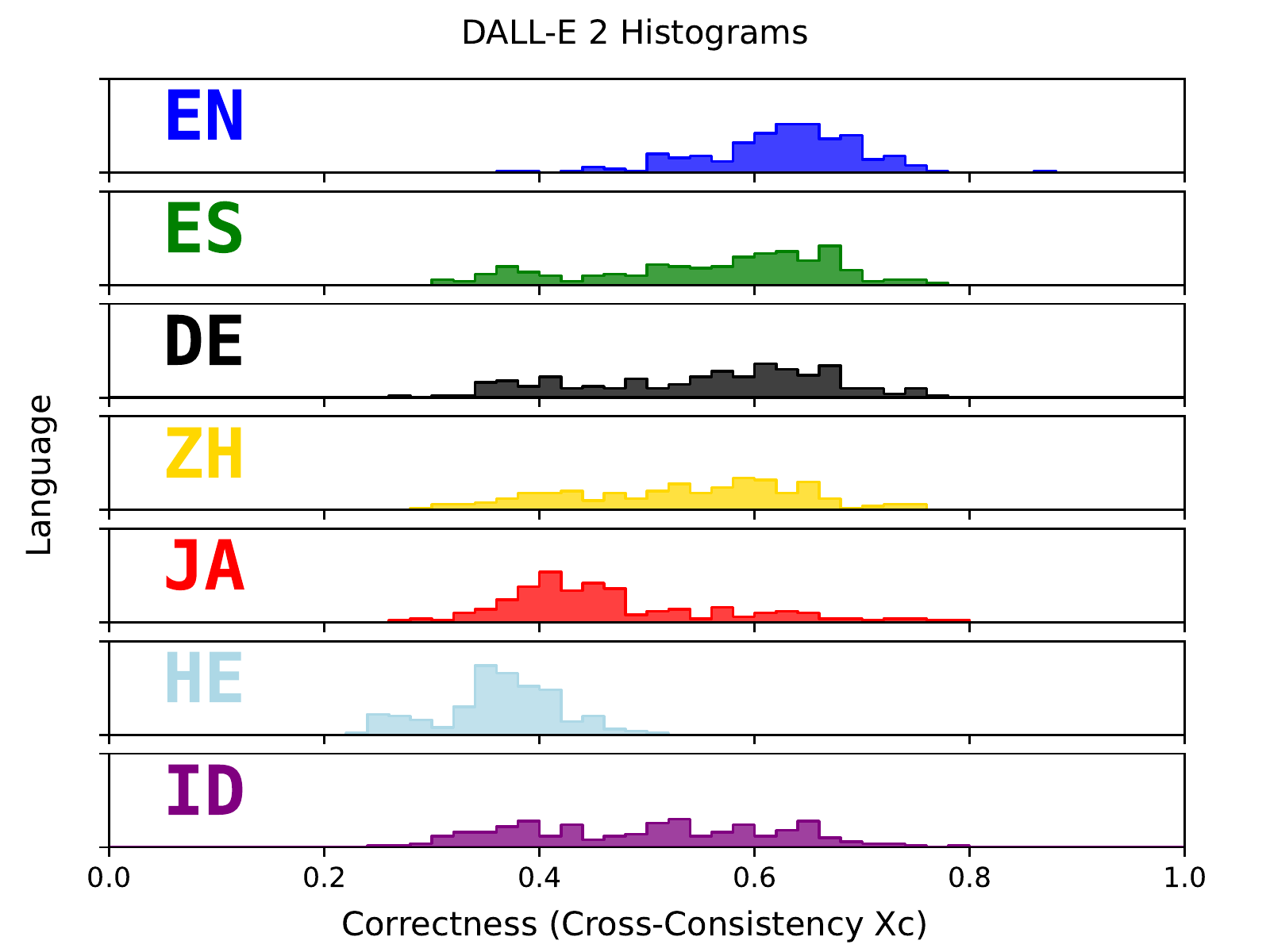}
\end{minipage}%

\begin{minipage}{.33\linewidth}
\centering
\includegraphics[width=1\linewidth]{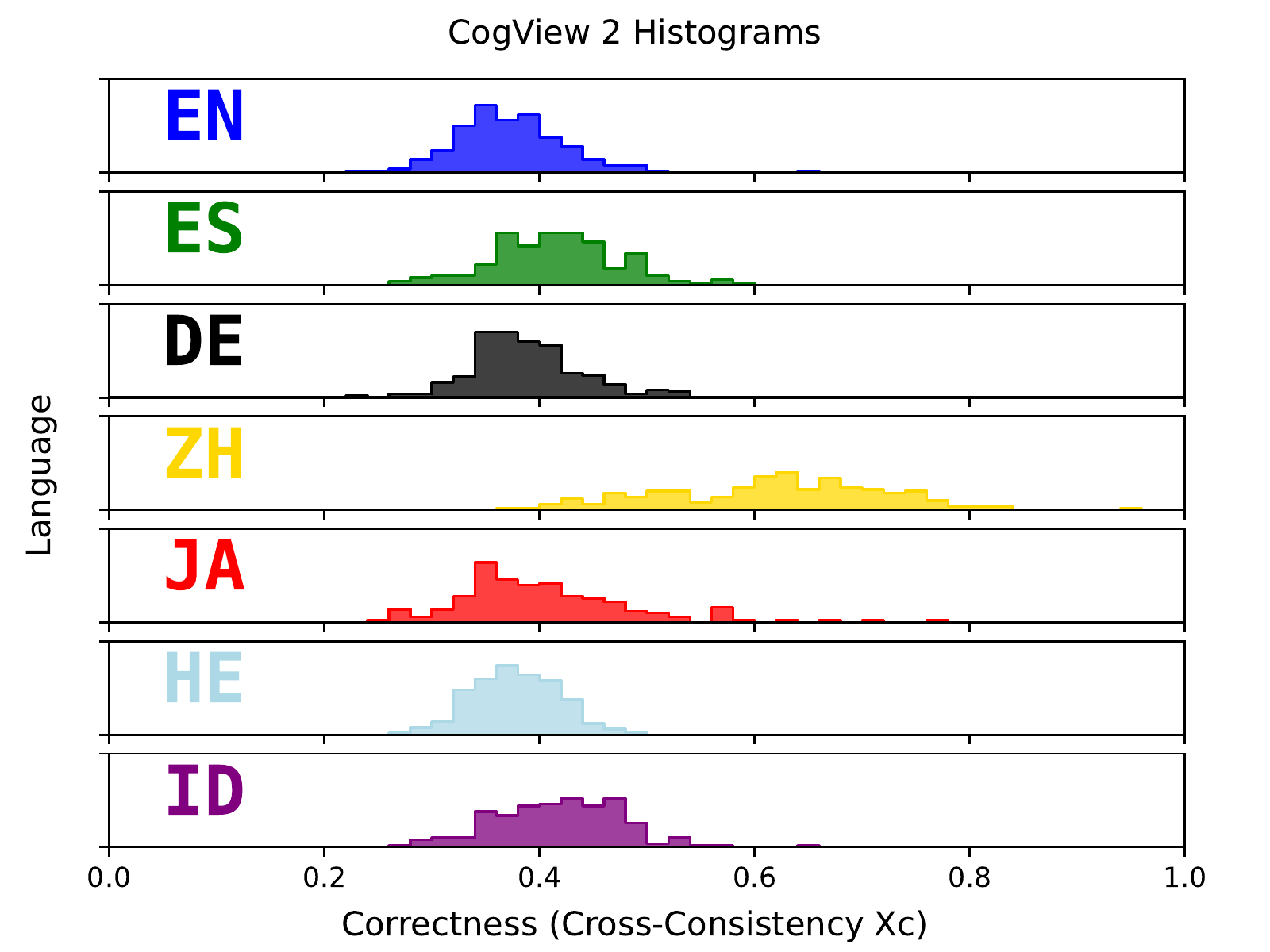}
\end{minipage}%
\begin{minipage}{.33\linewidth}
\centering
\includegraphics[width=1\linewidth]{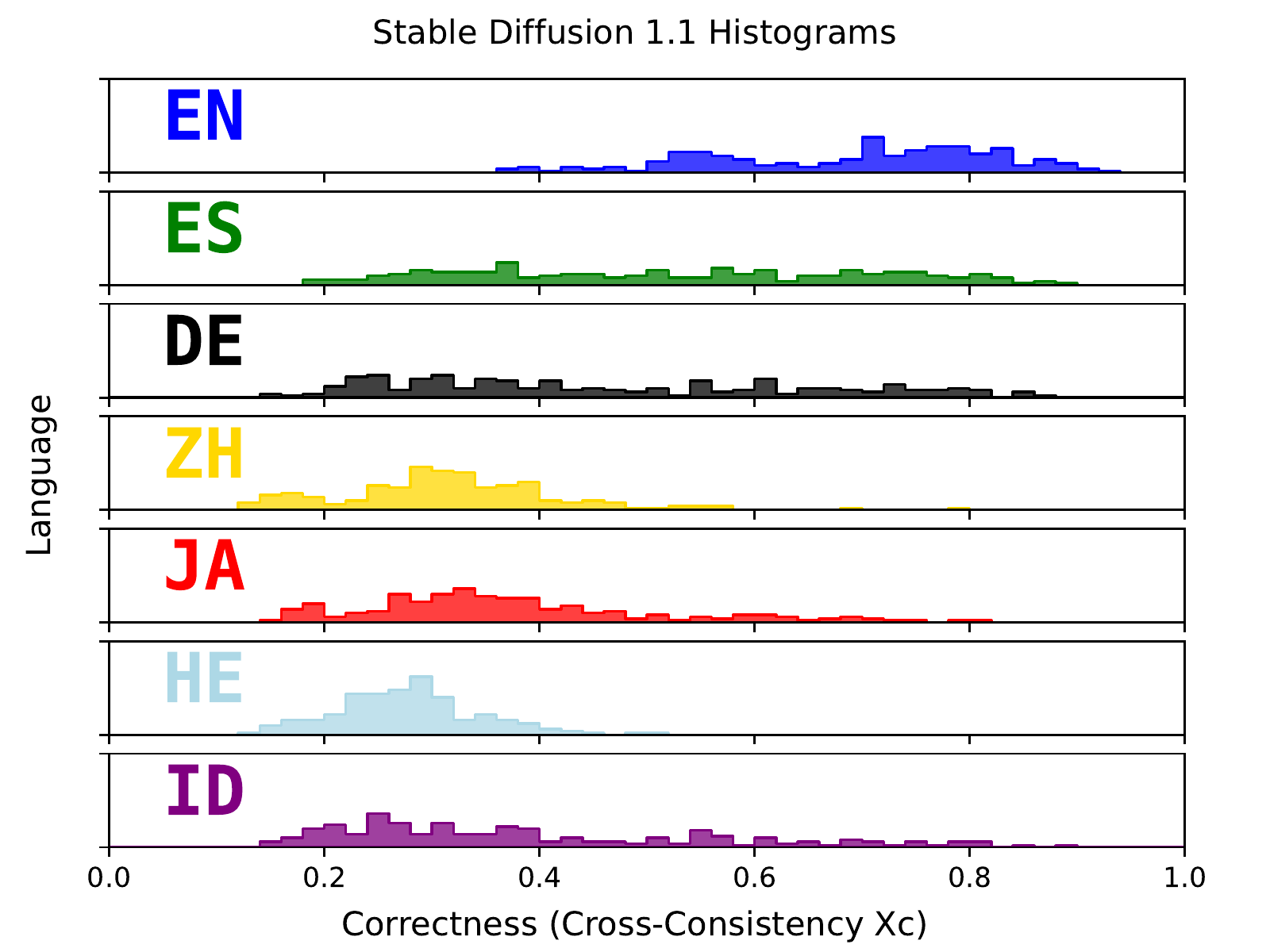}
\end{minipage}%
\begin{minipage}{.33\linewidth}
\centering
\includegraphics[width=1\linewidth]{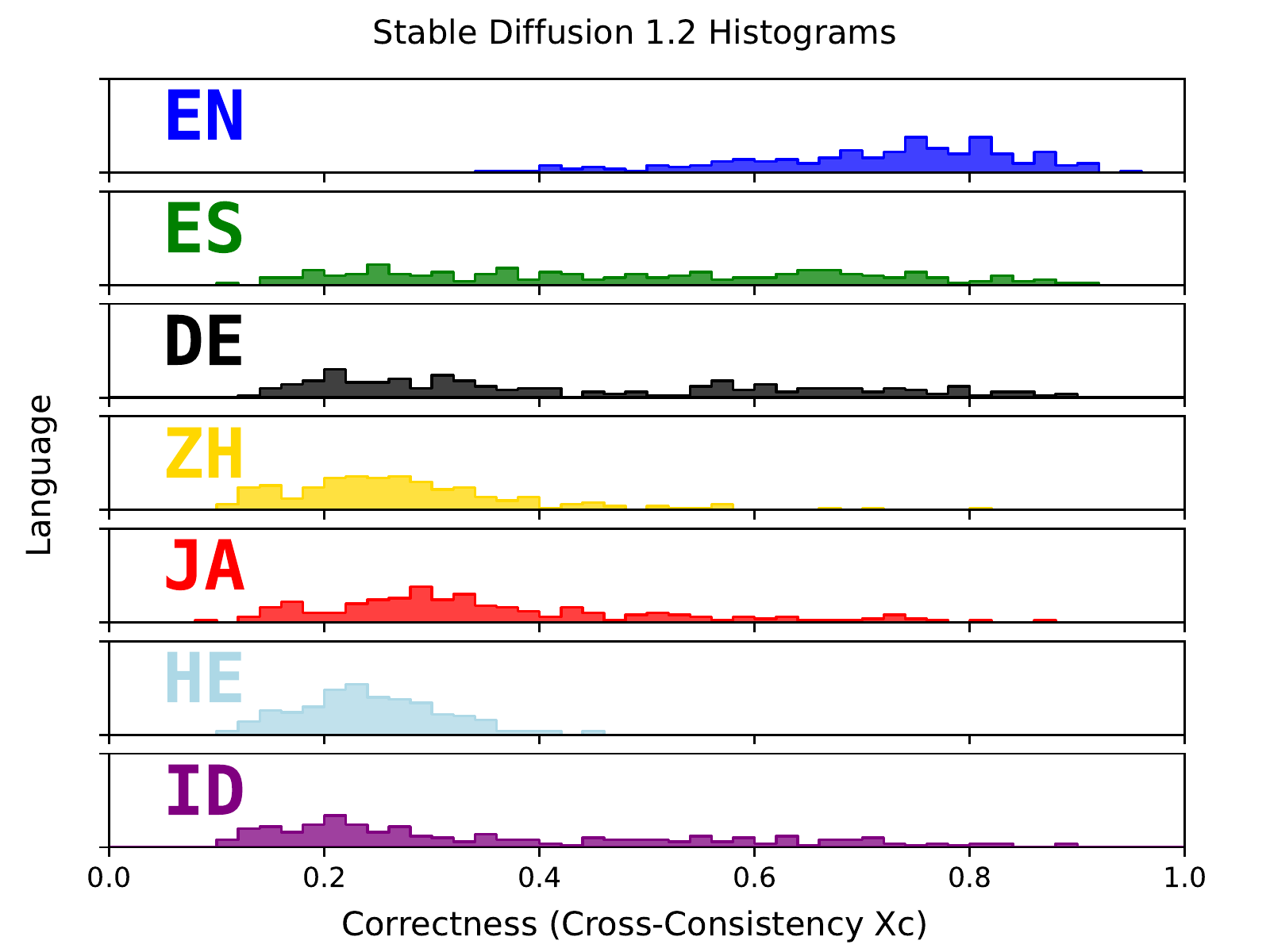}
\end{minipage}%

\begin{minipage}{.33\linewidth}
\centering
\includegraphics[width=1\linewidth]{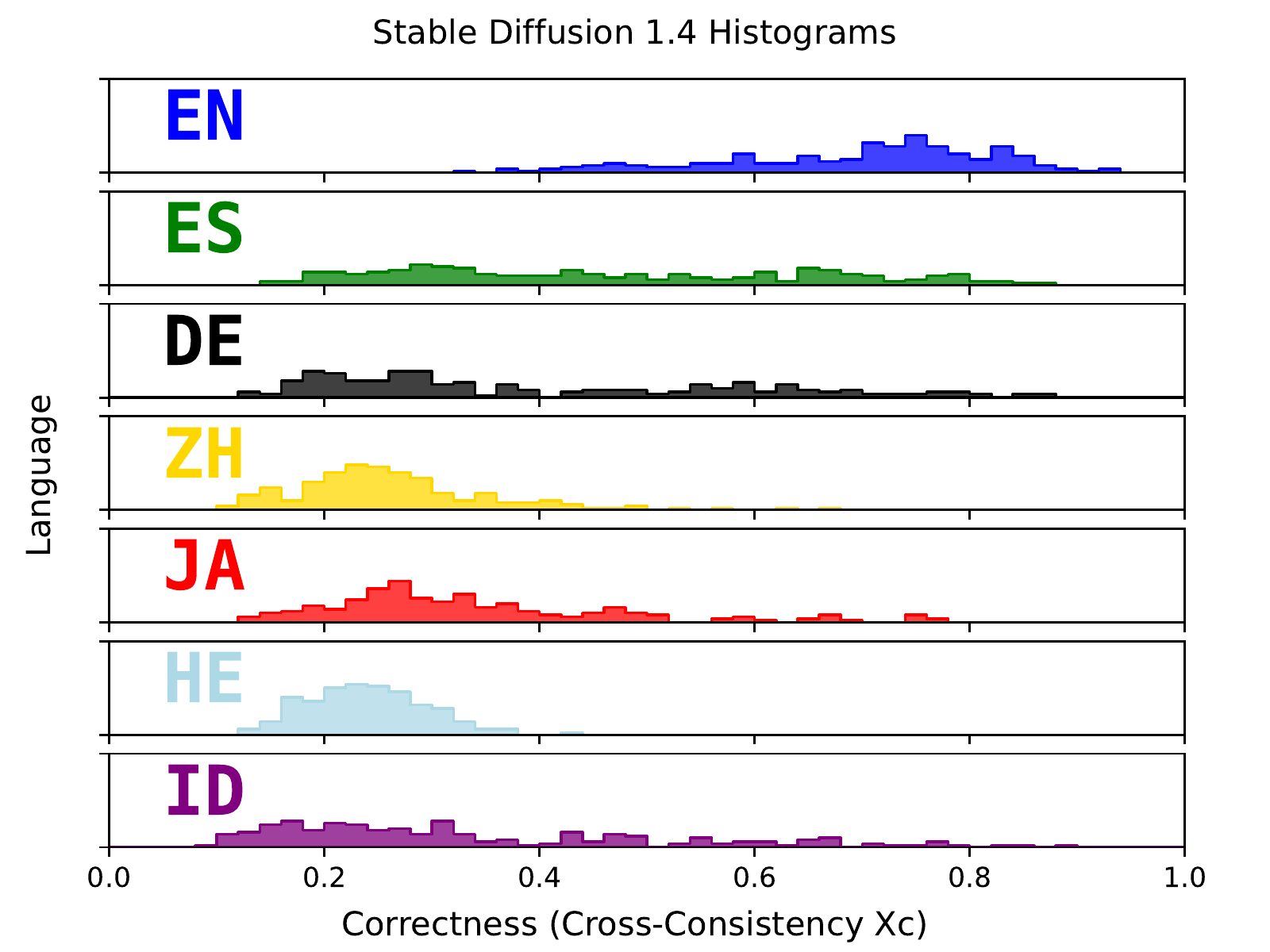}
\end{minipage}%
\begin{minipage}{.33\linewidth}
\centering
\includegraphics[width=1\linewidth]{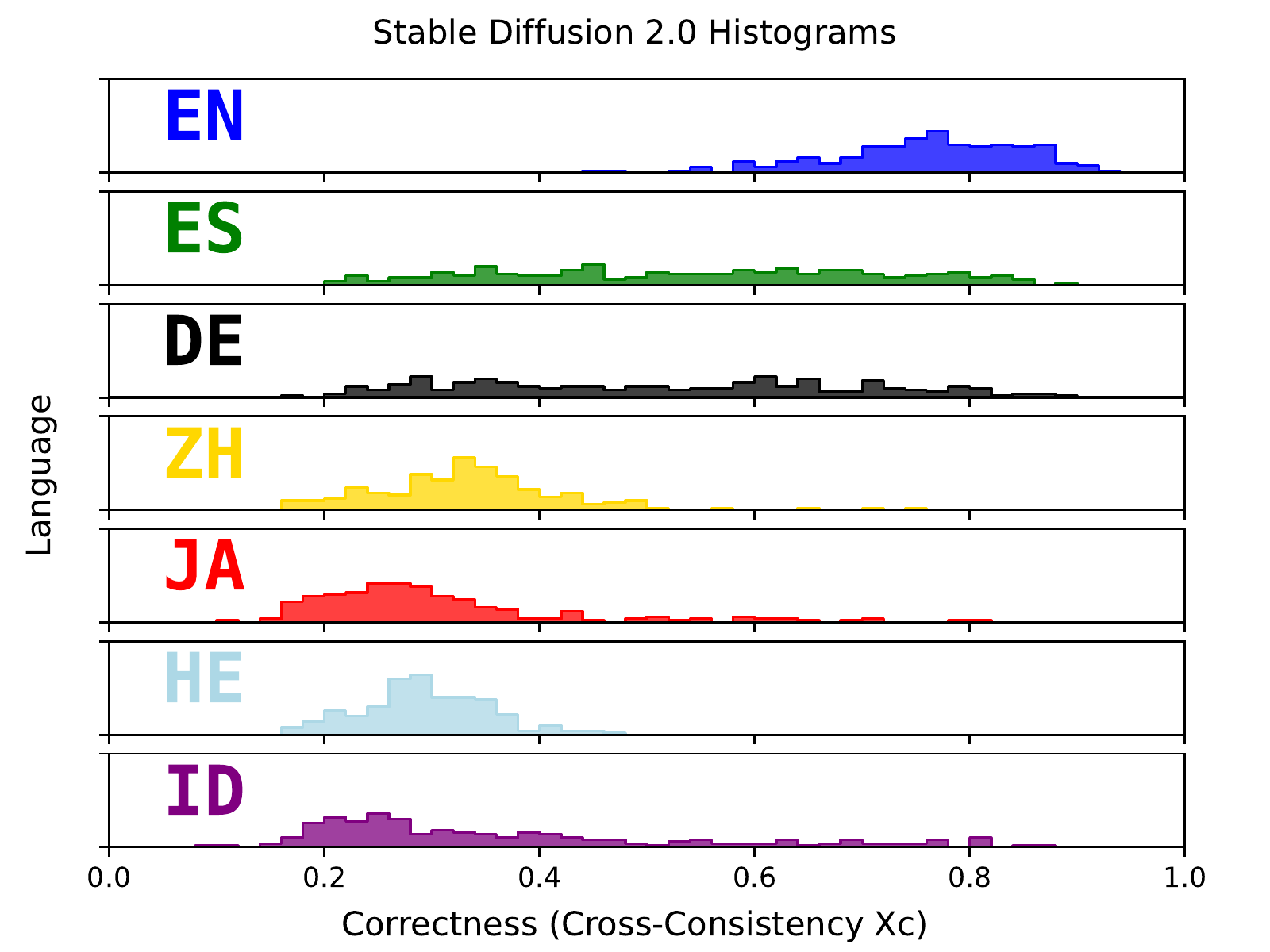}
\end{minipage}%
\begin{minipage}{.33\linewidth}
\centering
\includegraphics[width=1\linewidth]{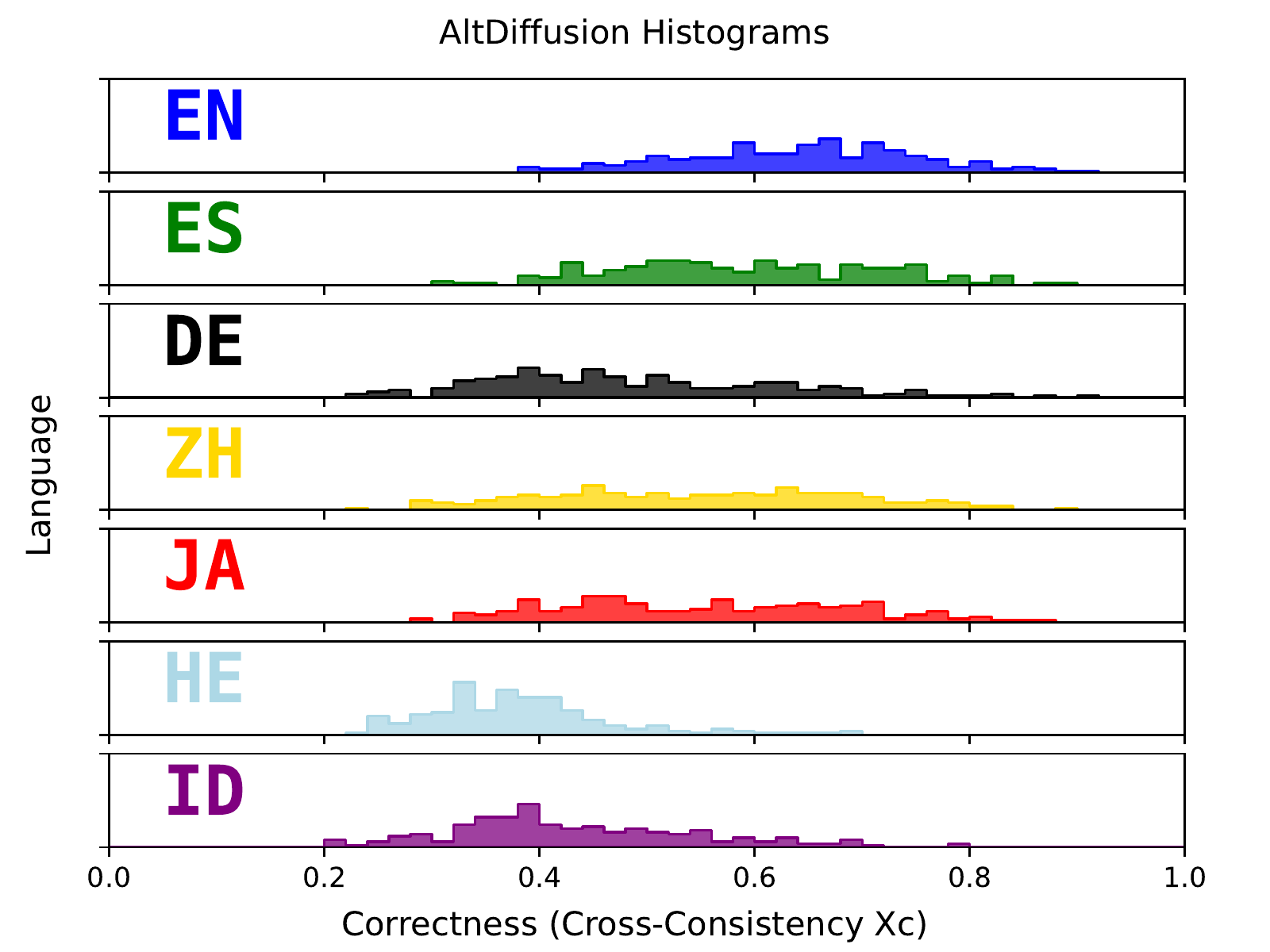}
\end{minipage}%

\caption{Histograms of the distribution of {\color{covBlue}\textbf{correctness}} cross-consistency (Xc) for each test language for all assessed models. Rightward probability mass reflects better conceptual coverage.}
\label{fig:corrful}
\vspace{-5pt}
\end{figure*}

\begin{figure*}[t!]
\begin{minipage}{.33\linewidth}
\centering
\includegraphics[width=1\linewidth]{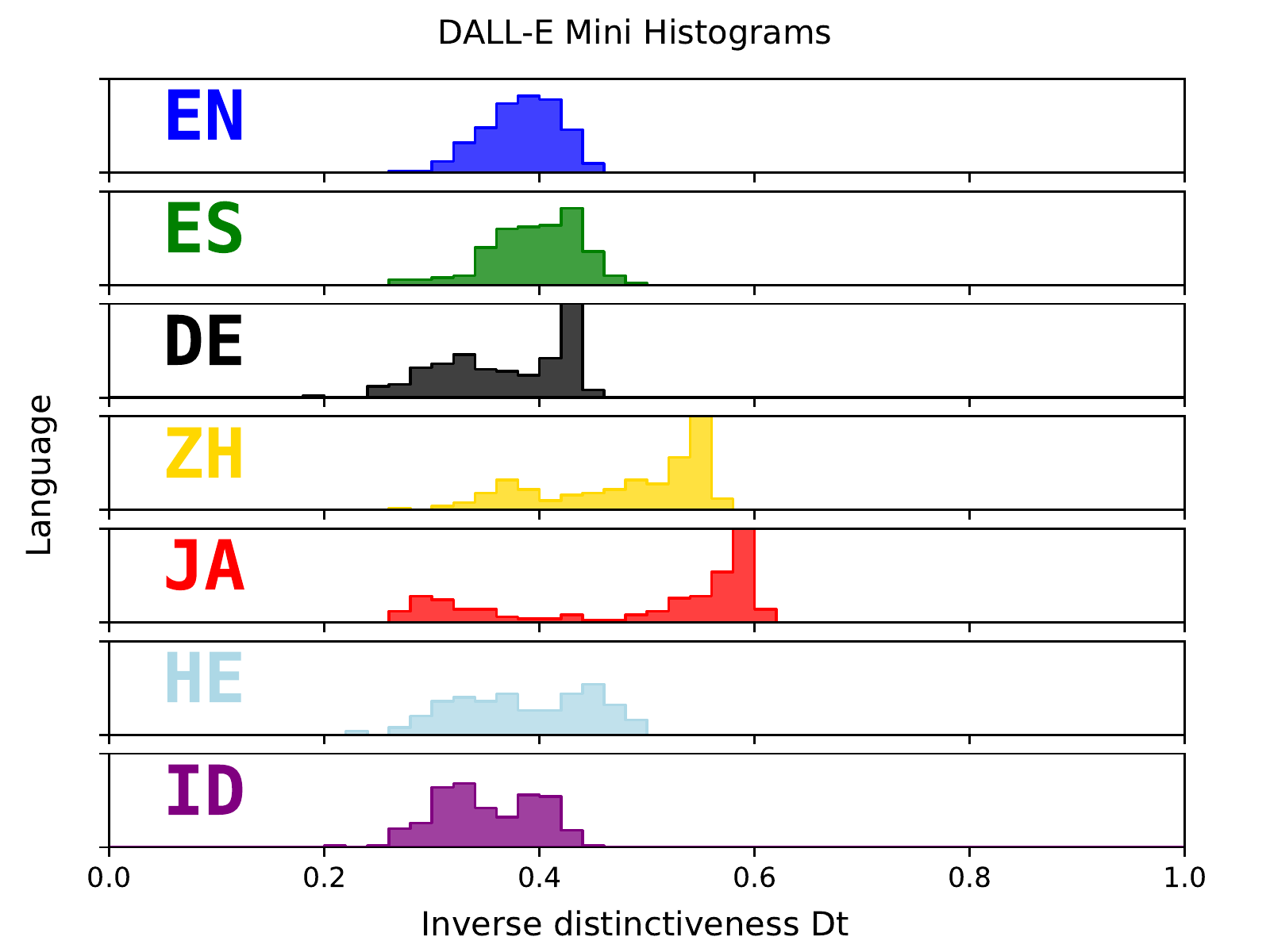}
\end{minipage}%
\begin{minipage}{.33\linewidth}
\centering
\includegraphics[width=1\linewidth]{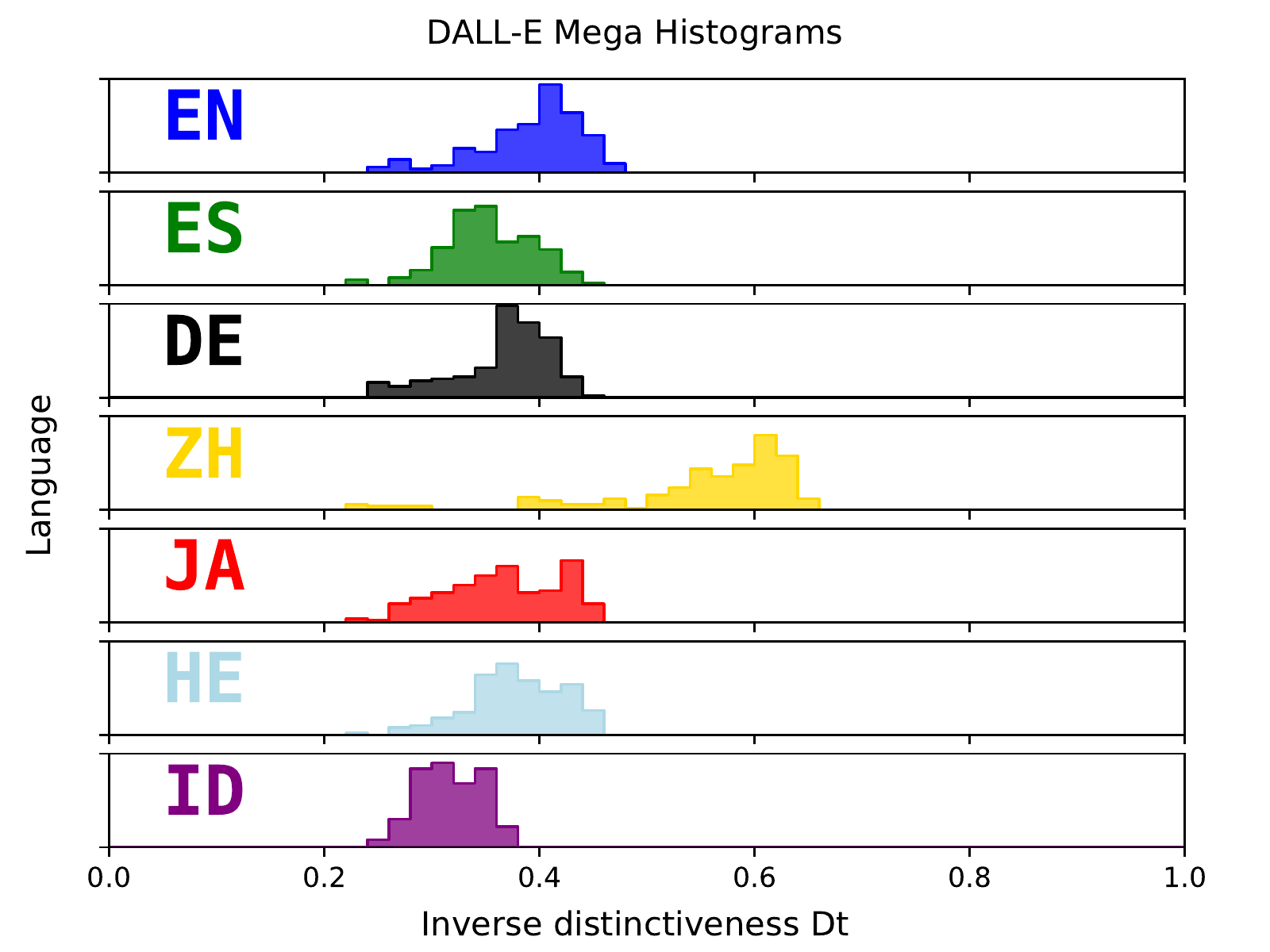}
\end{minipage}%
\begin{minipage}{.33\linewidth}
\centering
\includegraphics[width=1\linewidth]{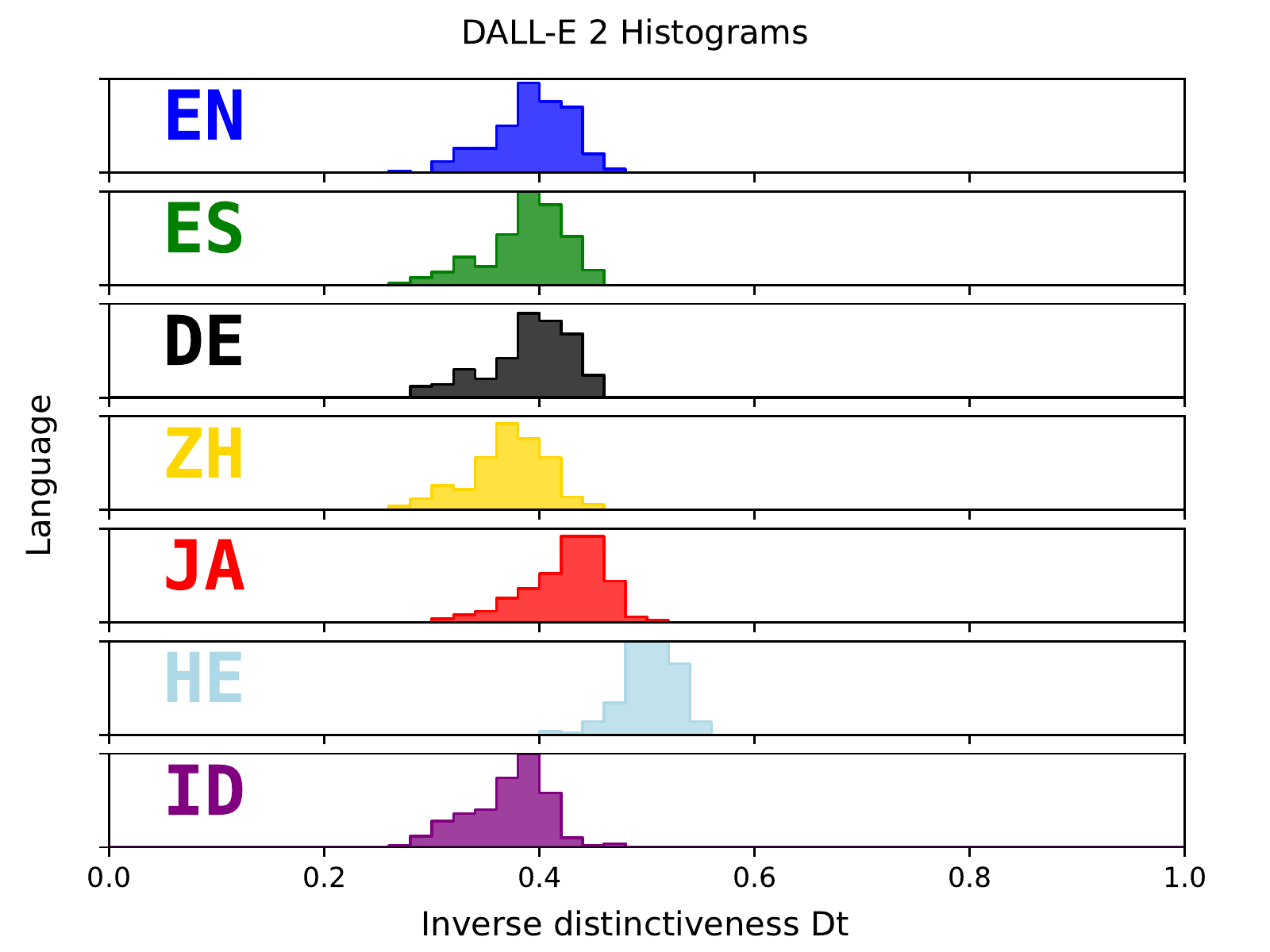}
\end{minipage}%

\begin{minipage}{.33\linewidth}
\centering
\includegraphics[width=1\linewidth]{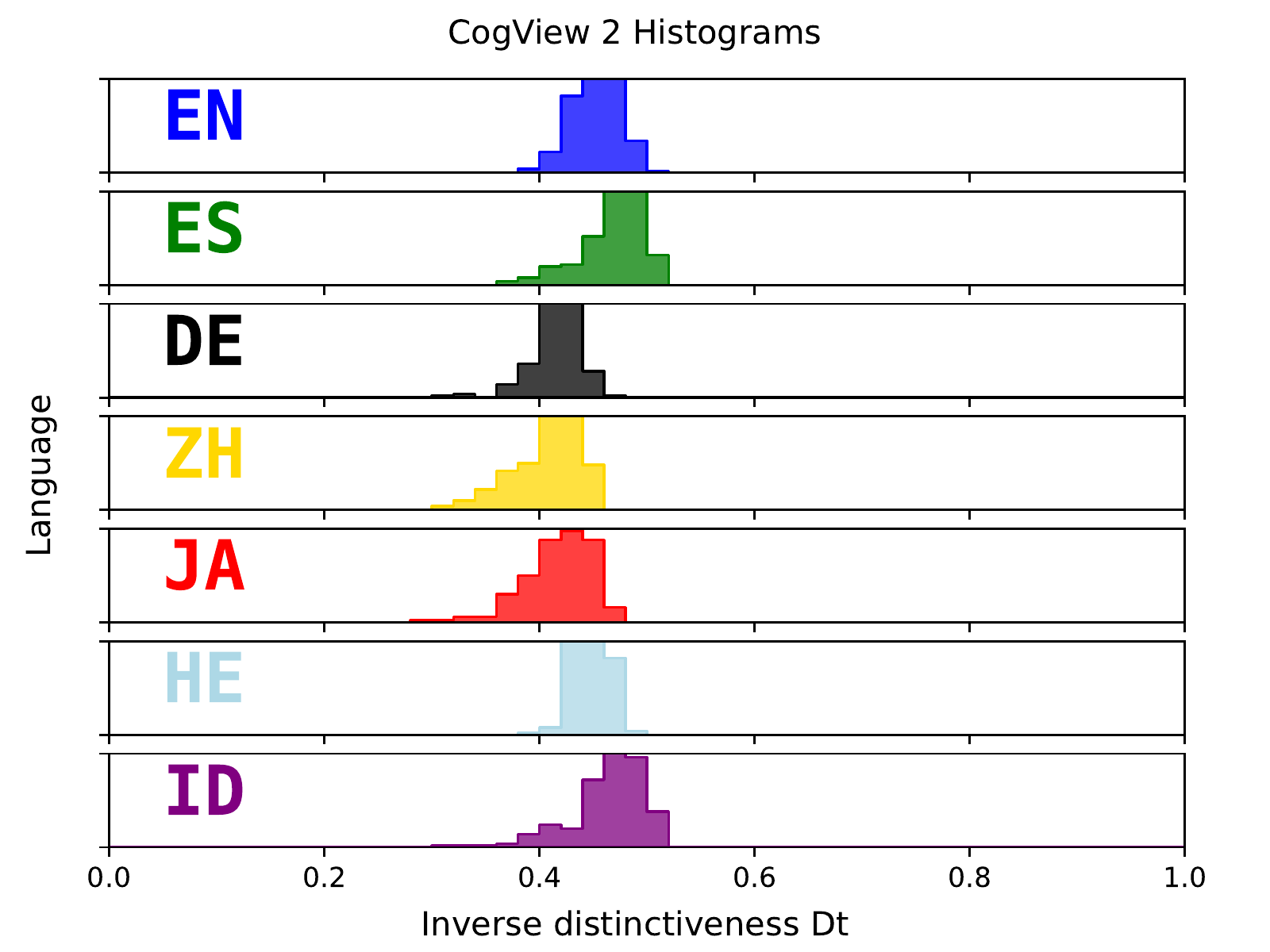}
\end{minipage}%
\begin{minipage}{.33\linewidth}
\centering
\includegraphics[width=1\linewidth]{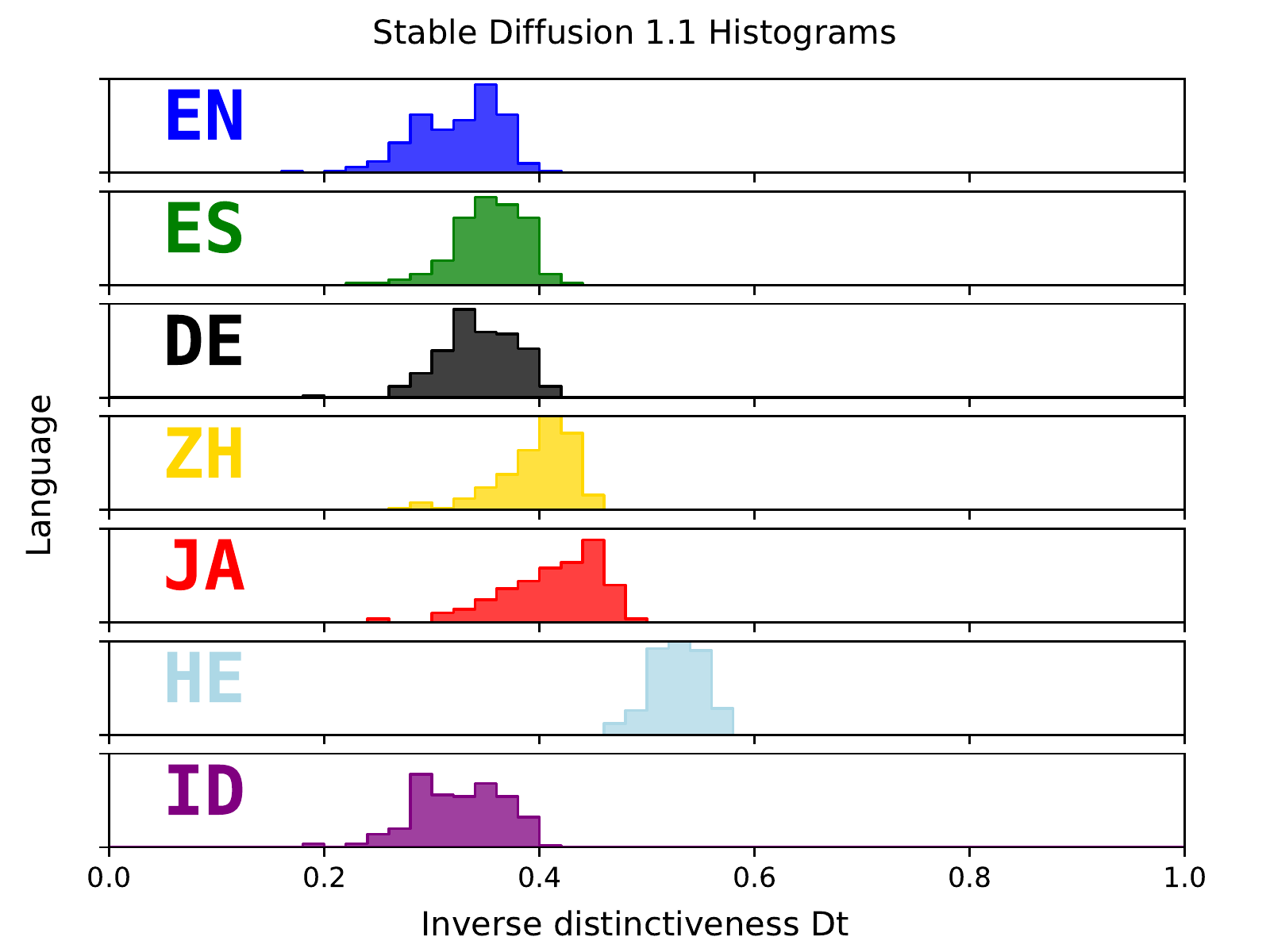}
\end{minipage}%
\begin{minipage}{.33\linewidth}
\centering
\includegraphics[width=1\linewidth]{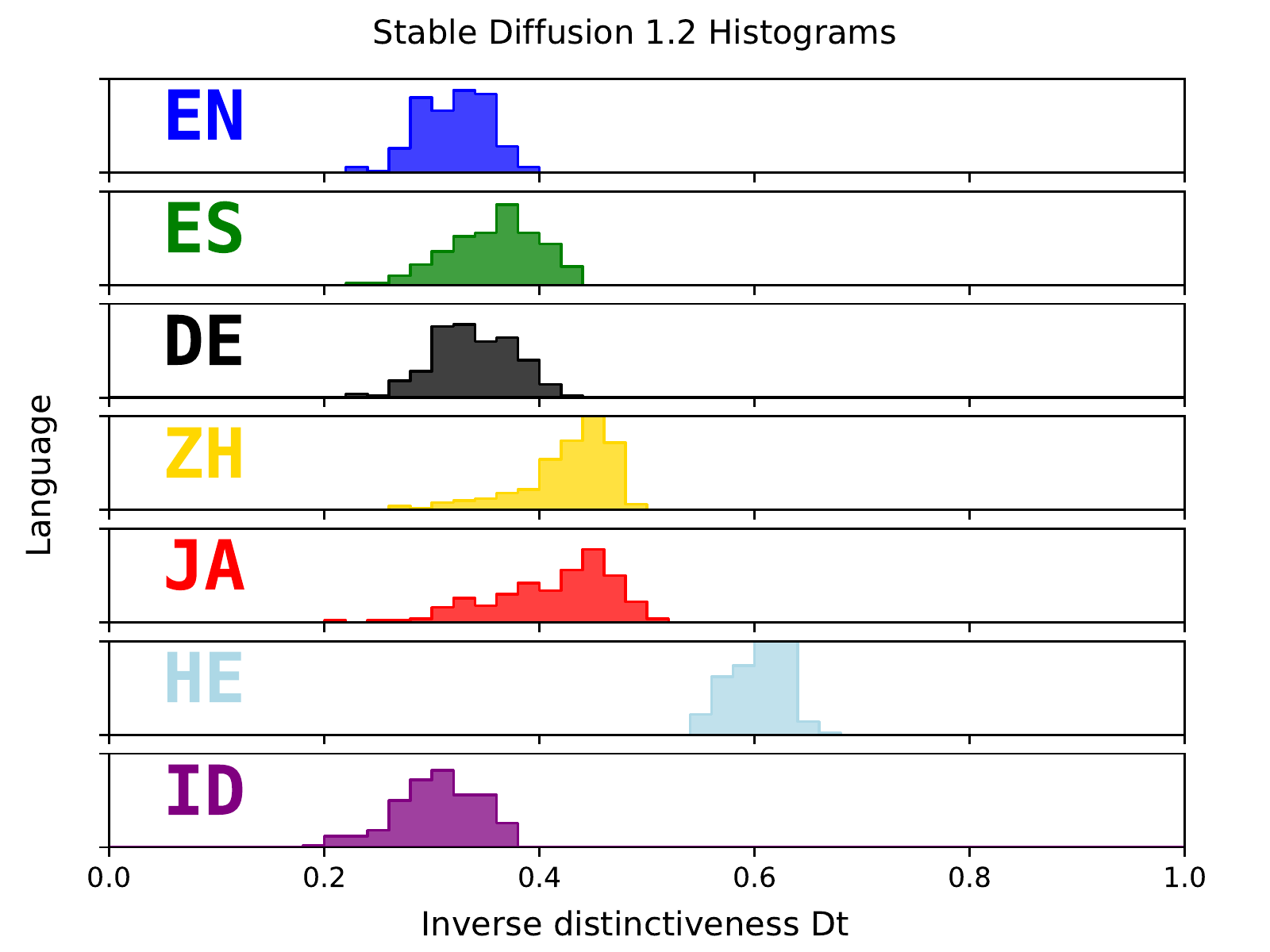}
\end{minipage}%

\begin{minipage}{.33\linewidth}
\centering
\includegraphics[width=1\linewidth]{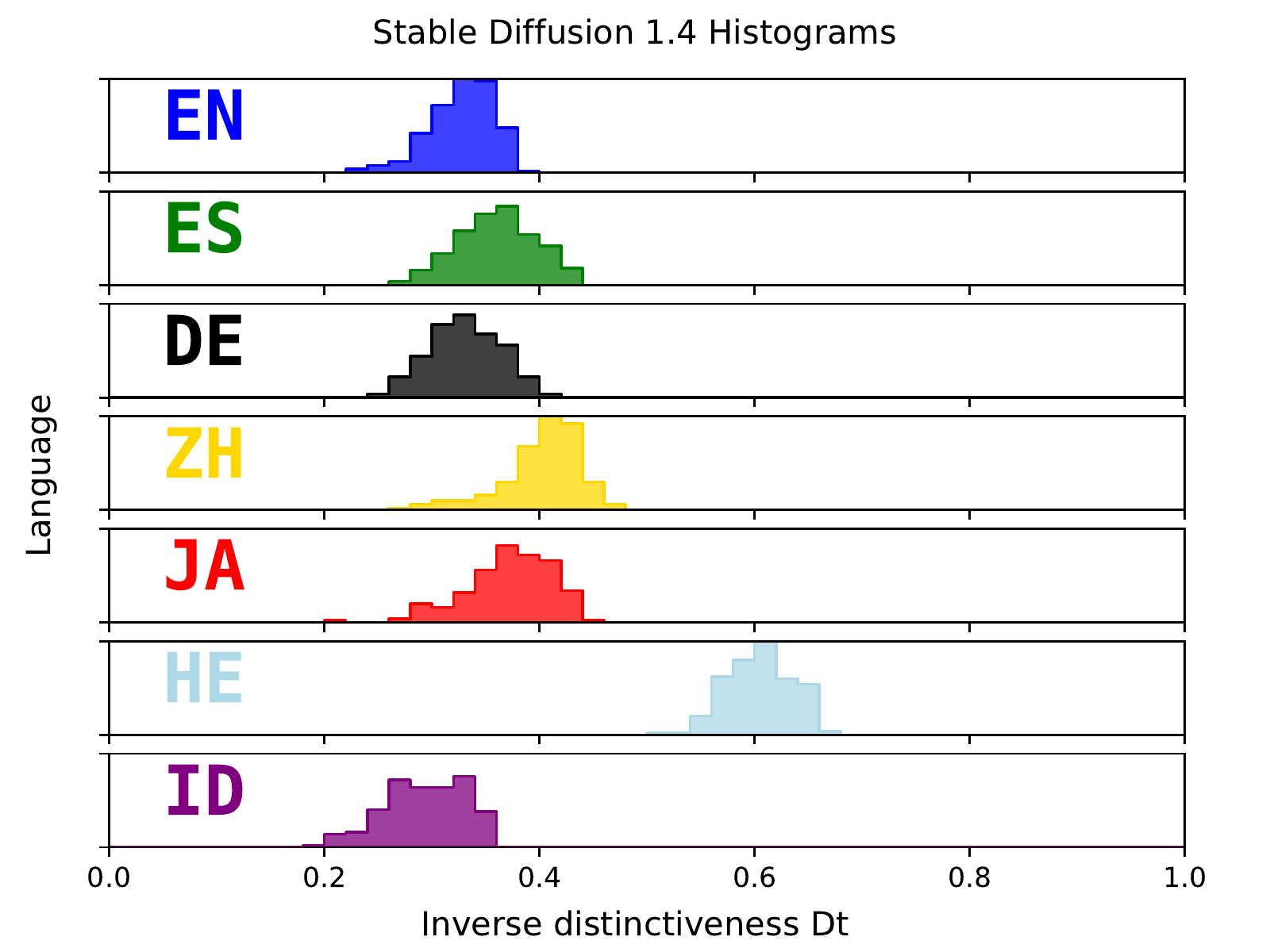}
\end{minipage}%
\begin{minipage}{.33\linewidth}
\centering
\includegraphics[width=1\linewidth]{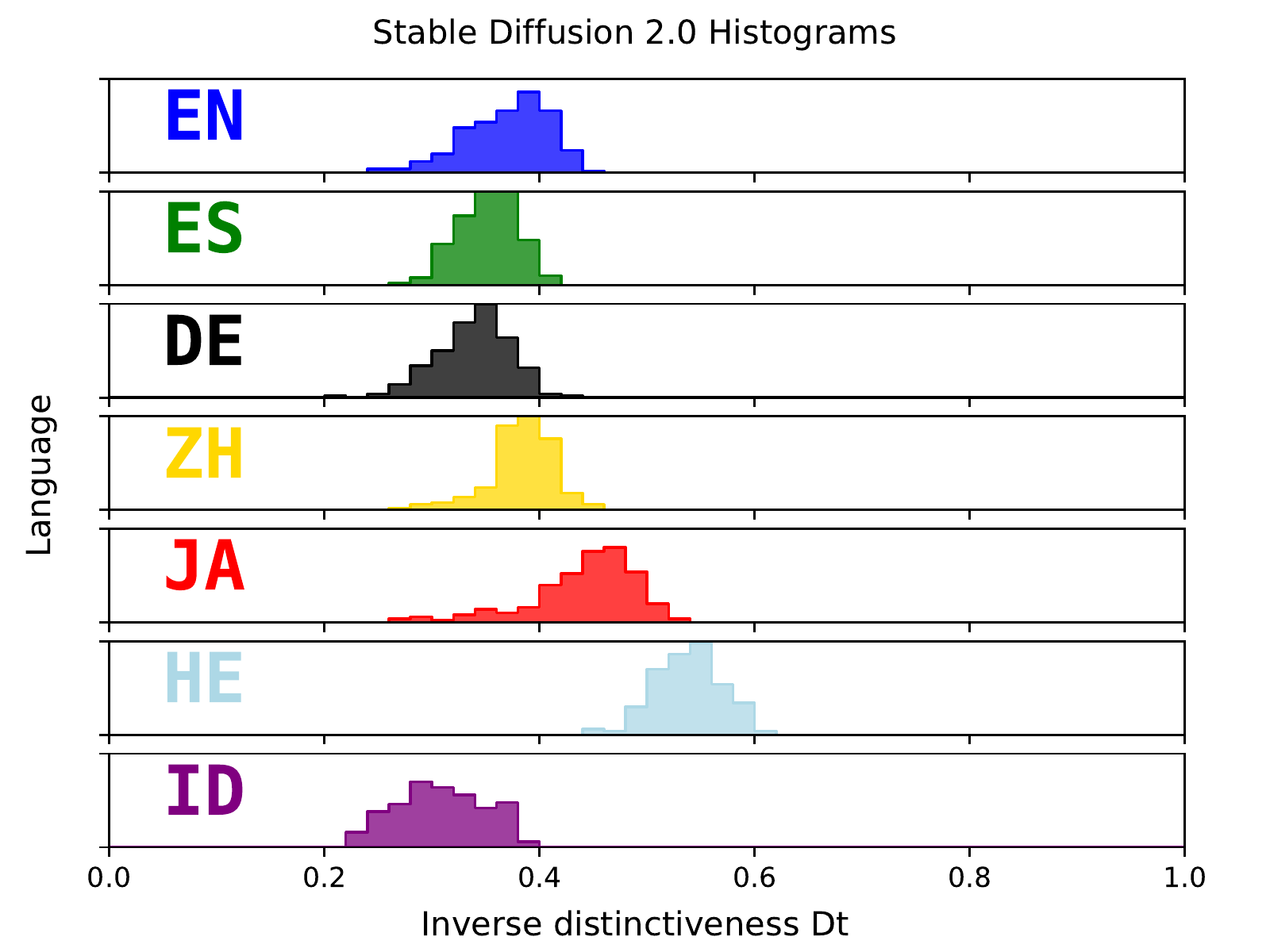}
\end{minipage}%
\begin{minipage}{.33\linewidth}
\centering
\includegraphics[width=1\linewidth]{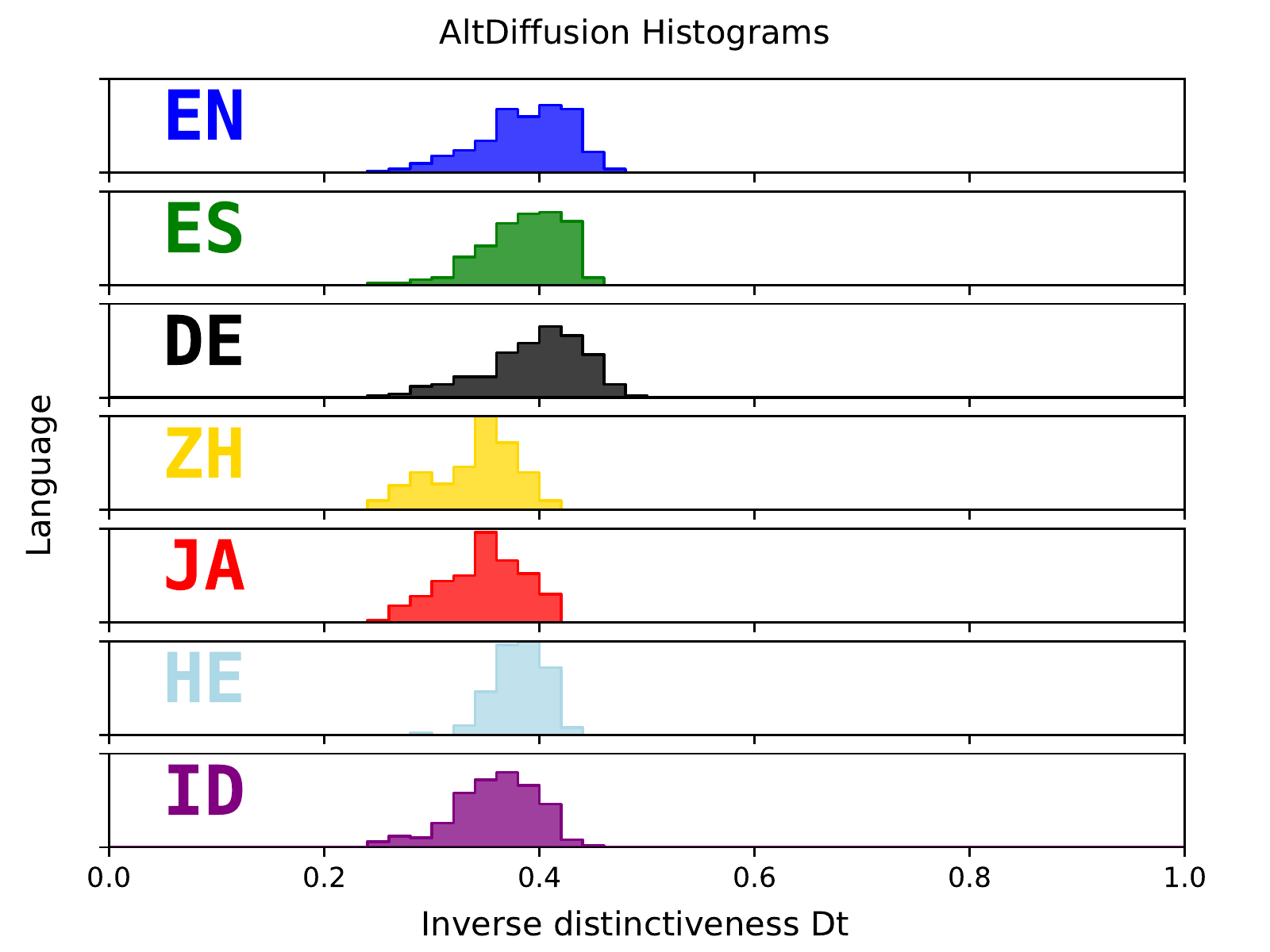}
\end{minipage}%

\caption{Distribution of {\color{disYellow}\textbf{inverse distinctiveness}} scores (Dt) for each test language for all models.}
\label{fig:disfull}
\vspace{-5pt}
\end{figure*}

\begin{figure*}[t!]
\begin{minipage}{.33\linewidth}
\centering
\includegraphics[width=1\linewidth]{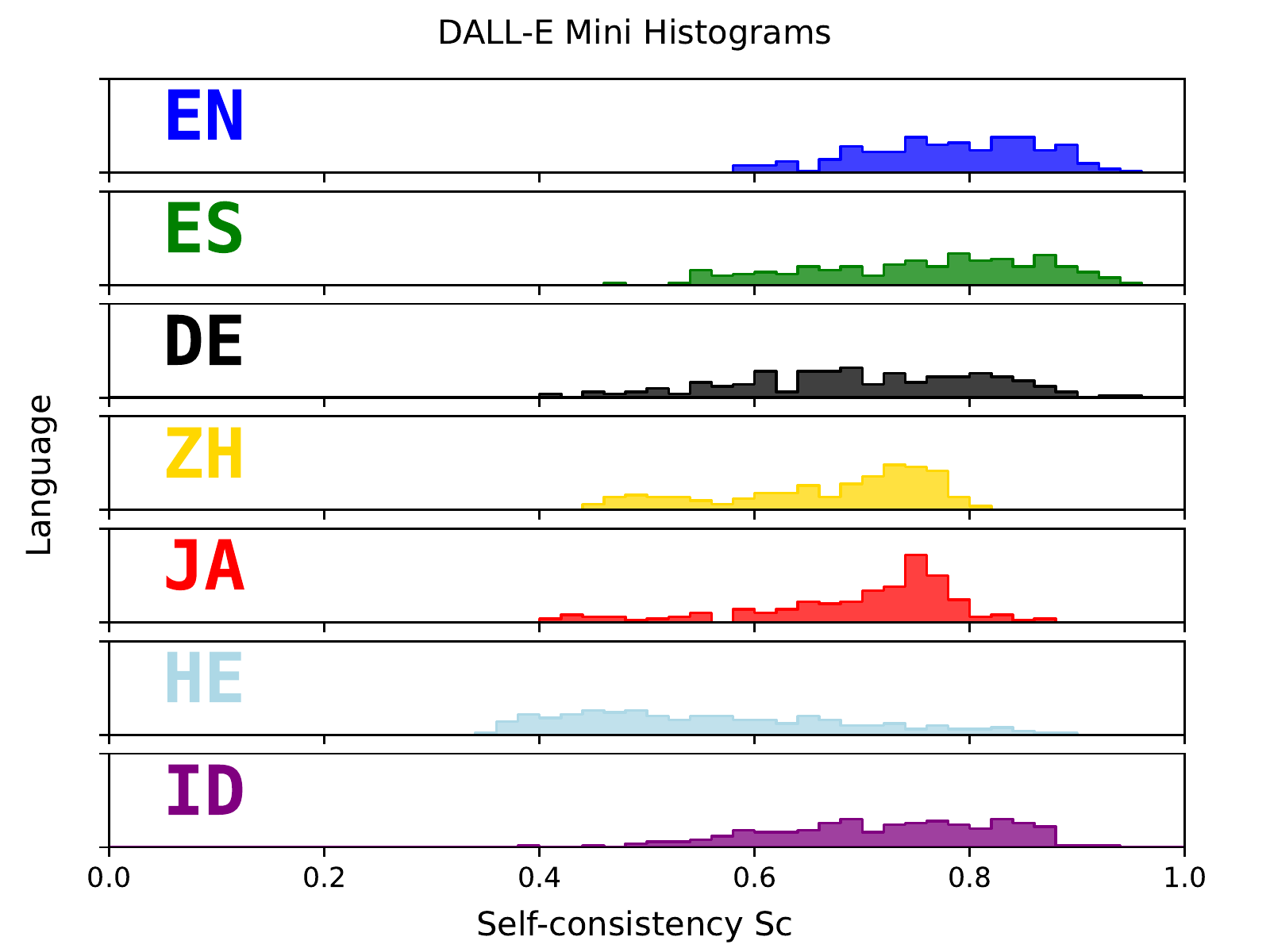}
\end{minipage}%
\begin{minipage}{.33\linewidth}
\centering
\includegraphics[width=1\linewidth]{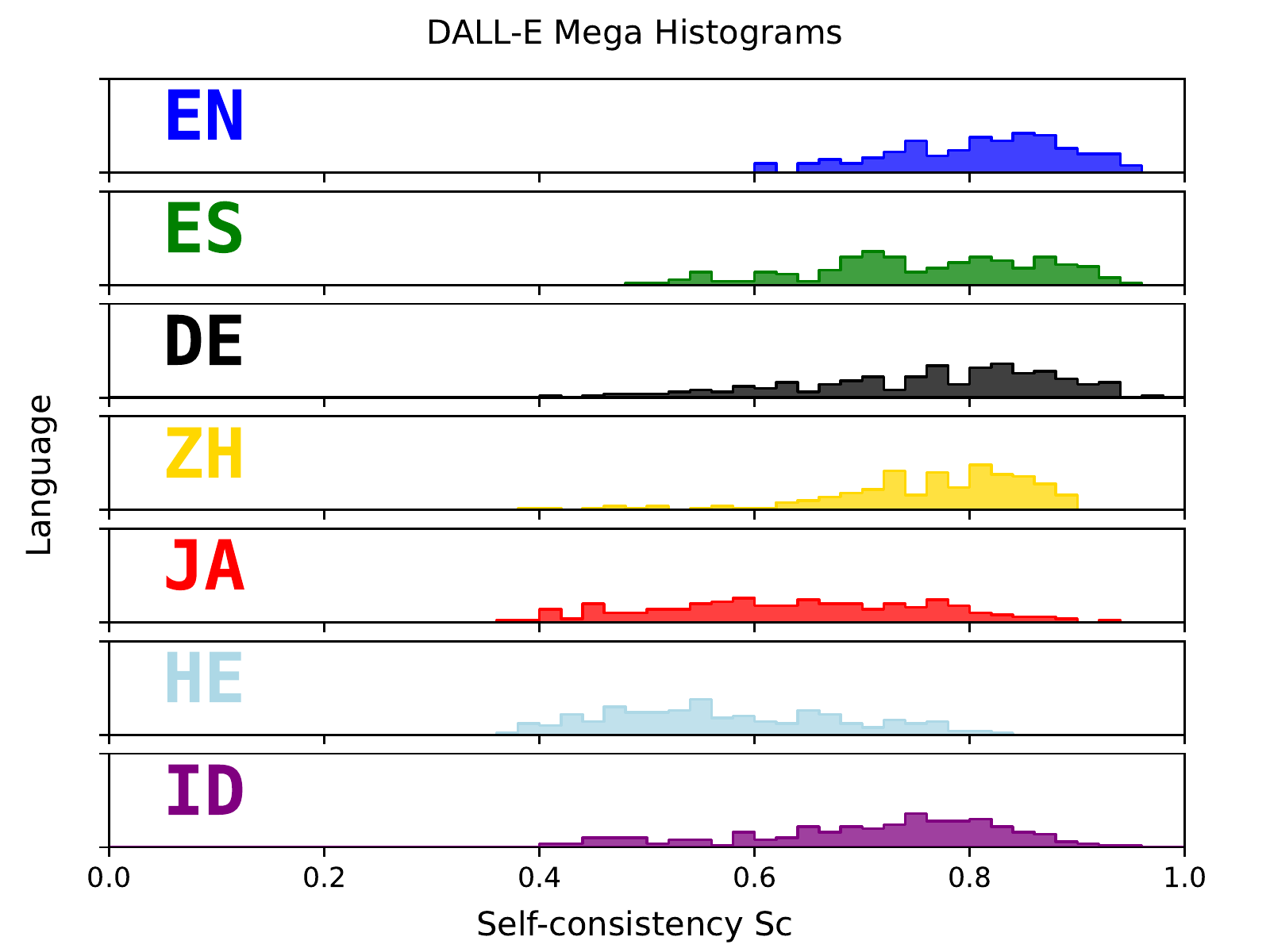}
\end{minipage}%
\begin{minipage}{.33\linewidth}
\centering
\includegraphics[width=1\linewidth]{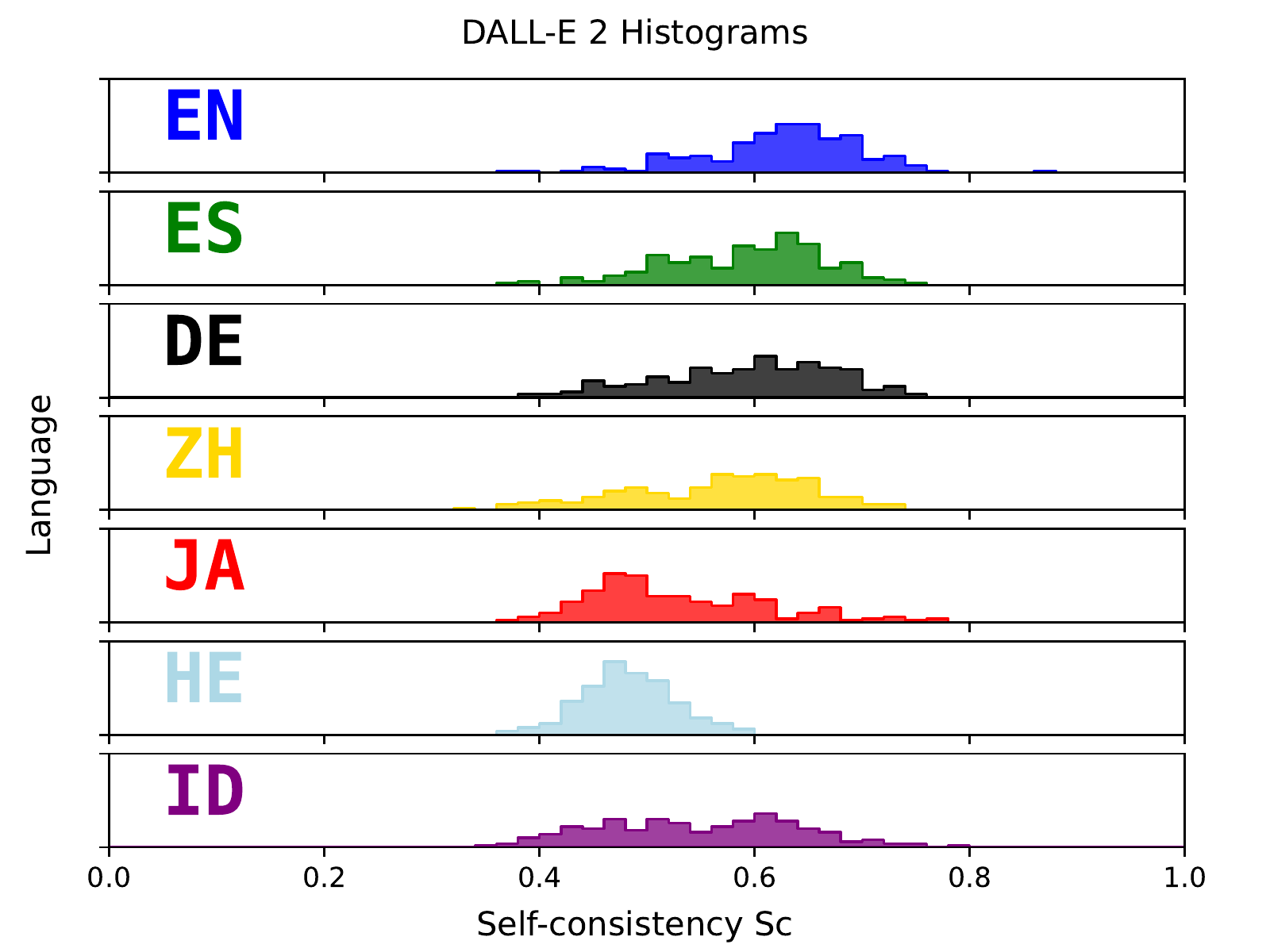}
\end{minipage}%

\begin{minipage}{.33\linewidth}
\centering
\includegraphics[width=1\linewidth]{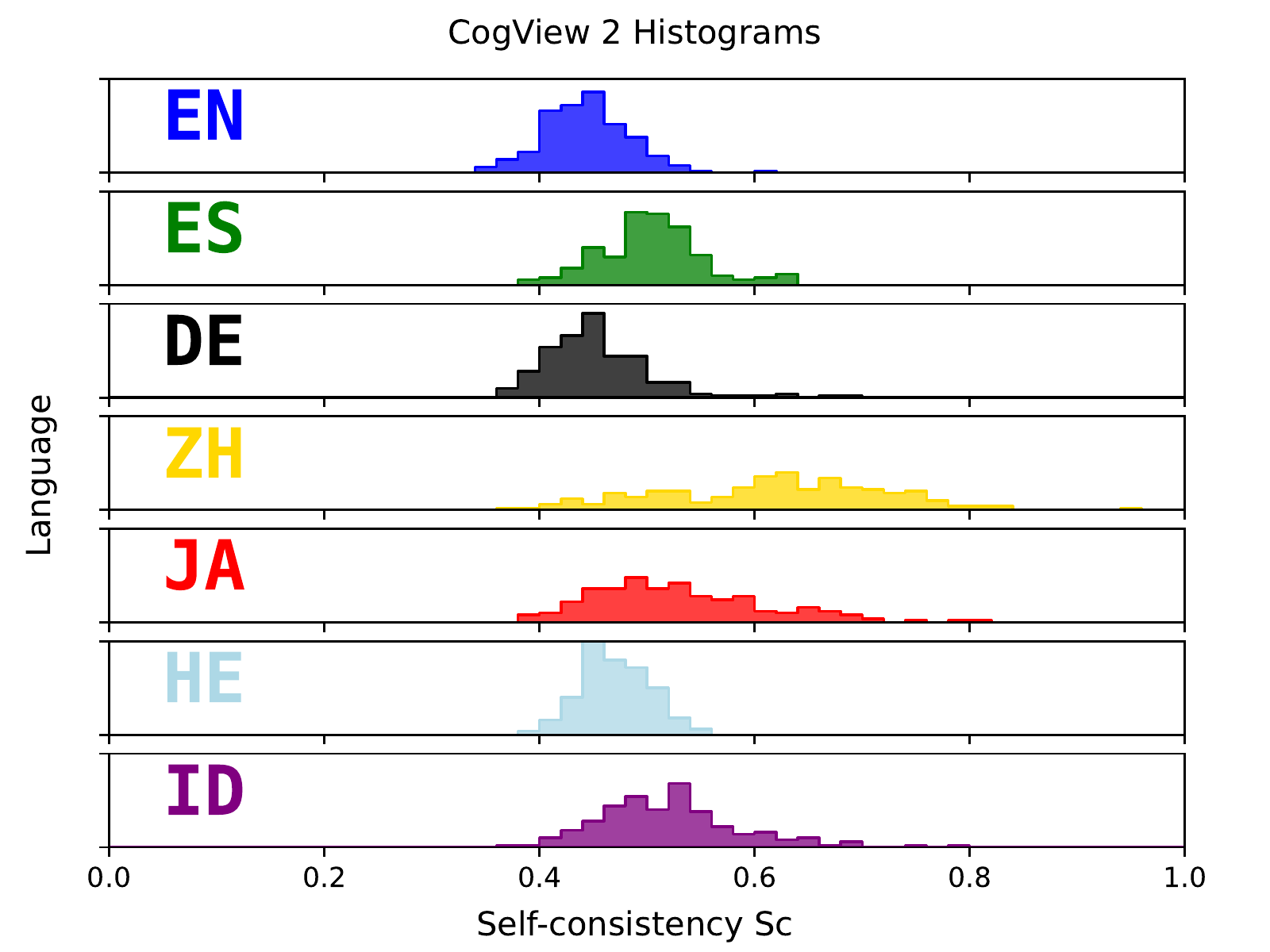}
\end{minipage}%
\begin{minipage}{.33\linewidth}
\centering
\includegraphics[width=1\linewidth]{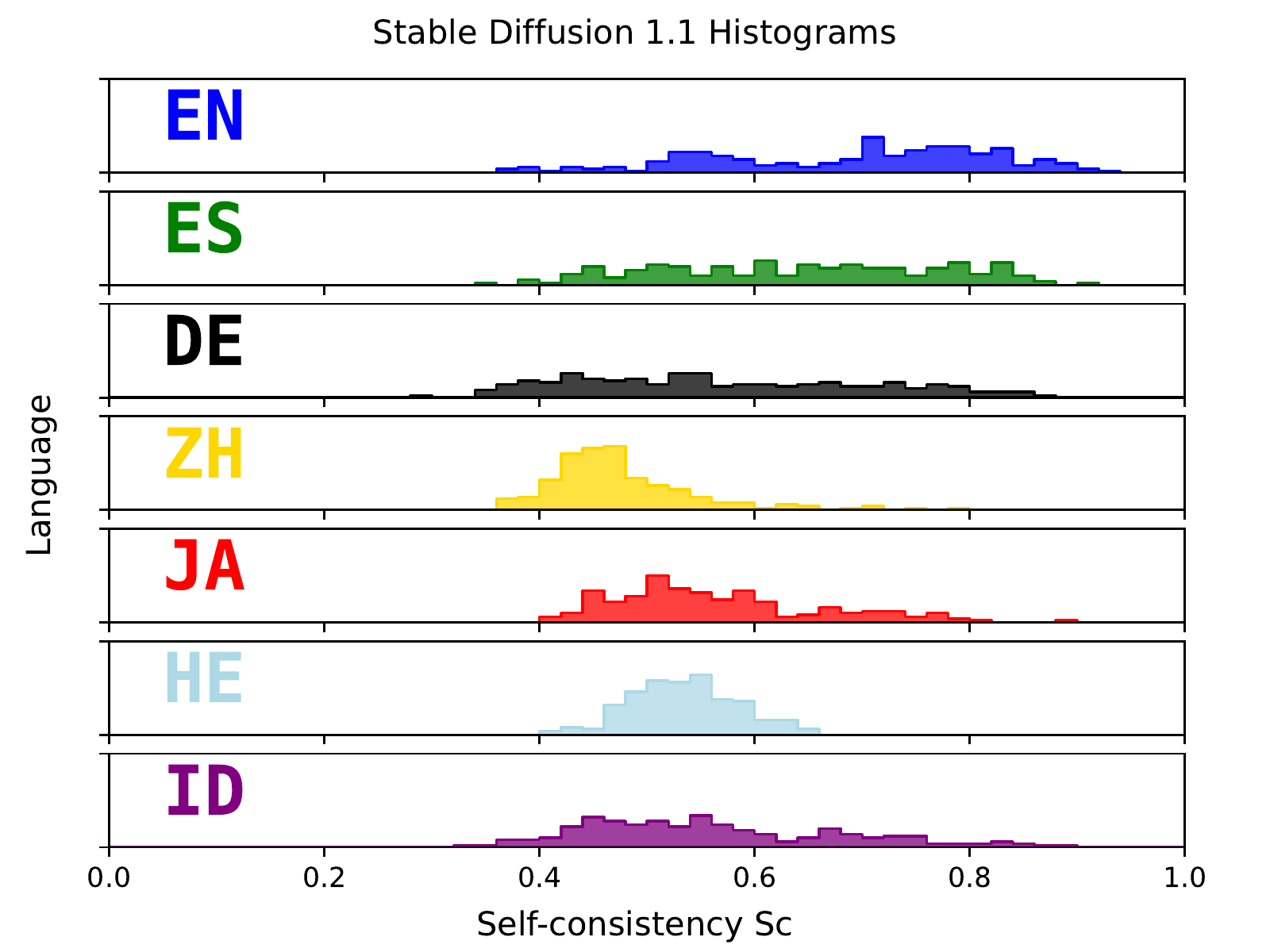}
\end{minipage}%
\begin{minipage}{.33\linewidth}
\centering
\includegraphics[width=1\linewidth]{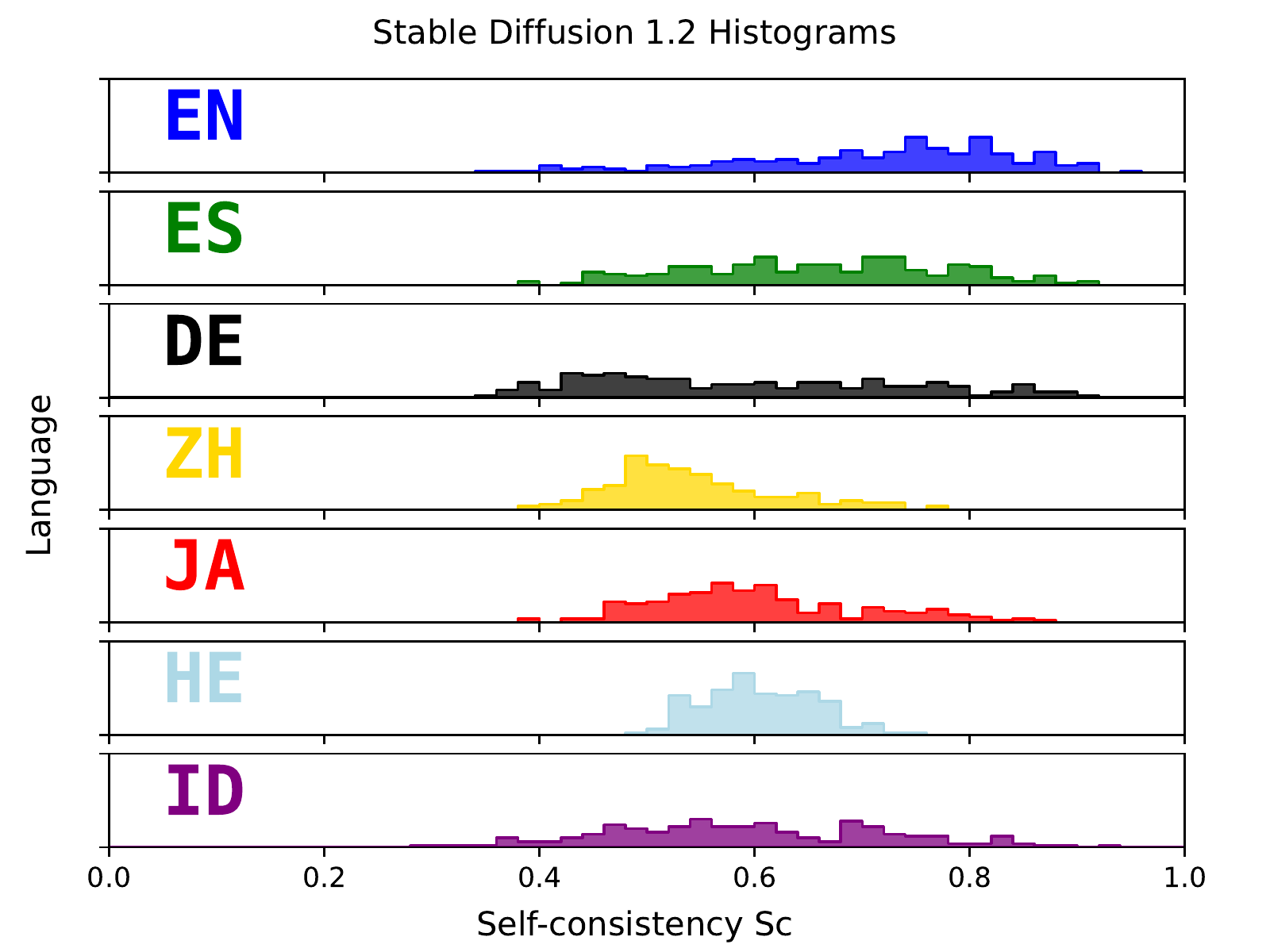}
\end{minipage}%

\begin{minipage}{.33\linewidth}
\centering
\includegraphics[width=1\linewidth]{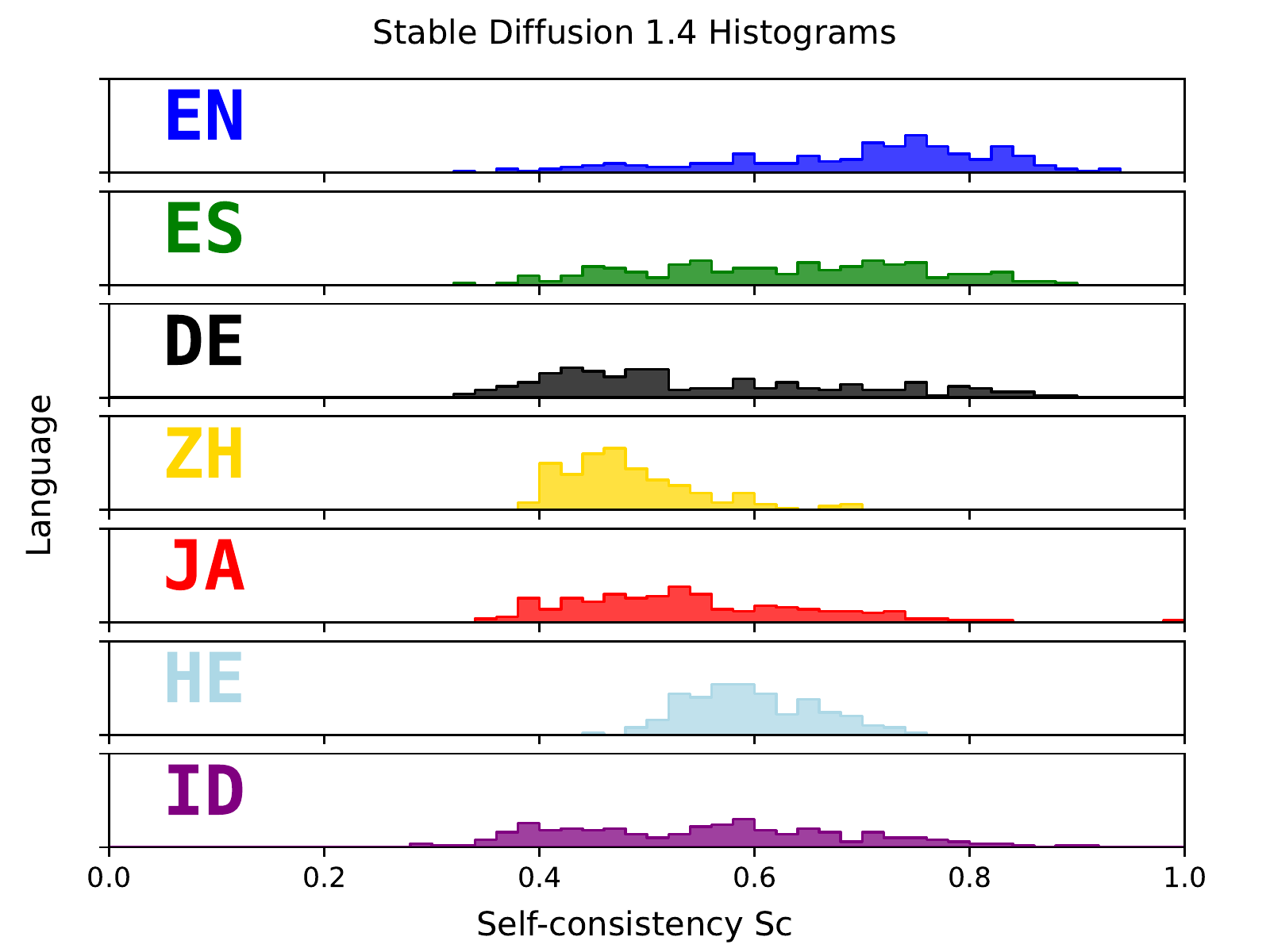}
\end{minipage}%
\begin{minipage}{.33\linewidth}
\centering
\includegraphics[width=1\linewidth]{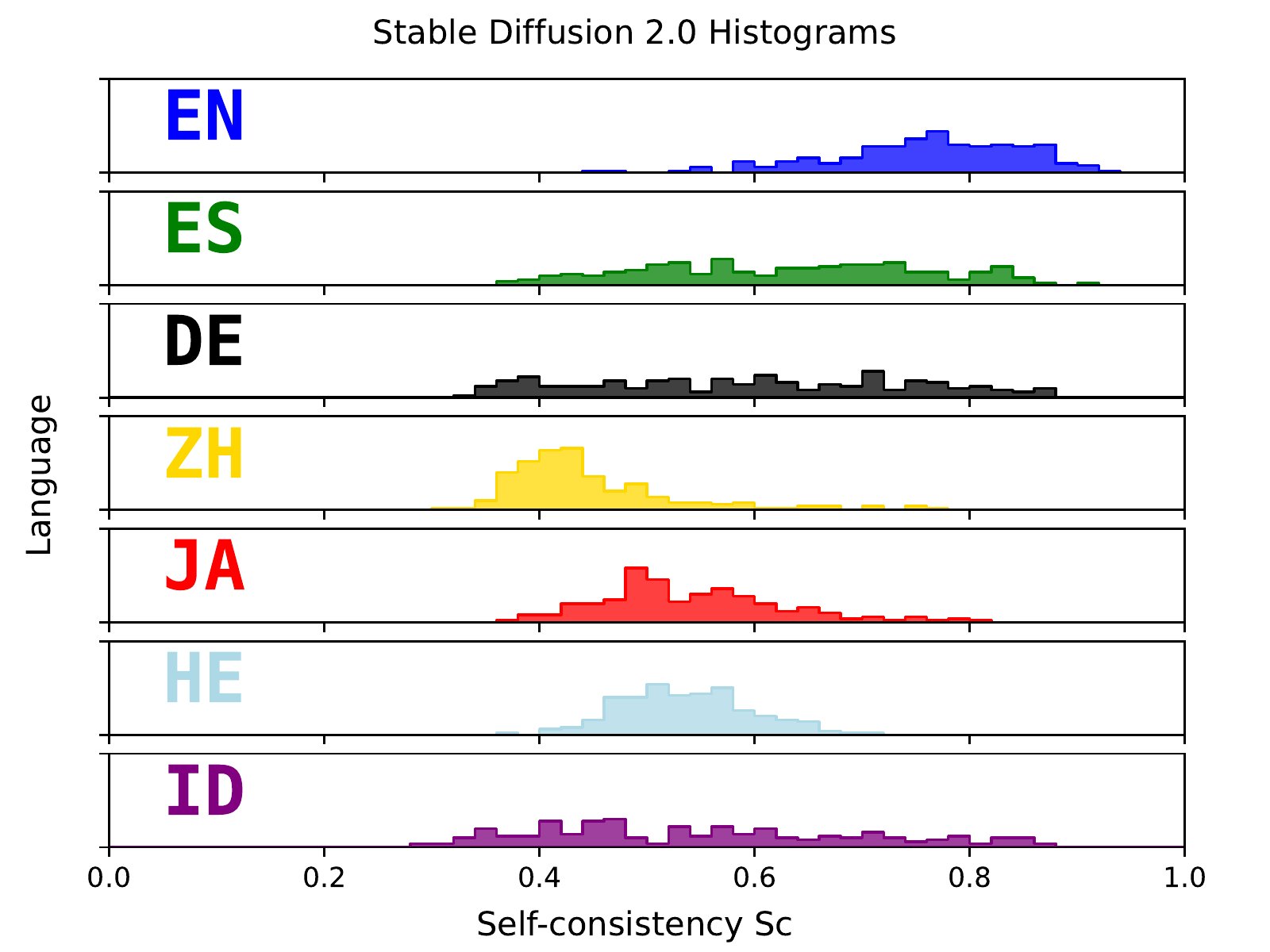}
\end{minipage}%
\begin{minipage}{.33\linewidth}
\centering
\includegraphics[width=1\linewidth]{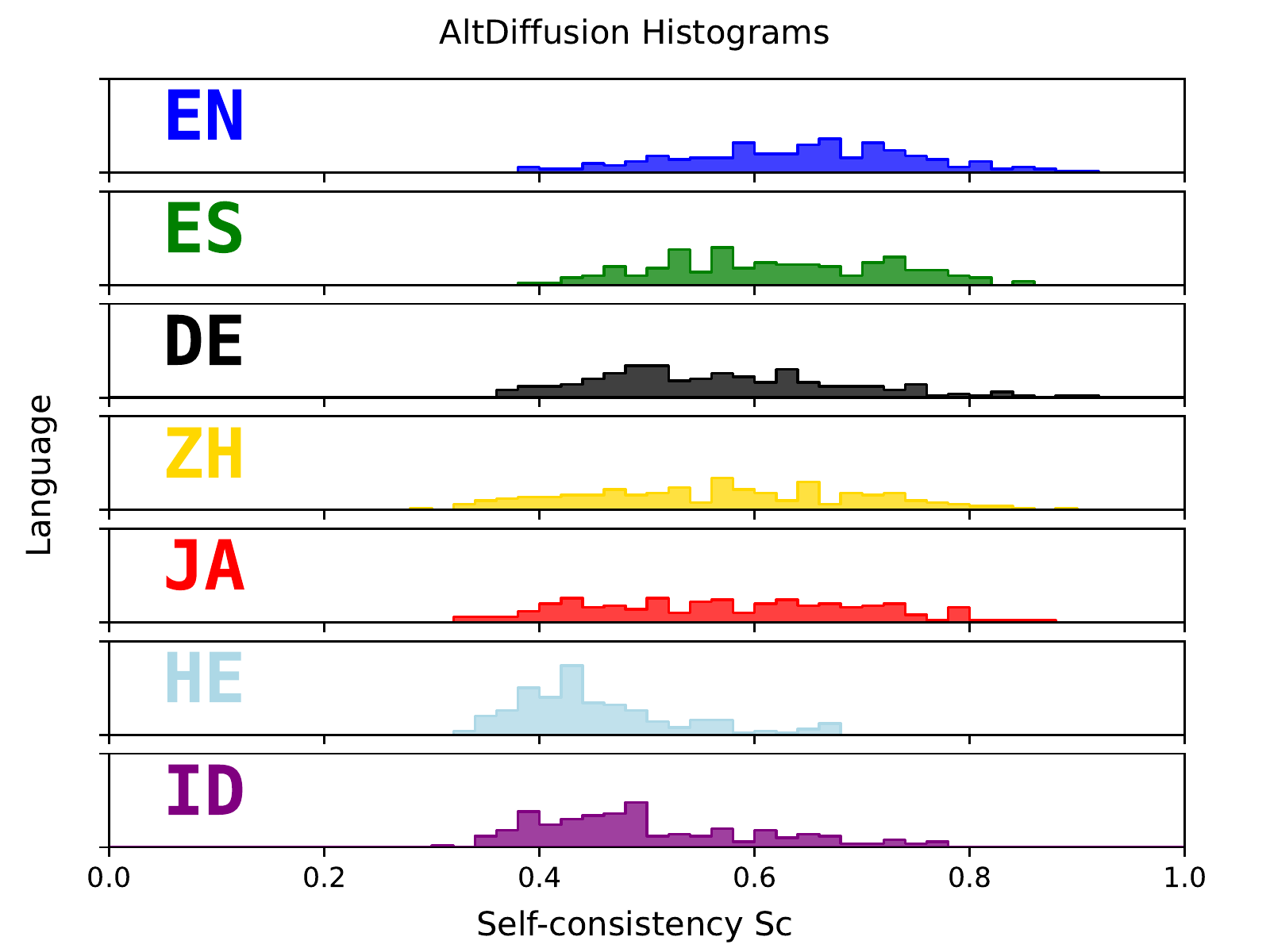}
\end{minipage}%

\caption{Distribution of {\color{conRed}\textbf{self-consistency}} scores (Sc) for each test language for all assessed models.}
\label{fig:confull}
\vspace{-5pt}
\end{figure*}

\subsection{Full concept list}

The (English) concepts are: 
eye, hand, head, smile, face, room, door, girl, person, man, love, watch, arm, hair, mother, car, mom, dad, table, phone, father, grin, mouth, kid, family, finger, world, shirt, ground, sister, chair, kitchen, woman, beer, hill, metal, hotel, princess, bench, detail, bird, cigarette, history, plastic, pizza, airplane, male, backpack, judge, dragon, sea, bike, female, garden, meal, toy, ship, flame, tail, library, weapon, cd, rope, cafeteria, porch, queen, duck, lake, television, boat, tent, roof, ticket, cop, milk, soldier, tank, thigh, belt, sandwich, bullet, teenager, apple, wine, supply, captain, cheese, feather, mask, prince, beaver, seal, stingray, shark, rose, bottle, mushroom, orange, pear, pepper, keyboard, lamp, telephone, couch, bee, beetle, butterfly, caterpillar, cockroach, tiger, wolf, bridge, castle, house, road, cloud, forest, mountain, camel, chimp, kangaroo, fox, raccoon, lobster, spider, worm, baby, crocodile, lizard, dinosaur, snake, turtle, hamster, rabbit, squirrel, tree, bicycle, train, tractor, jump, men, moon, clothes, neck, fire, tire, teacher, movie, dog, ring, eyebrow, sun, tall, doctor, sky, apartment, shoe, rock, daughter, girlfriend, bar, ball, hallway, tv, teeth, police, field, wife, brain, pants, tongue, cup, computer, bottom, bell, aunt, clock, suit, plate, chocolate, snow, guitar, truck, church, husband, van, blanket, bowl, mama, cookie, hat, monster, ceiling.

All translations are provided at the repository and demo page (\href{https://saxon.me/coco-crola}{\texttt{saxon.me/coco-crola}}).

\section{Validating the prompt templates}\label{subsec:grammareffects}

As mentioned in \autoref{subsec:minelicit}, the simple template-based approach to generating prompts for concepts leads to the introduction of grammatical errors, e.g. ``a photograph of dog.''

However, it is questionable whether small grammatical or logical errors like missing articles matters for high-resourced, well-covered languages like English. After all, the models are clearly able to generate high quality ``photograph of dog'' pictures without the word ``a'' in the sentence (\hyperref[fig:teaser]{Figure 1}). But, does the prompt phrasing matter for lower-performance languages in a model?
To investigate the impact of prompt phrasing on conceptual coverage, we tested a variety of English, Spanish and Chinese prompt phrasings on the concepts ``dog,'' ``sea,'' ``airplane,'' and ``ship'' (selected for their wide distribution across the cross-correlation correctness metric range).

For the English prompts, we experimented with including the articles ``a,'' ``the,'' ``my,'' and  ``an,'' as well as using the words ``photograph,''  ``image,'' ``photo,'' and ``picture.'' 
For Spanish, we used variations on the phrase ``un foto de'' (a photo of), including the same set of articles in English ``un/una,'' ``el/la,'' ``mi,'' ``tu,'' (your) and ``nuestra/o'' (our). 
For Chinese, we tried examples that both included and excluded the possessive particle ``\inlinezh{的}'' (de), as well as the words ``\inlinezh{照片}'' (zhaopian) and ``\inlinezh{图片}'' (tupian) for picture/photograph, and including or excluding the prepended phrase ``\inlinezh{一张}'' (yi zhang) to create the meaning ``one photograph.''
We reran the full 193 concept image generations in those three languages for Stable Diffusion 2 and AltDiffusion.

We found limited impact across all of these dimensions. Full details available in our anonymous demo at \href{https://saxon.me/coco-crola}{\texttt{saxon.me/coco-crola}}.

\section{Additional Plots}

\autoref{fig:corrful} shows the language-wise histograms for correctness scores for all nine models.

\autoref{fig:disfull} shows the language-wise histograms for inverse distinctiveness scores for all nine models. On this set of plots, the tendency for Hebrew to be an outlier in terms of inverse distinctiveness (ie, having lots of \textit{generic collisions} for concept failures) is clearly illustrated. However, other noteworthy outliers are DALL-E Mini and Mega performing worse on Chinese and Japanese (possibly script-driven) and CogView 2 having surprisingly low inverse distinctiveness for non-Chinese (non-training) languages in spite of low correctness.

\autoref{fig:confull} shows the language-wise histograms for the self-consistency scores for all nine models.

\end{document}